\documentclass[ijoc,sglanonrev]{informs4}
\usepackage{eqndefns-left} 
\RequirePackage{tgtermes}
\RequirePackage{newtxtext}
\RequirePackage{newtxmath}
\RequirePackage{bm}
\RequirePackage{endnotes}
\RequirePackage{amsmath,amsfonts,amssymb,indentfirst,subfigure,caption,hyperref,booktabs,multirow,threeparttable,verbatim,dsfont,colortbl,xcolor,makecell,bold-extra,fix-cm,fontaxes,graphicx}
\RequirePackage[numbers]{natbib}

\usepackage{hyperref}

\OneAndAHalfSpacedXII 

\usepackage{algorithm, algorithmic}
\usepackage{tikz}

\usepackage{natbib}
 \bibpunct[, ]{(}{)}{,}{a}{}{,}%
 \def\bibfont{\small}%
\EquationsNumberedThrough    

\TheoremsNumberedThrough     
\ECRepeatTheorems  %

\MANUSCRIPTNO{IJOC-0001-2024.00}

\begin{document}


 \RUNAUTHOR{Yi et al.} 

\RUNTITLE{Balancing Interpretability and Performance in Reinforcement Learning: An Adaptive Spectral Based Linear Approach}

\TITLE{Balancing Interpretability and Performance in Reinforcement Learning: An Adaptive Spectral Based Linear Approach}

\ARTICLEAUTHORS{%
\AUTHOR{Qianxin Yi\textsuperscript{a}, Shao-Bo Lin\textsuperscript{a}\thanks{corresponding author: sblin1983@gmail.com}, Jun Fan\textsuperscript{b}, Yao Wang\textsuperscript{a}}
\AFF{\textsuperscript{a}Center for Intelligent Decision Making and Machine Learning, School of Management, Xi’an Jiaotong University, Xi’an, China \\
\textsuperscript{b}Department of Mathematics, Hong Kong Baptist University, Kowloon, Hong Kong} 




} 

\ABSTRACT{Reinforcement learning (RL) has been widely applied to sequential decision making, where interpretability and performance are both critical for practical adoption. Current approaches typically focus on performance and rely on post hoc explanations to account for interpretability. Different from these approaches, we focus on designing an interpretability-oriented yet performance-enhanced RL approach. Specifically, we propose a spectral based linear RL method that extends the ridge regression–based approach through a spectral filter function. The proposed method clarifies the role of regularization in controlling estimation error and further enables the design of an adaptive regularization parameter selection strategy guided by the bias–variance trade-off principle. Theoretical analysis establishes near-optimal bounds for both parameter estimation and generalization error. Extensive experiments on simulated environments and real-world datasets from Kuaishou and Taobao demonstrate that our method either outperforms or matches existing baselines in decision quality. We also conduct interpretability analyses to illustrate how the learned policies make decisions, thereby enhancing user trust. These results highlight the potential of our approach to bridge the gap between RL theory and practical decision making, providing interpretability, accuracy, and adaptability in management contexts.
%
}%

\FUNDING{This research was supported by [grant number, funding agency].}




\maketitle


\section{Introduction}\label{sec:Intro}
In managerial environments, decision making involves both immediate consequences and long-term effects, as actions accumulate over time to shape future opportunities and overall performance. This cumulative effect makes it challenging to assess decision quality based solely on short-term outcomes, creating important modeling and optimization problems. Reinforcement learning (RL) provides a systematic framework for addressing these challenges by explicitly accounting for the impact of current actions on future outcomes \citep{cappart2022improving,du2025transfer}. Accordingly, RL \citep{gosavi2009reinforcement} has been successfully applied in various management domains, including personalized recommendation \citep{kokkodis2021demand}, dynamic treatment planning \citep{saghafian2024ambiguous}, customer acquisition \citep{song2025customer}, and behavioral operations \citep{bastani2025improving}.

A central challenge in applying RL to managerial environments is the limited ability to explore and evaluate innovative decision strategies. This limitation is particularly evident in domains such as healthcare and marketing, where decision policies often require formal approval before they can be implemented \citep{gong2024bandits}. For instance, the approval process for new drugs is typically lengthy and complex, delaying the ability to adapt treatment decisions in real time \citep{bravo2022flexible}. Similarly, in business contexts, modifications to marketing or operational strategies frequently involve formal governance procedures, which constrain opportunities for continuous experimentation \citep{hendricks1997delays}. In these settings, batch RL provides an efficient means of deriving optimal policies from fixed datasets of past decisions and outcomes, making it particularly suitable for applications with extensive historical records. Examples include treatment records in electronic health systems, driver movement logs from ride-hailing platforms, and pricing and inventory decisions routinely recorded by retail managers \citep{bastani2025improving}. Such historical datasets capture accumulated information and offer valuable opportunities for policy learning.

Interpretability remains a critical concern in the practical deployment of batch RL. For managers and frontline employees to trust, adopt, and effectively act on the recommendations produced by RL models, they must be able to understand the rationale underlying those decisions. However, many RL models currently used in decision making behave as “black boxes”, providing limited visibility into why a particular action is recommended, what information supports that recommendation, or how mistakes may arise \citep{puiutta2020explainable}. This lack of transparency can reduce user confidence and limit the broader adoption of RL in practice \citep{zhang2018exploring}. In response to these concerns, researchers and practitioners have increasingly focused on developing explainable RL methods, with several large-scale initiatives launched to advance progress in this area. For instance, the U.S. Defense Advanced Research Projects Agency (DARPA) launched the Explainable Artificial Intelligence (XAI) program in 2018 to encourage the development of high-performing models whose decision logic can be understood by human users \citep{gunning2019darpa}. More recently, scholars in the Information Systems field highlighted the importance of incorporating explainability into the design of machine learning models \citep{berente2021managing}.

While interpretability is essential for building managerial trust and supporting practical adoption, achieving high performance is equally critical to ensure the effectiveness and impact of adopted actions. In high-stakes domains such as dynamic pricing and precision medicine, suboptimal decisions can lead to financial losses or harmful interventions. For example, in pricing, RL models that fail to adapt to market dynamics may cause revenue decline, inventory misallocation, or reduced customer satisfaction \citep{bozkurt2019customers}. Similarly, in healthcare, inaccurate predictive models may recommend inappropriate treatments, thereby jeopardizing patient safety \citep{bastani2020online}. Moreover, in accuracy-sensitive areas like finance and operations, even minor errors can have significant consequences. For instance, in financial decision making, errors in reward estimation can result in poorly timed portfolio adjustments, which may cause substantial financial losses or expose the portfolio to unforeseen market risks \citep{ju2024reinforcement}. 

Considering the simultaneous need for interpretability and high performance, developing batch RL algorithms that effectively balance both objectives remains a fundamental challenge. Classical linear least squares RL approaches \citep{murphy2005generalization, goldberg2012q} provide strong interpretability, as the contribution of each feature to the decision can be clearly understood. However, these methods often perform poorly in complex or high-dimensional environments. In contrast, kernel or neural network based RL approaches \citep{wang2023kernel, fan2020theoretical} typically achieve higher prediction accuracy and better policy performance, but their complex and opaque structures make it difficult to interpret how decisions are made. 

To address the trade-off between interpretability and performance, a line of RL methods has emerged that focuses on performance-driven post hoc explanation, in which a black box model is trained first and explanations are subsequently derived. Techniques used for post hoc explanation include SHapley Additive exPlanations (SHAP) \citep{lundberg2017unified} and Local Interpretable Model-Agnostic Explanations (LIME) \citep{ribeiro2016should}. However, post hoc explanations face inherent limitations. First, the “explanations” in post hoc approaches provide concern only the model’s internal operations, not the underlying real-world mechanisms. Moreover, explanation models can be misleading: although they may match a black box’s predictive performance, they often rely on different features and thus fail to reflect the model’s true computations. Second, explanations are unavoidably imperfect. A perfectly faithful post hoc explanation would be indistinguishable from the black box itself, rendering the latter redundant. As a result, any post hoc method inevitably misrepresents the black box in parts of the feature space, making such explanations often unreliable and sometimes misleading \citep{rudin2019stop, chen2020concept}.

Given the limitations of post hoc explanations, inherently interpretable models have been suggested as an alternative \citep{rudin2019stop}, leading to growing interest in interpretability-oriented yet performance-enhanced RL methods. For example, Lasso-based RL methods \citep{oh2022generalization} promote sparsity to support feature selection, thereby improving model transparency without severely compromising performance. Nevertheless, their success is highly sensitive to the choice of regularization parameters, which typically requires computationally intensive grid search.

To address the limitations highlighted in Fig. \ref{fig:balance}, we propose an adaptive, interpretability-oriented and performance-enhanced RL method: a spectral based linear RL approach. This method improves performance via a spectral filter function and incorporates an adaptive strategy for selecting regularization parameters. These developments aim to bridge the gap between RL methodology and practical decision making, supporting management applications that demand both transparency and accuracy. Our main contributions can be summarized as follows:

\begin{figure}[htp]
    \centering
    \includegraphics[width=\linewidth]{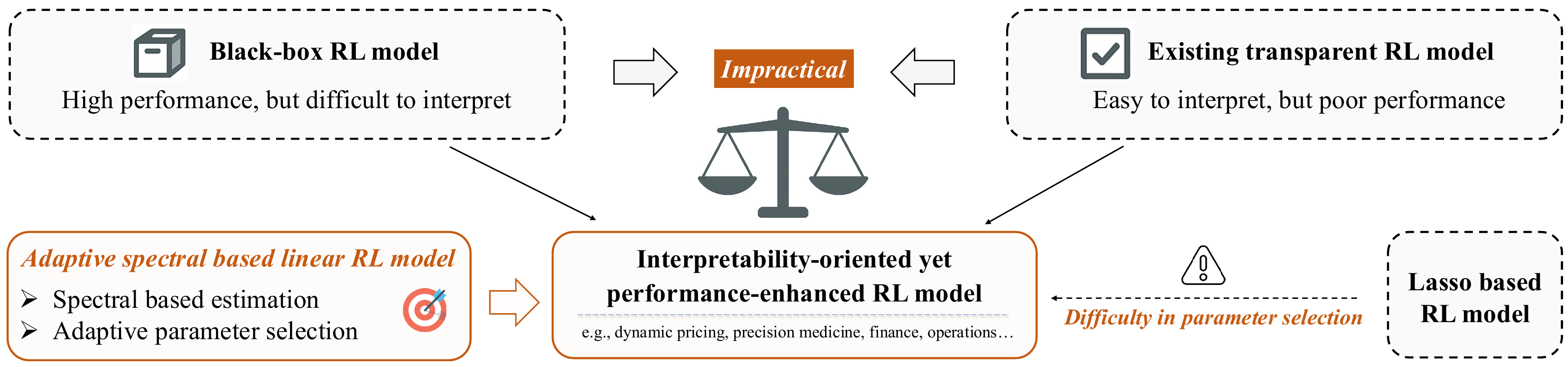}
    \caption{Motivation behind this work}
    \label{fig:balance}
\end{figure}

\begin{itemize}
    \item Methodologically, to balance interpretability and performance, we propose a spectral based linear RL method that alleviates the saturation phenomenon \citep{gerfo2008spectral, yao2007early} arising from the limited utility of additional prior information, thereby improving generalization performance. This framework also enables us to design an adaptive parameter selection strategy for the spectral based linear RL, grounded in the bias–variance trade-off principle.
    \item Theoretically, based on the relationship between batch Q-learning and multi-stage regression, we develop a novel error decomposition that incorporates multi-stage error. Leveraging this decomposition, we first derive a parameter error bound for linear regression with adaptive parameter selection through bias–variance analysis. We then adopt a recursive approach to transfer these regression results to the RL framework, establishing a near-optimal generalization error bound.
    \item Experimentally, we conduct evaluations on both simulated environments and real-world datasets from Kuaishou and Taobao, demonstrating that the proposed method either outperforms or matches relevant baselines in learning efficiency and decision quality. Furthermore, we provide detailed analyses of model interpretability, showing how the decision making process can be clearly understood and trusted. These findings also yield practical management insights, such as the advantage of simpler models or feature sets, illustrating the “less is more” principle.
\end{itemize}

The rest of this paper is as follows. Section \ref{sec:related} introduces batch Q-learning, highlighting its connection to multi-stage regression, and discusses the trade-off between interpretability and performance. It also reviews related work, including RL for decision making, explainable RL, and adaptive RL. Building on this foundation, Section \ref{sec:method} proposes spectral based linear Q-learning and then its adaptive parameter selection version. Section \ref{sec:theory} presents the theoretical analysis, including parameter estimation and generalization error bounds. Section \ref{sec:experiments} provides experimental results using both simulated environments and real-world data from Kuaishou and Taobao. Finally, Section \ref{sec:conclusion} concludes the paper and discusses future research directions.

\section{Problem setting and related work}\label{sec:related}

This section first introduces batch Q-learning and its connection to multi-stage regression, followed by a discussion of the trade-off between interpretability and performance. It then reviews related work, including RL for sequential decision making, explainable RL and adaptive RL.

\subsection{Formulation connection: from batch Q-learning to multi-stage regression}

We consider a $T$-stage decision problem. For each stage $t$, $s_t \in \mathcal{S}_t$ denotes the state and $a_t \in \mathcal{A}_t$ represents the action, where $\mathcal{S}_t$ and $\mathcal{A}_t$ are the respective state and action spaces. The cumulative state and action spaces are denoted as $\mathcal{S}_{1:t} = \mathcal{S}_1 \times \mathcal{S}_2 \times \cdots \times \mathcal{S}_t$ and $\mathcal{A}_{1:t} = \mathcal{A}_1 \times \mathcal{A}_2 \times \cdots \times \mathcal{A}_t$. The outcome $R_t : (\mathcal{S}_{1:t+1}, \mathcal{A}_{1:t}) \rightarrow \mathbb{R}$ depends on the state transition $s_{1:t+1}$ and past actions $a_{1:t}$, where $s_{1:t} = \{s_1, s_2, \ldots, s_t\}$ and $a_{1:t} = \{a_1, a_2, \ldots, a_t\}$ capture the historical states and actions up to stage $t$, respectively.  The trajectory is  $\mathcal{T}_T = \{s_{1:T+1}, a_{1:T}\}$, with $s_{T+1}$ being the state following all actions. We consider a setting in which only datasets $D:=\left\{\mathcal{T}_{i, T}, r_{i, 1: T}\right\}_{i=1}^{|D|}$ are available, with $\left\{\mathcal{T}_{i, T}\right\}_{i=1}^{|D|}= \left\{\left(s_{i, 1: T+1}, a_{i, 1: T}\right)\right\}_{i=1}^{|D|}$, $ r_{i, t}:= R_t\left(s_{i, 1: t+1}, a_{i, 1: t}\right)$, and $|D|$ denoting the dataset cardinality.

A policy $\pi = (\pi_1, \ldots, \pi_T)$ is a set of decision rules, where $\pi_t : \mathcal{S}_{1:t} \times \mathcal{A}_{1:t-1} \rightarrow \mathcal{A}_t$, specifying the action selection strategy at each stage. The optimal policy maximizes the total outcome $\sum_{t=1}^T R_t(s_{1:t+1}, a_{1:t})$. The transition probability $\rho_t(s_t \mid s_{1:t-1}, a_{1:t-1})$ defines the probability of transitioning to state $s_t$ given prior states and actions. The value of $\pi$ at stage $t$ is 
$$
V_{\pi, t}\left(s_{1: t}, a_{1: t-1}\right)=E_\pi\left[\sum_{j=t}^T R_j\left(S_{1: j+1}, A_{1: j}\right) \mid S_{1: t}=s_{1: t}, A_{1: t-1}=a_{1: t-1}\right],
$$
where $E_\pi$ denotes the expectation under the distribution
$$
P_\pi=\rho_1\left(s_1\right) 1_{a_1=\pi_1\left(s_1\right)} \prod_{t=2}^T \rho_t\left(s_t \mid s_{1: t-1}, a_{1: t-1}\right) 1_{a_t=\pi\left(s_{1: t}, a_{1: t-1}\right)} \rho_{T+1}\left(s_{T+1} \mid s_{1: T}, a_{1: T}\right),
$$
and $1_W$ denotes the indicator function for event $W$. The optimal value function of $\pi$ at stage $t$ is 
$
V_t^*\left(s_{1: t}, a_{1: t-1}\right)=\max _{\pi \in \Pi} V_{\pi, t}\left(s_{1: t}, a_{1: t-1}\right),
$
where $\Pi$ represents the set of all possible policies. Our goal is to identify a policy $\hat{\pi}$ to minimize $V_1^*\left(s_1\right)-V_{\hat{\pi}, 1}\left(s_1\right)$. The time-dependent Q-function is 
$$
Q_{\pi,t}\left(s_{1: t}, a_{1: t}\right)
=E\left[R_t\left(S_{1: t+1}, A_{1: t}\right)+V_{\pi,t+1}\left(S_{1: t+1}, A_{1: t}\right) \mid S_{1: t+1}=s_{1: t+1},A_{1: t}=a_{1: t}\right],
$$
and the corresponding optimal time-dependent Q-function is given by
\begin{equation}
   Q_t^*\left(s_{1: t}, a_{1: t}\right)=E\left[R_t\left(S_{1: t+1}, A_{1: t}\right)+V_{t+1}^*\left(S_{1: t+1}, A_{1: t}\right) \mid S_{1: t+1}=s_{1: t+1},A_{1: t}=a_{1: t}\right], \label{Q general}
\end{equation}
where $E$ denotes the expectation taken with respect to the distribution $P:=P_{T+1}$ and
$$
P_t=\rho_1\left(s_1\right) p_1\left(a_1 \mid s_1\right) \prod_{j=2}^t \rho_j\left(s_j \mid s_{1: j-1}, a_{1: j-1}\right) p_j\left(a_j \mid s_{1: j}, a_{1: j-1}\right),
$$
where $p_t\left(a_t \mid s_{1: t}, a_{1: t-1}\right)$ denote the probability of choosing action $a_t$ given the history $\left\{s_{1: t}, a_{1: t-1}\right\}$. 
According to the definition of $V_t^*$, it follows that \citep{murphy2005generalization}
$$
V_t^*\left(s_{1: t}, a_{1: t-1}\right)=V_{\pi^*, t}\left(s_{1: t}, a_{1: t-1}\right)
=E_{\pi^*}\left[\sum_{j=t}^T R_j\left(S_{1: j+1}, A_{1: j}\right) \mid S_{1: t}=s_{1: t}, A_{1: t-1}=a_{1: t-1}\right],
$$
where $\pi^*$ represents the optimal policy. Consequently, we have
\begin{equation}
V_t^*\left(s_{1: t}, a_{1: t-1}\right)=\max _{a_t} Q_t^*\left(s_{1: t}, a_{1: t}\right).\label{vq general}
\end{equation}
This formulation demonstrates that optimal decisions can be determined by maximizing the optimal Q-functions. With $Q_{T+1}^*\left(s_{1: T+1}, a_{1: T+1}\right)=0$, combining equations  (\ref{Q general}) and (\ref{vq general}) shows that
\begin{equation}
\begin{aligned}
&Q_t^*\left(s_{1: t}, a_{1: t}\right)\\
=&E\left[R_t\left(S_{1: t+1}, A_{1: t}\right)+\max _{a_{t+1}} Q_{t+1}^*\left(S_{1: t+1}, A_{1: t}, a_{t+1}\right)\mid S_{1: t+1}=s_{1: t+1}, A_{1: t}=a_{1: t}\right].\label{property:q1 general}
\end{aligned}
\end{equation}
This property links Q-functions with the regression function \citep{gyorfi2006distribution}. Let $\mathcal{X}_t=\mathcal{S}_{1: t+1} \times \mathcal{A}_{1: t}$, $x_t:=\left\{s_{1: t+1}, a_{1: t}\right\} \in \mathcal{X}_t$, and $
    y_t^*:=r_t\left(s_{1: t+1}, a_{1: t}\right)+\max _{a_{t+1}} Q_{t+1}^*\left(s_{1: t+1}, a_{1: t}, a_{t+1}\right)
$, then $Q_t^*=E[Y_t^*\mid X_t]$. Therefore, the standard approach in statistical learning theory \citep{gyorfi2006distribution} yields
\begin{equation}\label{Q regression general}
\begin{aligned}
Q_t^*=\arg \min _{Q_t } E\left[\left(Y_t^*-Q_t(X_t)\right)^2\right], \quad t=T, T-1, \ldots, 1,
\end{aligned}
\end{equation}
showing that optimal Q-functions can be obtained by solving $T$ least squares problems. 

\subsection{Intrinsic phenomenon: interpretability and performance trade-off}


One widely adopted approach is to represent the Q-function $Q_t(x_t)$ as a linear function $
Q_t(x_t) = \langle x_t, \theta_t\rangle
$,
where $\theta_t$ is a stage-dependent parameter vector \citep{murphy2005generalization, goldberg2012q}. This linear representation provides explicit transparency, as each coefficient in $\theta_t$ reflects the weight of a feature, enabling domain experts to trace how states and actions affect decision quality. Nonetheless, achieving such high interpretability typically entails a reduction in performance. In more complex scenarios with nonlinear dependencies, the linear model may fail to capture the true underlying structure, often leading to poor predictive accuracy for $Y_t^*$ and suboptimal policies.

To overcome the performance limitations of the linear representation, researchers have turned to nonlinear function representations. In kernel-based RL approaches \citep{wang2023kernel}, the Q-function is
$
Q_t(x_t) = \sum_{i=1}^{n} \alpha_i K(x_t, x_i)
$,
where $K(\cdot,\cdot)$ is a kernel function and $\alpha_i$ are coefficients learned from the data. In deep RL approaches \citep{fan2020theoretical}, the Q-function is parameterized as
$
Q_t(x_t) = f_{\theta_t}^{\text{NN}}(x_t)
$,
where $f_{\theta_t}^{\text{NN}}$ is a multi-layer network with parameters $\theta_t$ that captures complex nonlinear relationships. By expanding the space of candidate functions, these nonlinear models often achieve superior performance, but this improvement comes at the cost of interpretability. The learned decision rules cannot be easily broken down into the contributions of individual features, and the mechanisms behind predictions are generally treated as black boxes. This lack of transparency may limit trust and raise concerns in safety-critical or regulated applications.

\begin{figure}\vspace{-4mm}
    \centering
    \includegraphics[width=0.45\linewidth]{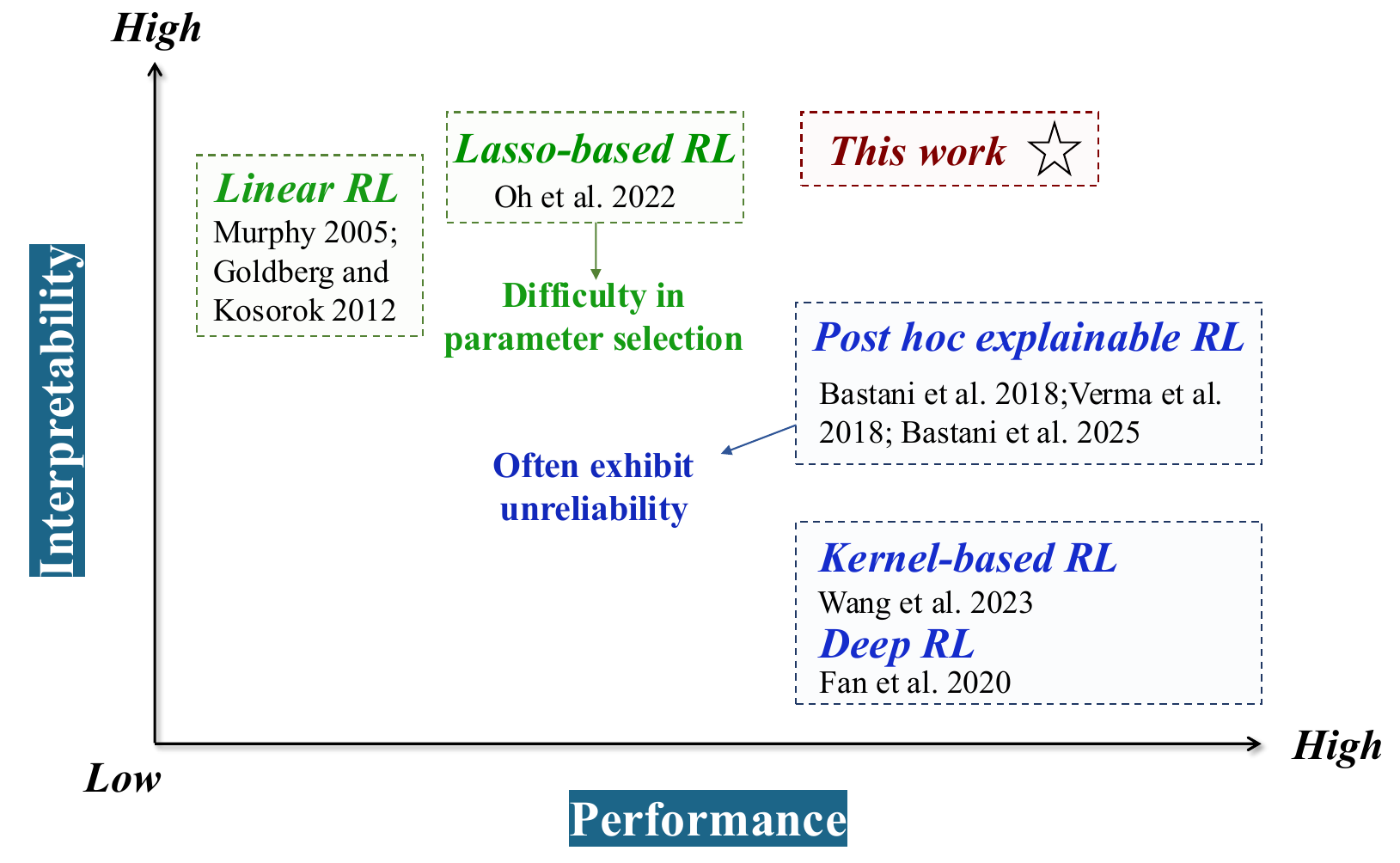}
    \caption{Interpretability and performance trade-off}
    \label{fig:tradeoff}
\end{figure}

This motivates the exploration of the trade-off between interpretability and performance. One approach is post hoc explainable methods, which first train a black-box model and then derive explanations \citep{bastani2018verifiable, verma2018programmatically,bastani2025improving}. However, such methods often suffer from unreliability \citep{rudin2019stop, chen2020concept}. An alternative is inherently interpretable models, such as linear models incorporating sparsity via Lasso regularization \citep{oh2022generalization}. In this case, the parameter vector is estimated as
$
\hat{\theta}_{D,\lambda_t,t} ^{Lasso}= \arg \min_{\theta_t}  \widehat{E}_{D}[(Y_{t}^* - \langle X_{t}, \theta_t \rangle)^2]+ \lambda_t \|\theta_t\|_1, 
$
where $\widehat{E}_{D}[(Y_{t}^* - \langle X_{t}, \theta_t \rangle)^2]=\frac{1}{|D|} \sum_{i=1}^{|D|} \left( y_{t,i}^* - \langle x_{t,i}, \theta_t \rangle \right)^2$, $\lambda_t$ is the regularization parameter, and the $\ell_1$ penalty promotes sparsity by shrinking irrelevant coefficients to zero. However, the effectiveness of Lasso critically depends on the choice of $\lambda_t$. A value that is too small fails to eliminate noisy variables, reducing interpretability and increasing the risk of overfitting, whereas a value that is too large may exclude important features, resulting in underfitting and weaker policy performance. Selecting $\lambda_t$ usually relies on grid search or cross-validation, which is computationally expensive and often impractical. Thus, while sparse regularization offers an effective way to balance interpretability and performance, its dependence on parameter tuning presents challenges for scalability. Based on the above, Fig. \ref{fig:tradeoff} summarizes the trade-off between interpretability and performance.

\subsection{Related work}

This study relates to three main areas of research. One line of work focuses on RL methods for sequential decision making. Another concerns explainable RL, which aims to improve the transparency and interpretability of model behavior. A third area involves adaptive RL, which develops methods to enhance the practicality of RL through adaptive parameter selection mechanisms.

\subsubsection{Reinforcement learning for sequential decision making.}

RL has emerged as a widely adopted and powerful framework for sequential decision making, offering significant advantages over traditional multi-armed bandit (MAB) models \citep{lattimore2020bandit}. While MABs focus on short-term outcomes, RL explicitly models state dynamics and long-term returns, enabling its successful application across diverse domains. For instance, in recommendation systems, RL facilitates dynamic learning pathways that guide users in acquiring in-demand skills by forecasting market trends and maximizing long-term outcomes \citep{kokkodis2021demand}. 

Among various RL methods, Q-learning \citep{watkins1992q} is one of the most widely adopted value-based algorithms, recognized for its model-free nature, ease of implementation, and solid theoretical foundation. This type of method works by estimating the action-value function (Q-function), which guides the agent in selecting actions that maximize long-term rewards. To support a wide range of application scenarios, several variants of Q-learning have been proposed, including linear Q-learning, kernel-based Q-learning, and deep Q-learning. Linear Q-learning employs linear models, which are computationally efficient and interpretable, making them well suited for structured data \citep{murphy2005generalization}. Kernel-based Q-learning uses reproducing kernel Hilbert spaces (RKHS) to enable modeling of complex, nonlinear patterns in the state space \citep{wang2023kernel}. Deep Q-learning employs deep neural networks to approximate the Q-function, enabling the algorithm to handle high-dimensional and unstructured inputs effectively \citep{lin2023lifting}. Among these, the setting in \citep{lin2023lifting} is most similar to ours, differing only in the solution approach.

\subsubsection{Explainable reinforcement learning.}

Explainability is increasingly important in RL, especially for management decision making tasks \citep{berente2021managing}. However, ensuring interpretability remains difficult, as RL models are often built on black box architectures and involve sequential decisions aimed at long-term objectives, which makes their logic harder to trace \citep{song2025customer}. This lack of transparency can frustrate users, weaken their confidence in the system, and limit its practical adoption \citep{zhang2018exploring}. Enhancing model interpretability enables decision-makers to understand the rationale behind actions and the relevance to decision processes.

Most existing explainable RL approaches are performance-driven and obtain interpretability post hoc, training a black box model first and deriving explanations afterward. For example, \cite{bastani2018verifiable} introduced the Verifiability via Iterative Policy ExtRaction (VIPER) algorithm, which extracts a neural network policy into a decision tree to improve interpretability and enable formal verification. Similarly, \cite{verma2018programmatically} developed Programmatically Interpretable Reinforcement Learning (PIRL), which approximates a neural policy using programs written in a high-level, domain-specific language, allowing symbolic reasoning about policy behavior. More recently, \cite{bastani2025improving} inferred an interpretable decision rule (tip) that minimizes the difference between existing human policies and black box recommendations. 

Despite recent progress, post hoc explainable methods often exhibit unreliability \citep{rudin2019stop, chen2020concept}. An alternative is to learn inherently transparent policies. Linear RL models exemplify this approach by representing the value function linearly, revealing the relationship between features and decision outcomes. For instance, least-squares RL methods \citep{murphy2005generalization} allow direct evaluation of each feature weight, providing immediate interpretability, although their predictive performance is often limited. Additionally, Lasso-based RL methods \citep{oh2022generalization} aim to identify the most relevant features through sparsity regularization, which enhances interpretability but requires careful tuning of the regularization parameters.

\subsubsection{Adaptive reinforcement learning.}

Adaptive parameter selection is critical in RL, but it remains underexplored in most current methods. This has recently attracted attention in the bandit literature, since the regret performance of Upper Confidence Bound (UCB) based algorithms is sensitive to confidence bound parameters. These parameters often vary with application contexts and are difficult to tune in real-time \citep{bouneffouf2020hyper, ding2022syndicated}. Traditional tuning methods such as cross-validation \citep{stone1974cross} or Bayesian optimization \citep{frazier2018tutorial} are not well suited for online decision making. To address this, \cite{bouneffouf2020hyper} proposed a two-level framework that treats parameter choices as arms in a bandit problem, using Thompson Sampling (TS) or decision trees for selection. \cite{ding2022syndicated} extended this idea by employing the EXP3 algorithm \citep{auer2002nonstochastic}, enabling the selection of multiple parameters. More recently, \cite{kang2024online}, building on the Bandit-over-Bandit (BOB) framework \citep{cheung2019learning}, proposed a method based on Zooming TS to select parameters from continuous spaces.

\section{Methodology}\label{sec:method}
This section first outlines the road-map of the proposed spectral based linear Q-learning, followed by a practical algorithm capable of adaptive parameter selection.

\subsection{Road-map: spectral based linear Q-learning}

To address the trade-off between interpretability and performance, we propose a spectral based linear RL approach. We begin with the linear RL framework, where the linear Q-function is 
$$
Q_{\pi,t}\left(s_{1: t}, a_{1: t}\right)=\langle x_t,\theta_{\pi,t}\rangle
=E\left[R_t\left(X_t\left(S_{1: t+1}, A_{1: t}\right)\right)+V_{\pi,t+1}\left(X_t\left(S_{1: t+1}, A_{1: t}\right)\right) \mid X_t=x_t\right],
$$
and the corresponding optimal linear Q-function is given by
\begin{equation}
\begin{aligned}
&Q_t^*\left(s_{1: t}, a_{1: t}\right)=\langle x_t,\theta_t^*\rangle\\
=&E\left[R_t\left(X_t\left(S_{1: t+1}, A_{1: t}\right)\right)+\max _{a_{t+1}} \langle X_t\left(S_{1: t+1}, A_{1: t}, a_{t+1}\right),\theta_{t+1}^*\rangle \mid X_t=x_t\right].\label{property:q1}
\end{aligned}
\end{equation}
Denote
\begin{equation}\label{ytstar}
    y_t^*:=r_t\left(x_t(s_{1: t+1}, a_{1: t})\right)+\max _{a_{t+1}} \langle\theta_{t+1}^*,x_t\left(s_{1: t+1}, a_{1: t}, a_{t+1}\right)\rangle.
\end{equation}
From (\ref{property:q1}), with $\theta_{T+1}^*=0$, the following holds:
\begin{equation}\label{property:q}
\begin{aligned}
&\langle x_t, \theta_t^*\rangle=E[Y_t^*\mid X_t=x_t]\\
=&E\left[R_t\left(X_t(S_{1: t+1}, A_{1: t})\right)+\max _{a_{t+1}} \langle\theta_{t+1}^*,X_t\left(S_{1: t+1}, A_{1: t}, a_{t+1}\right)\rangle \mid X_t=x_t\right].
\end{aligned} 
\end{equation}

To estimate the parameter $\theta_t^*$, we utlize the spectral based linear estimation. Let the covariance matrix be defined as $\Sigma_t=E[X_tX_t^\top]$ and the empirical covariance matrix as $\widehat{\Sigma}_{D,t}=\frac{1}{|D|} \sum_{i=1}^{|D|}x_{i,t}x_{i,t}^\top$. Given regularization parameters $\lambda_t$ for $t=1, \ldots, T$, with $\lambda_{T+1}=0$ and $\theta_{D, \lambda_{T+1}, T+1}=0$, the parameter vectors $\theta_t^*$ are empirically estimated using the spectral based linear method, defined as
$$\theta_{D, \lambda_t, t}=g_{\lambda_t}\left(\widehat{\Sigma}_{D,t}\right) \widehat{E}_{D}[X_t Y_t],$$ where $g_{\lambda_t}$ is the spectral filter function (see examples in Table~\ref{sb examples}), $\widehat{E}_{D}[X_t Y_t]=\frac{1}{|D|} \sum_{i=1}^{|D|}x_{i,t}y_{i,t}$, and
\begin{equation}
y_{i, t}:=r_{i, t}(x_{i,t}\left(s_{i, 1: t+1}, a_{i, 1: t}\right))+\max _{a_{t+1} \in \mathcal{A}_{t+1}} \langle\theta_{D, \lambda_{t+1}, t+1},x_{i,t}\left(s_{i, 1: t+1}, a_{i, 1: t}, a_{t+1}\right)\rangle.\label{prop:y bound}
\end{equation}
Define 
\begin{equation}
\begin{aligned}
&\langle x_t,\theta^*_{D, \lambda_t, t}\rangle=E[Y_t\mid X_t=x_t]\\
=&E\left[R_t\left(X_t( S_{1:t+1},A_{1:t})\right)+\max _{a_{t+1}\in \mathcal{A}_{t+1}} \langle\theta_{D, \lambda_{t+1}, t+1},X_t\left( S_{1:t+1},A_{1:t}, a_{t+1}\right)\rangle \mid X_t=x_{ t}\right].\label{qstard}
\end{aligned}
\end{equation}

\begin{table*}[htbp]\vspace{-9mm}
	\small
		\centering
		\caption{Examples of spectral filter function}\label{sb examples}
		\begin{tabular*}{0.9\textwidth}{cccc}
			\toprule
			 \ \ \ \ \ Method \ \ \ \ \ & \ \ \ \ \ Filter Function $ g_\lambda(\sigma)$\ \ \ \ \ & \ \ \ \ \  $ b$ \ \ \ \ \ &\ \ \ \ \ $ \nu_g$\ \ \ \ \ \\
			\midrule
		Tikhonov regularization / regularized least squares  & $\frac{1}{\sigma+\lambda}$& $1$&$1$\\
		Spectral cut-off &$
 \begin{cases}\frac{1}{\sigma}, & \text { if } \sigma \geqslant \lambda \\ 0, & \text { if } \sigma<\lambda .\end{cases}
$  &$1$&$\infty$\\
  Gradient descent & $\sum_{i=0}^{p-1}(1-\sigma)^i$  &$1$  &$\infty$\\
			\bottomrule
		\end{tabular*}
	\end{table*}
The spectral filter function $g_{\lambda_t}\left(\widehat{\Sigma}_{D,t}\right)$ mentioned above is used to approximate the inverse of the empirical covariance matrix $\widehat{\Sigma}_{D,t}^{-1}$. Furthermore, the performance of spectral based linear method depends on the choice of the spectral filter function, which must be carefully designed to satisfy the following conditions that guarantee desirable properties.

\begin{definition}\label{filter}
Let $C_x$ denote the upper bound of $\|x_t\|_2$. We say that $g_\lambda:\left[0, C_x^2\right] \rightarrow \mathbb{R}$, with $0<\lambda \leq C_x^2$, is a filter function with qualification $\nu_g \geqslant 1/2$ if there exists a positive constant $b$ independent of $\lambda$ such that $\sup _{0<\sigma \leq C_x^2}\left|g_\lambda(\sigma)\right| \leq \frac{b}{\lambda}$, $\sup _{0<\sigma \leq C_x^2}\left|g_\lambda(\sigma) \sigma\right| \leq b$, and $\sup _{0<\sigma \leq C_x^2}\left|1-g_\lambda(\sigma) \sigma\right| \sigma^\nu \leq \gamma_\nu \lambda^\nu \text{ for } \forall 0<\nu \leq \nu_g$,
where $\gamma_\nu>0$ is a constant depending only on $\nu $.
\end{definition}

First, the spectral based linear method penalizes low-variance directions that are more sensitive to noise, thereby improving the robustness of the estimation. Specifically, the regularization parameter $\lambda_t$ serves to reduce the influence of small eigenvalues. For instance, under classic Tikhonov regularization, the spectral adjustment is given by $(\widehat{\Sigma}_{D,t} + \lambda_t I)^{-1}$. In this case, for directions corresponding to small eigenvalues $\sigma_{t,j}\ (j=1,\ldots,d)$ of the empirical covariance matrix $\widehat{\Sigma}_{D,t}$, the expression $\frac{1}{\sigma_{t,j} + \lambda_t}$ becomes smaller as $\lambda_t$ increases, thereby reducing their influence in the parameter estimate. 

Second, the spectral based linear approach can address the saturation phenomenon \citep{gerfo2008spectral, yao2007early}, a limitation inherent in Tikhonov regularization. The saturation phenomenon refers to the situation where, as prior information increases, the rate of performance improvement gradually decelerates. Specifically, as indicated by the constants of different spectral based linear methods in Table \ref{sb examples}, Tikhonov regularization is restricted to a maximum qualification $\nu_g$ of 1, while spectral cut-off and gradient descent can attain arbitrarily large values of $\nu_g$. This highlights the saturation effect in Tikhonov regularization and demonstrates how spectral cut-off and gradient descent methods effectively overcome this limitation.

\subsection{Algorithm design: adaptive spectral based linear Q-learning}

Considering that regularization parameters $\lambda_t$ play a key role in the spectral based linear method, this subsection aims to explore adaptive data-driven strategies for selecting these parameters. As a foundation for adaptive parameter selection, we first analyze the parameter estimation error and formalize its decomposition by introducing two auxiliary estimators. Because $
    \widehat{\Sigma}_{D, t}\theta_t^*
    =\widehat{E}_D[X_tX_t^\top \theta_t^*]
    =\widehat{E}_D[X_tE[Y_t^*\mid X_t]].
$
That is, $\widehat{\Sigma}_{D, t}\theta_t^*$ is the noise-free version of $\widehat{E}_{D}\left[X_t Y_t^*\right]$. Therefore, in addition to the solution $\theta_{D, \lambda_t, t}=g_{\lambda_t}\left(\widehat{\Sigma}_{D,t}\right) \widehat{E}_{D}[X_t Y_t]$, we define the following two estimators
$$
\hat{\theta}_{D, \lambda_t, t}:=g_{\lambda_t}\left(\widehat{\Sigma}_{D, t}\right) \widehat{E}_{D}\left[X_t Y_t^*\right], \quad \theta_{D, \lambda_t, t}^{\diamond}:=g_{\lambda_t}\left(\widehat{\Sigma}_{D, t}\right) \widehat{E}_D[X_tE[Y_t^*\mid X_t]].
$$
Then $\hat{\theta}_{D, \lambda_t, t}$ can be interpreted as the result of applying a spectral based linear estimation to the data $\left\{\left(x_{i, t}, y_{i, t}^*\right)\right\}_{i=1}^{|D|}$, while $\theta_{D, \lambda_t, t}^{\diamond}$ is a noise-free version of $\hat{\theta}_{D, \lambda_t, t}$. We now present the parameter estimation error decomposition. Applying the triangle inequality, the error is upper bounded by
\begin{equation}
\begin{aligned}
& \left\|\left(\Sigma_{t}+\lambda_t I\right)^{1 / 2}\left(\theta_{D, \lambda_t, t}-\theta_t^*\right)\right\|_{2} \leq\left\|\left(\Sigma_{ t}+\lambda_t I\right)^{1 / 2}\left(\theta_{D, \lambda_t, t}^{\diamond}-\theta_t^*\right)\right\|_{2} \\
+ & \left\|\left(\Sigma_{ t}+\lambda_t I\right)^{1 / 2}\left(\theta_{D, \lambda_t, t}^{\diamond}-\hat{\theta}_{D, \lambda_t, t}\right)\right\|_{2}+\left\|\left(\Sigma_{ t}+\lambda_t I\right)^{1 / 2}\left(\theta_{D, \lambda_t, t}-\hat{\theta}_{D, \lambda_t, t}\right)\right\|_{2} ,\label{error decomposition}
\end{aligned}
\end{equation}
where the three terms on the right-hand side of (\ref{error decomposition}) correspond to the bias, variance, and multi-stage error, respectively. Based on the theoretical analysis presented later in the appendix, we observe that the bias term increases with the regularization parameter $\lambda_t$, while the variance term decreases. The multi-stage error term is more complicated: it partly follows the variance trend and also captures the accumulation of estimation errors over time. Consequently, an overall trade-off exists between bias and variance, with some components increasing and others decreasing as $\lambda_t$ varies. This trade-off motivates the development of an adaptive approach for selecting $\lambda_t$. Specifically, for the regularization parameter $\lambda_t$, denote $
K_{D,q,t}:=\log_q\left(\frac{C_{sa}}{q_t\sqrt{|D|_\gamma}}\right)
$ with $C_{sa}:=\frac{21C_x(1+2C_x)(\sqrt{C_0}+1)}{\tilde{c}}\log\frac{2}{\delta}\ (0<\delta<1)$, we choose $\lambda_{k_{t}}=q_{t} q^{k_{t}}$ ($q_{t}>0$, $0<q<1$) with $k_t=K_{D,q,t}, \ldots, 1$, define $\hat{k}_{t}$ to be the first $k_t$ satisfying
\begin{equation}
\begin{aligned}
&\left\|\left(\widehat{\Sigma}_{D,t}+\lambda_{k_t+1} I\right)^{1 / 2}\left({\theta}_{D, \lambda_{k_t+1},t}-{\theta}_{D, \lambda_{k_t},t}\right)\right\|_2\\
\ge&C_{ada}\left(84((T-t+2)M+\Phi_{t+1})(1+C_x)\mathcal{W}_{D,\lambda_{k_t+1},t}\log^2 \frac{2}{\delta}\right),\label{adas}
\end{aligned}
\end{equation}
where $C_{ada}=8b\sqrt{\frac{1-\tilde{c}}{1-2 \tilde{c}}}\sqrt{\frac{1}{1-2 \tilde{c}}}$, $M$ is the upper bound of $|R_t|$, $\Phi_{t+1}$ is the upper bound of $| \langle \theta_{D, \lambda_{\hat{k}_{t+1}}, t+1}, x_t \rangle|$, and $\mathcal{W}_{D,\lambda_{k_t+1},t}=\left(\frac{\left(1+4\left(\frac{13C_x }{\sqrt{\lambda_{k_t+1}\ell_3}}+\frac{21C_x^2 }{\lambda_{k_t+1} \ell_3}
\right)\right)\overline{\sqrt{\mathcal{N}_{\text{empirical}}(\lambda_{k_t+1})}}}{\sqrt{|D|_\gamma}}+\frac{1}{|D|_\gamma \sqrt{\lambda_{k_t+1}}} \right)$ with $\ell_3=\frac{|D|b_0}{2\left(\max\left\{1 , \log \left(b_0c_0 |D|\frac{2\sqrt{d}}{C_x}\right)\right\}\right)^{1 / \gamma_0}}$, $\overline{\sqrt{\mathcal{N}_{\text{empirical}}(\lambda_{k_t+1})}}=\max\{\sqrt{\mathcal{N}_{\text{empirical}}(\lambda_{k_t+1})} ,1\}$, $
\mathcal{N}_{\text{empirical}}(\lambda_{k_t+1})=\operatorname{Tr}\left(\widehat{\Sigma}_{D, t}\left(\widehat{\Sigma}_{D, t}+\lambda I\right)^{-1}\right) 
$, and $|D|_\gamma=\frac{|D|b_0}{2\left(\max\{1 , \log \left(c_1^* |D|\right)\}\right)^{1 / \gamma_0}}$ with $b_0>0$, $c_0\ge0$, $\gamma_0>0$, $c_1^*=c_0 b_0\max\{\frac{\sqrt{2}\max\{M+2C_x\|\theta^*\|_2,C_x\}}{2C_xM},\frac{1}{C_x}\}$. If there is no $k_t$ satisfying (\ref{adas}), define $\hat{k}_{t}=K_{D,q,t}$. 

\begin{algorithm}[h]
\caption{Adaptive Spectral Based Linear Q-Learning (SB-LinQL\_ada)}\label{alg:linsb}
\begin{algorithmic}[1]
\REQUIRE  The confidence level $0<\delta<1$, filter function $g$, dataset $D$. 
            \STATE Initialize $\theta_{D, \lambda_{T+1}, T+1}=0$ with $\lambda_{T+1}=0$; 
\FOR{$t = T, \ldots, 1$}
    \STATE Construct the outcome
   $y_{i, t}:=r_{i, t}+\max \limits_{a_{t+1} \in \mathcal{A}_{t+1}} \langle\theta_{D, \lambda_{t+1}, t+1},x_{i,t}\left(s_{i, 1: t+1}, a_{i, 1: t}, a_{t+1}\right)\rangle, i=1,\ldots,|D|$;
   \vspace{-6mm}
    \STATE Choose the regularization parameter $\lambda_{\hat{k}_{t}}$ by (\ref{adas}), and compute
    $\theta_{D, \lambda_{\hat{k}_{t}}, t}=g_{\lambda_{\hat{k}_{t}}}\left(\widehat{\Sigma}_{D,t}\right) \widehat{E}_{D}[x_t y_t]$;
\ENDFOR

\STATE Return the estimated action $\pi_{D, \vec{\lambda}_{\hat{k}}}=\left(\pi_{D, \lambda_{\hat{k}_{1}}, 1}, \ldots, \pi_{D, \lambda_{\hat{k}_{T}}, T}\right)$ satisfying
$
{\pi}_{D,\lambda_{\hat{k}_{t}},t}(x_t(s_{1: t}, a_{1: t-1})) = \arg \max\limits_{a_t\in\mathcal{A}_t} \langle x_t(s_{1: t}, a_{1: t-1}, a_t),{\theta}_{D,\lambda_{\hat{k}_t},t} \rangle, 
$
where $t = 1, \ldots, T$.
\end{algorithmic}
\end{algorithm}

Algorithm~\ref{alg:linsb} outlines the proposed Adaptive Spectral Based Linear Q-Learning (SB-LinQL\_ada). First, by employing a linear representation, the method inherently provides interpretability. Second, in line 4 of the algorithm, we implement a spectral based estimation that enhances numerical stability while mitigating the saturation phenomenon, thereby improving overall performance. Consequently, the proposed algorithm effectively balances interpretability and performance. Moreover, an adaptive parameter selection strategy is incorporated in line 4, guided by the bias–variance trade-off principle.

\section{Theoretical 
behavior}\label{sec:theory}

This section provides a comprehensive theoretical analysis of the proposed linear RL algorithm, with the key distinctions from standard linear regression outlined in Appendix~\ref{appendix: theoretical_challenges}. Following the “no free lunch” theorem \citep{gyorfi2006distribution}, no learning algorithm can achieve satisfactory generalization error bounds without certain assumptions about the data-generating process. Accordingly, we begin by outlining the assumptions about the data and the distribution.

The first assumption involves dependence on the dataset $D$. In many real-world settings, such as time series, data often exhibit diminishing correlations as the time gap increases \citep{sun2022nystrom}. This behavior is formally characterized by the mixing property, defined as follows. Let $\mathcal{C}_{\text {Lip}}$ denote the set of bounded Lipschitz functions defined over $\mathcal{X}$, and define
$
C_{L i p}(f):=\|f\|_{L i p(\mathcal{X})}:=\sup \left\{\frac{\left|f(x)-f\left(x^{\prime}\right)\right|}{\left\|x-x^{\prime}\right\|_2} \mid x, x^{\prime} \in \mathcal{X}, x \neq x^{\prime}\right\}.
$
and
$
\|f\|_{\mathcal{C}_{L i p}}:=\|f\|_{L^{\infty}(\mathcal{X})}+C_{L i p}(f) .
$
Let $\mathcal{C}_1$ be the ``semi-ball" of functions $f \in \mathcal{C}_{\text {Lip}}$ such that $C_{\text {Lip}}(f) \leq 1$. Within this framework, $\tau$-mixing is defined as follows.

\begin{definition}[$\tau$-mixing, \cite{maume2006exponential}]
For $i,j \in \mathbb{N}$, the $\tau$-mixing coefficients are defined as
$
	\tau_j=  \sup \left\{\left\|E\left( f\left(z_{i+j}\right)\mid \mathcal{M}_{i} \right)-E\left(f\left(z_{i+j}\right)\right)\right\|_\infty \mid  f \in \mathcal{C}_1\right\} ,
$ where $\mathcal{M}_i$ is the sigma algebra generated by $z_1, \ldots, z_i$.
A sequence $\left\{z_i\right\}_{i=1}^{\infty}$ is said to be $\tau$-mixing if $\lim _{j \rightarrow \infty} \tau_j=0$. Specifically, if there exist constants $b_0>0, c_0 \geq 0, \gamma_0>0$ satisfying the inequality $
\tau_j \leq c_0 \exp \left(-\left(b_0 j\right)^{\gamma_0}\right), \text{ for } \forall j \geq 1$,
then the sequence $\left\{z_i\right\}_{i=1}^{\infty}$ is referred to as geometrically $\tau$-mixing. 
\end{definition}

\begin{assumption}\label{assump:mixing}
    For any $t=1,\ldots,T$, sequences $\left\{x_{i,t},y_{i,t}\right\}_{i=1}^{|D|}$ and $\left\{x_{i,t},y_{i,t}^*\right\}_{i=1}^{|D|}$ exhibit geometrically $\tau$-mixing with mixing coefficients $\tau_j$.
\end{assumption}

Assumption~\ref{assump:mixing} is a generalization of the commonly used i.i.d. sampling assumption. In particular, the assumption reduces to the i.i.d. case when \(\tau_j = 0\) for all \(j\). Next, we introduce a standard boundedness assumption widely used in the literature \citep{wang2023kernel}.

\begin{assumption}\label{assump:bounded}
For any $t=1, \ldots, T$, there exists $C_x,M \ge 0$ such that $\|x_t\|_2\le C_x$ and $|R_t| \leq M$.
\end{assumption}

Based on (\ref{property:q}) and $\theta_{T+1}^*=0$, Assumption \ref{assump:bounded} implies that
$
\left|y_t^*\right| \leq(T-t+2) M.
$
\begin{assumption}\label{assump:target}
For any $t=1, \ldots, T$, there holds
$
\left\|\Sigma_t^{-r}\theta_t^*\right\|_2\le C,  \text { for some } r\ge 0,C>0.
$
\end{assumption}

Assumption \ref{assump:target} imposes a structural constraint on the unknown parameter vector \(\theta_t^*\) by bounding the norm \(\|\Sigma_t^{-r} \theta_t^*\|_2\). Since \(\Sigma_t^{-r}\) increases the effect of components associated with small eigenvalues, the assumption prevents \(\theta_t^*\) from having too much weight in directions that are poorly identified. These directions correspond to feature subspaces that have low variance and potentially high noise, which can cause instability in parameter estimation. Therefore, this assumption ensures that \(\theta_t^*\) lies within a well-conditioned subspace of the feature space so that learning algorithms can achieve reliable convergence. Moreover, in practical management settings, this means that model parameters do not rely heavily on unstable or noisy features, such as customer attributes with limited variability or unreliable measurements. By imposing this assumption, overfitting to unreliable feature directions can be avoided, which contributes to the development of more stable decision policies.

\begin{assumption}\label{assump:effective dimension}
For all $t$, there exists $s \in[0,1]$ such that the effective dimension $\mathcal{N}_t(\lambda)$ satisfies
$
\mathcal{N}_t(\lambda)=\operatorname{Tr}\left(\Sigma_t\left(\Sigma_t+\lambda I\right)^{-1}\right) \leq C_0 \lambda^{-s}, 
$
where $\lambda>0$ and $C_0 \geq 1$ is a constant independent of $\lambda$.
\end{assumption}

Assumption \ref{assump:effective dimension} is the effective dimension assumption, which characterizes the decay of the eigenvalues of the covariance matrix. Based on the inequality $\mathcal{N}_t(\lambda)=\operatorname{Tr}\left(\Sigma_t\left(\Sigma_t+\lambda I\right)^{-1}\right)\le\operatorname{Tr}\left(\Sigma_t\right)\frac{1}{\lambda_{\min}(\Sigma_t+\lambda I)}\le\operatorname{Tr}\left(\Sigma_t\right)\lambda^{-1}:=C_0\lambda^{-1}$, this assumption always holds when 
$s=1$.

To connect the generalization error with the parameter estimation error, we introduce the following assumption, which characterizes the conditional distribution of action selection.

\begin{assumption} \label{assump:conditional probability}
Let $\mu \geq 1$ be a constant, for $\forall a \in \mathcal{A}_t$ and $ t=1, \ldots, T $,
$
p_t\left(a \mid s_{1: t}, a_{1: t-1}\right) \geq \mu^{-1}.
$
\end{assumption}

Assumption \ref{assump:conditional probability}, a standard assumption in RL \citep{murphy2005generalization, goldberg2012q, wang2023kernel}, ensures that conditioned on prior information, each action in the finite set $\mathcal{A}_t$ is chosen with probability no less than $\mu^{-1}$.
Based on Assumption \ref{assump:conditional probability}, and Eq. (16) in \citep{goldberg2012q}, we obtain that for any parameter vector $\theta_t$, and the policy $\pi=\left(\pi_1, \ldots, \pi_T\right)$ is defined by $\pi_t\left(s_{1: t}, a_{1: t-1}\right)=\arg \max _{a_t \in \mathcal{A}_t} \langle\theta_t,x_t\left(s_{1: t}, a_{1: t-1}, a_t\right)\rangle$, the following inequality holds.
\begin{equation}
\begin{aligned}
E\left[V_1^*\left(S_1\right)-V_{\pi, 1}\left(S_1\right)\right] \leq \sum_{t=1}^T 2 \mu^{t / 2} \sqrt{E\left[\langle \theta_t-\theta_t^*,X_t\rangle^2\right]}=\sum_{t=1}^T 2 \mu^{t / 2} \left\|\theta_t-\theta_t^*\right\|_{\Sigma_t},\label{ine:comparison}
\end{aligned}
\end{equation}
where $\|z\|_A^2=z^\top A z$ is the weighted 2-norm of  $z\in\mathbb{R}^d$ with a positive definite matrix $A\in\mathbb{R}^{d\times d}$.

Equipped with the assumptions detailed above, we establish the generalization error bound for Algorithm~\ref{alg:linsb}, which quantifies the performance of the learned policy.

\begin{theorem}\label{thm:main}
Let $0 \leq \delta \leq 1/2$ satisfy $
\delta \geq 2 \exp \left\{-\frac{\sqrt{2 r+s}}{(\log d)^{\frac{1}{\gamma_0}}\sqrt{\log_q(|D|_\gamma^{-1/2})}}|D|_\gamma^{\frac{r}{4 r+2 s+1}} \right\} $.
Under Assumptions \ref{assump:mixing}-\ref{assump:conditional probability}, with $r \geq 0$ and $0\leq s \leq 1$, if $\lambda_{\hat{k}_{t}}$ is chosen by (\ref{adas}) for $t=1, \ldots, T$, then with probability at least $1-\delta$,
\begin{equation*}
\begin{aligned}
&E\left[V_1^*\left(S_1\right)-V_{\pi_{D, \vec{\lambda}_{\hat{k}}}, 1}\left(S_1\right)\right]  \notag\\
\leq &C(T,\mu) |D|_\gamma^{-\frac{r+1/2}{2r+s+1}} \log_q\left(|D|_\gamma^{-1/2}\right)(\log d)^{\frac{2}{\gamma_0}}\log^2 \frac{2}{\delta} \left(1+\left(\log \frac{2}{\delta}\right)^{\min\{1,r\}}\mathbb{I}_{r>1/2}\right),\notag
\end{aligned}
\end{equation*}
where $C(T,\mu)=\sum_{t=1}^T  \mu^{\frac{t}{2}} C_{c}\sum_{\ell=t}^T\left((T-\ell+2)M+  M\prod_{k=\ell+1}^{T-1}\left(T-k+3\right) -M\right)$ with $C_{c}$ a constant.
\end{theorem}

Theorem~\ref{thm:main} provides the generalization error bound for the proposed linear RL method with adaptively selected regularization parameters (Algorithm~\ref{alg:linsb}). The result shows that the generalization error of the learned policy decreases as the effective sample size $|D|_\gamma$ increases. Although the regularization parameters $\lambda_{\hat{k}_t}$ are chosen adaptively, the resulting error bound remains close to the rate achieved with optimally tuned $\lambda_t=|D|_\gamma^{-\frac{1}{2 r+s+1}}$, differing only by logarithmic factors. This demonstrates that the proposed method achieves a satisfactory generalization error bound in an adaptive manner, making it well-suited for practical applications. 

The generalization error bound represents a meaningful improvement over those reported in previous studies on linear Q-learning \citep{murphy2005generalization,oh2022generalization}. In particular, our generalization error bound of $|D|_\gamma^{-(r+1/2)/(2r+s+1)}$ is sharper than the bounds commonly obtained in earlier work, which are often limited to $|D|^{-1/4}$ and rely on more restrictive assumptions. Specifically, when  $r$ is sufficiently large, the error bound can reach $|D|^{-1/2}$, illustrating the statistical efficiency of the proposed method in favorable settings. From a practical standpoint, a smaller generalization error bound enables more accurate value estimation, which contributes to better decision outcomes.

\begin{remark}
    A tighter generalization error bound can be obtained under a margin-type condition. Specifically, the comparison inequality improves from (\ref{ine:comparison}) to $E\left[V_1^*\left(S_1\right)-V_{\pi, 1}\left(S_1\right)\right] \leq \sum_{t=1}^T 2 \mu^{t / 2} \left\|\theta_t-\theta_t^*\right\|_{\Sigma_t}^{(2+2\alpha)/(2+\alpha)} $ for some $\alpha \ge 0$, which subsequently leads to a sharper generalization error bound of order $|D|_\gamma^{-\frac{(2r+1)(1+\alpha)}{(2r+s+1)(2+\alpha)}}$. Please refer to the appendix for a detailed specification of the margin condition and the full derivation of the generalization error bound.
\end{remark}

\section{Experiments}\label{sec:experiments}

To evaluate the effectiveness and interpretability of the proposed algorithm, we conduct experiments on both synthetic and real-world datasets. First, we evaluate our algorithm in a synthetic environment with fully specified ground-truth parameters, allowing precise assessment of parameter estimation, policy performance, and interpretability by directly comparing true and learned feature weights. Second, we evaluate our algorithm on recommendation data to assess its effectiveness in complex scenarios with dynamic interactions. In addition to policy performance metrics, we evaluate the interpretability of the learned policy, a crucial factor for practical deployment. 

The comparison algorithms are listed as follows. LS \citep{murphy2005generalization} denotes linear Q-learning estimated via least squares, and LASSO \citep{oh2022generalization} applies linear Q-learning with Lasso. KRR \citep{wang2023kernel} denotes kernel Q-learning estimated via kernel ridge regression, and KRR+SHAP combines KRR with the post-hoc explanation method SHAP. DNN \citep{lin2023lifting} implements deep Q-learning using a fully connected feedforward network with sigmoid activation and Xavier initialization \citep{glorot2010understanding}, with DNN+SHAP leveraging SHAP explanations. Finally, our proposed methods, referred to as LRR, LGD, and LCO, correspond to SB-LinQL\_ada estimated through linear ridge regression, gradient descent, and spectral cut-off, respectively.

\subsection{Synthetic simulations}\label{experiment: simulation}

\textbf{Performance comparison.} We construct a synthetic environment simulating a video recommendation scenario. The environment includes 10 users and 30 candidate videos. At each time step, the state is represented by the current user and the currently displayed video, and the agent selects an action corresponding to a candidate video. The reward and the resulting next state are generated according to a linear model with time-varying parameters and additive Gaussian noise; details of the trajectory generation process are provided in Appendix~\ref{appendix: simulation}. We generate 1000 trajectories with a horizon of $T=20$, split evenly into training and test sets.

For our method, the regularization schedule follows an exponential decay scheme, given by \(\lambda_k = \lambda_0 \cdot 0.9^k\), where \(\lambda_0 = 100\) for both LRR and LGD, and \(\lambda_0 = 30\) for LCO. The universal constant \(C_{\text{ada}}\) is set to \(0.5 \times 10^{-5}\), \(1 \times 10^{-5}\), and \(1 \times 10^{-4}\) for LRR, LGD, and LCO, respectively. The budget \(K_{D,q}\) is fixed at 100 for all methods. We evaluate each model using two metrics: the parameter estimation error \textit{(parameter gap)} and the policy discrepancy \textit{(policy gap)}. The parameter gap is computed as the average root mean squared error (RMSE) between the estimated parameters and the ground-truth parameters across all time steps:
$
\textit{Parameter gap} = \sqrt{\frac{1}{T} \sum_{t=1}^T \|\hat{\theta}_t - \theta^*_t\|_2^2},
$
where \(\hat{\theta}_t\) is the estimated parameter and \(\theta^*_t\) is the true parameter. The policy gap quantifies the loss in decision quality due to inaccuracies in parameter estimation. 
At each time step \(t\), we evaluate the discrepancy between the predicted outcomes \(\hat{y}_t\), 
obtained using the estimated parameter \(\hat{\theta}_t\), and the ground-truth outcomes \(y_t\), 
generated using the true parameter \(\theta^*_t\). The policy gap is computed as the root mean squared error (RMSE) across all time steps:
$
\textit{Policy gap} = \sqrt{\frac{1}{T} \sum_{t=1}^T (\hat{y}_t - y_t)^2},
$
where \(\hat{y}_t\) is the estimated outcome under the learned policy and \(y_t\) is the corresponding true outcome.

\begin{figure}[htbp]\vspace{-5mm}
	\centering
	\subfigure
	{
		\begin{minipage}[b]{0.38\linewidth}
			\includegraphics[width=1\linewidth]{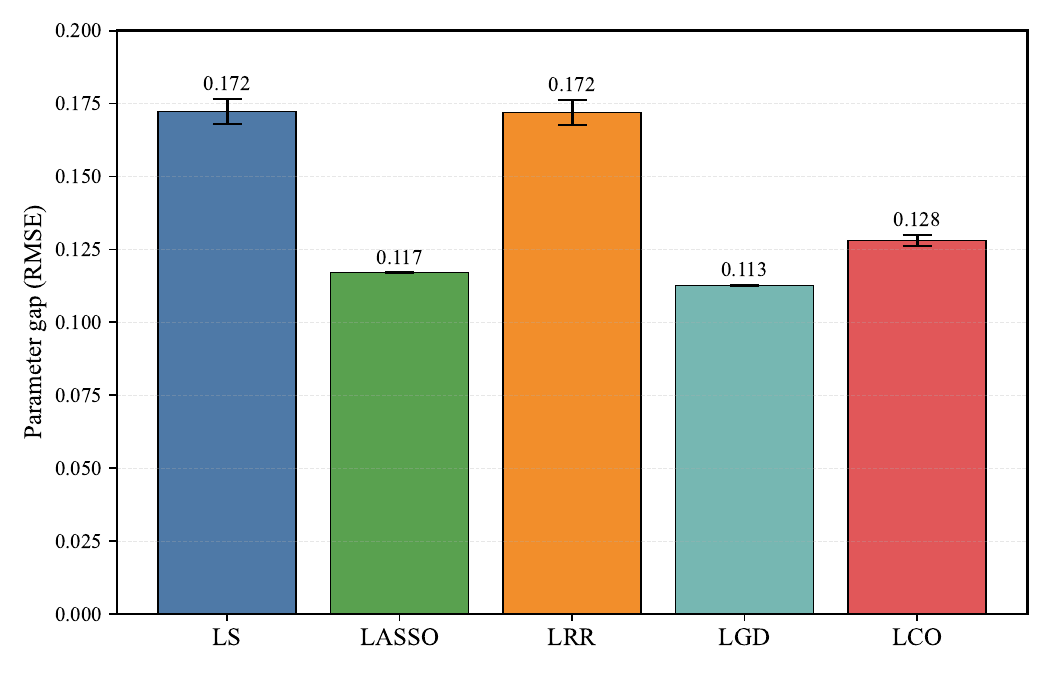}\vspace{-5pt}
		\end{minipage}
	}
	\subfigure
	{
		\begin{minipage}[b]{0.38\linewidth}
			\includegraphics[width=1\linewidth]{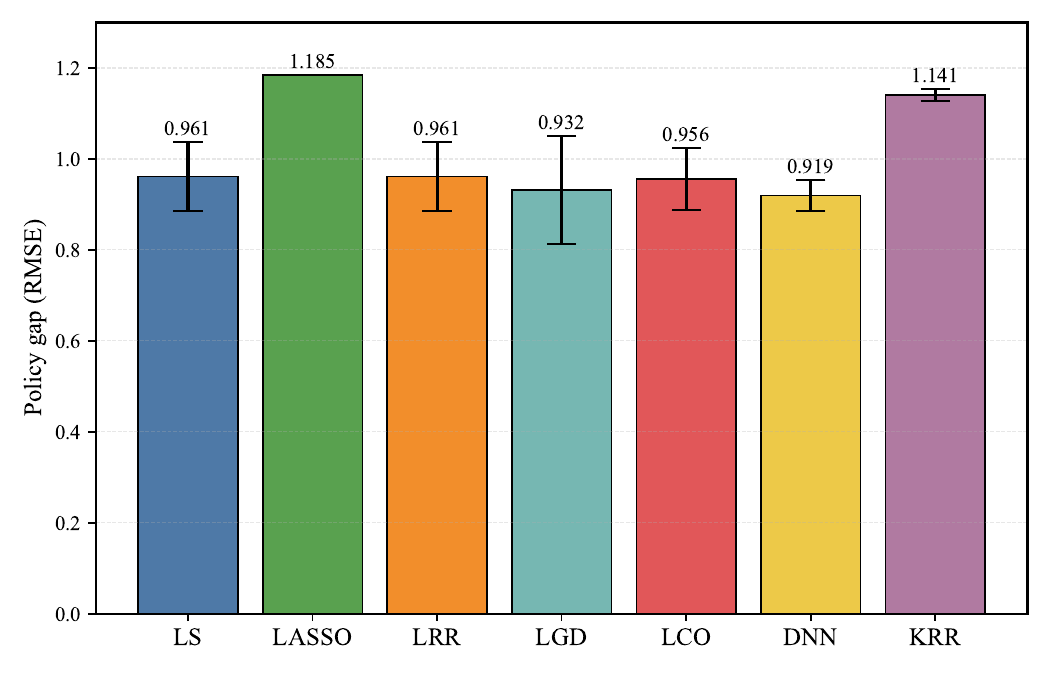}\vspace{-5pt}
		\end{minipage}
	}\hspace{-5mm}
	\caption{Parameter gap and policy gap on simulation data}\label{fig: sim performance}
\end{figure}

Each experiment is repeated five times, and the average results are presented in Fig.~\ref{fig: sim performance}. Regarding parameter estimation error (parameter gap), our methods, particularly LGD and LCO, achieve significantly lower RMSE compared to LS, LRR, and LASSO, with LGD showing the highest overall accuracy. Note that parameter gap is not reported for KRR and DNN, as they are nonparametric models. In terms of policy performance (policy gap), DNN achieves the smallest gap, outperforming all other methods, followed closely by LGD and LCO, both of which consistently outperform LASSO and LS. While DNN offers the best policy accuracy, it comes with substantially higher computational costs (see Table \ref{tab:train_time} in Appendix \ref{appendix: time}), reflecting a clear trade-off between decision quality and training efficiency. The comparatively modest performance of kernel methods can be attributed to the fact that our synthetic data were generated using a linear model. Moreover, LS completes training the fastest but suffers from the largest parameter gap and weak decision making performance. These results highlight spectral based estimation methods improve policy performance, while linear models allow for fast training and efficient computation. Additionally, our methods incorporate adaptive regularization schedules that remove the need for manual hyperparameter tuning, thereby improving scalability in large-scale sequential decision making problems.

\textbf{Interpretability analysis.} Similarly, we construct a controlled synthetic video recommendation environment in which each trajectory spans six time steps, and the ground-truth feature weights are fixed and identical across user, video, and action features. The details of the trajectory generation process are provided in Appendix~\ref{appendix: simulation 2}. We benchmark our algorithm against interpretable RL methods, including LS, LASSO, KRR+SHAP and DNN+SHAP.

\begin{figure}[htp]\vspace{-4mm}
    \centering
    \includegraphics[width=0.55\linewidth]{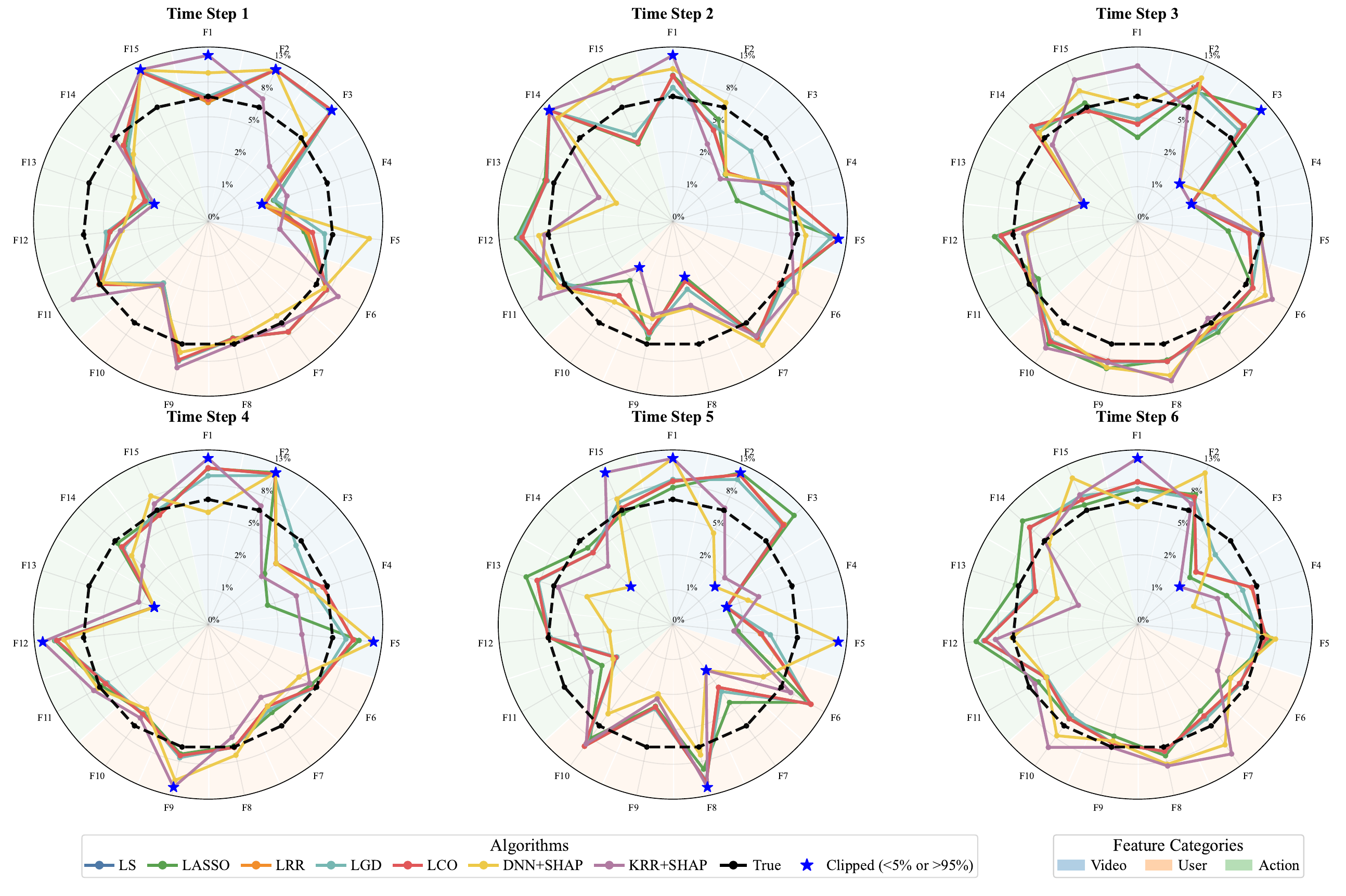}
    \caption{Visualization of feature weights across different time steps on synthetic data}
    \label{fig:radar}
\end{figure}

Fig. \ref{fig:radar} presents the ground-truth feature weights (black dashed lines) alongside the estimated feature weights produced by the different algorithms over time. Curves that remain close to the dashed lines indicate accurate interpretability, whereas large deviations imply distortion of feature relevance. To further quantify this, we mark clipped feature weights (blue stars), which are defined as values falling into the top or bottom 5\% across all algorithms, features, and time steps. The results show that our algorithm consistently aligns with the ground-truth weights across all six steps, avoiding both over- and under-emphasis on individual features. By contrast, KRR+SHAP produces the largest number of extreme weights (18 cases), followed by DNN+SHAP (12) and LASSO (11). Overall, our method demonstrates more stable and faithful interpretability than existing approaches.

\subsection{Real-world evaluation}
This section presents empirical validation on two real-world datasets: Kuaishou video recommendation and Taobao ad recommendation. The constants and hyperparameters of our adaptive regularization framework remain consistent with those used in prior simulation studies, with no additional tuning applied to these datasets. This consistency underscores the practical effectiveness and fully adaptive nature of the proposed approach, making it especially well-suited for large-scale real-world systems where manual tuning is often impractical and computationally costly. 

We evaluate the proposed approach from two perspectives: algorithmic performance and model interpretability. While performance reflects the quality of decision making, interpretability helps clarify how the learned policy makes decisions and offers practical insights for real-world applications. To assess interpretability, we conduct an analysis using LCO as a representative example. The analysis focuses on two aspects: (1) the visualization of feature weights across different time steps, and (2) the evaluation of cumulative rewards associated with the top-ranked feature vectors. Each experiment is repeated five times, and the average results are reported to ensure reliability.

\subsubsection{Case study on the Kuaishou dataset}\label{experiment: kuaishou}

The experimental evaluation is conducted using KuaiRand-1K, a large-scale sequential recommendation dataset compiled from real-world user interaction logs of the Kuaishou video platform \citep{gao2022kuairand}. This dataset comprises 11.7 million interactions involving 1,000 users and approximately 4.37 million videos. It provides fine-grained, time-stamped feedback, making it well-suited for modeling sequential decision making scenarios in which item exposure and user responses evolve dynamically over sessions. Each interaction records rich behavioral signals such as clicks, likes, and long views, along with comprehensive contextual features for both users and items. A detailed description can be found in Appendix \ref{appendix:kuaishou}.

\begin{figure}[htbp]\vspace{-5mm}
	\centering
	\subfigure[KuaiRand-1K dataset]
	{
		\begin{minipage}[b]{0.38\linewidth}
			\includegraphics[width=1\linewidth]{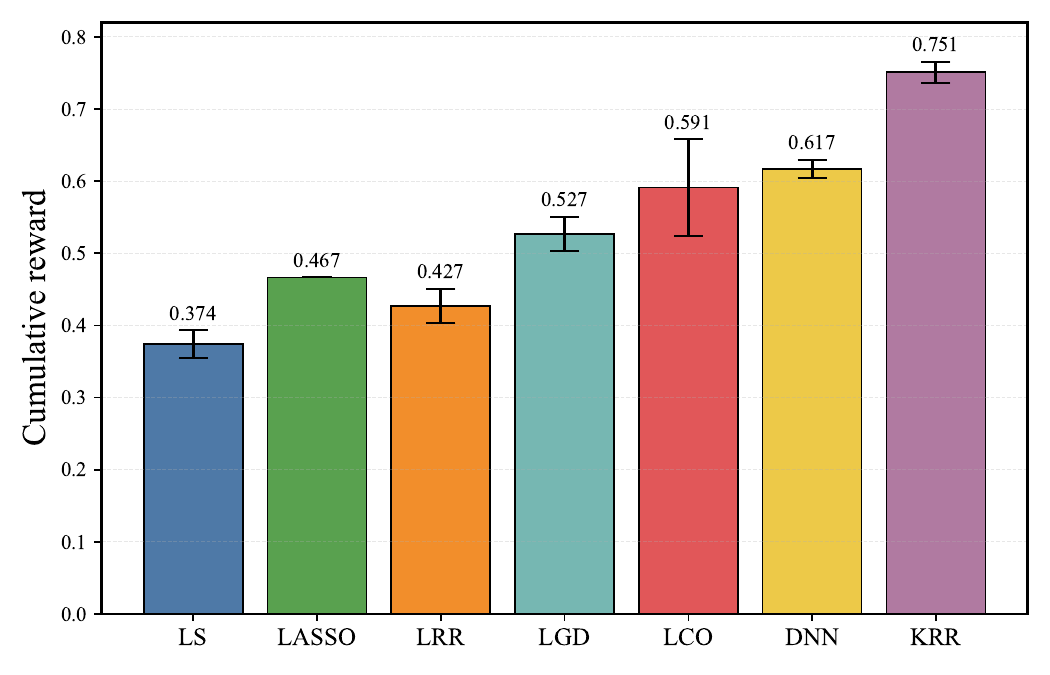}\vspace{-5pt}
		\end{minipage}
	}
	\subfigure[Ali\_Display\_Ad\_Click dataset]
	{
		\begin{minipage}[b]{0.38\linewidth}
			\includegraphics[width=1\linewidth]{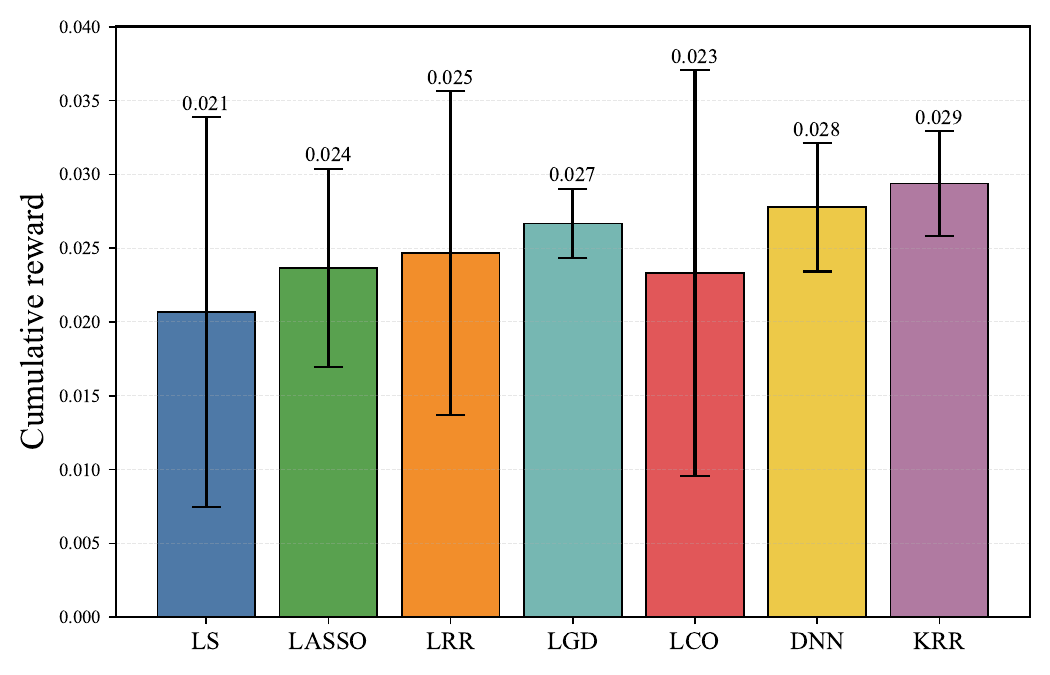}\vspace{-5pt}
		\end{minipage}
	}\hspace{-5mm}
	\caption{Cumulative reward comparison}\label{fig: real performance}
\end{figure}

\textbf{Performance comparison.} The experimental results shown in Fig.~\ref{fig: real performance} (a) demonstrate that LRR consistently outperforms LS, confirming the effectiveness of adaptive regularization in improving algorithmic performance. Furthermore, the LGD and LCO variants achieve even greater improvements over LRR, demonstrating that incorporating gradient descent optimization and spectral cutoff techniques yields significant gains. Although KRR and DNN provide higher accuracy, their lengthy training durations (see Table \ref{tab:train_time} in Appendix \ref{appendix: time}) limit their feasibility in real-world applications. These findings highlight the crucial role of spectral based algorithms in effectively capturing the complex and dynamic patterns in user-video interactions. Overall, this evidence highlights the practical value of our proposed algorithm for real-world sequential recommendation tasks.

\begin{figure}[htbp]\vspace{-4mm}
	\centering
	\subfigure[Visualization of feature weights across different time steps]
	{
		\begin{minipage}[b]{0.5\linewidth}
			\includegraphics[width=1\linewidth]{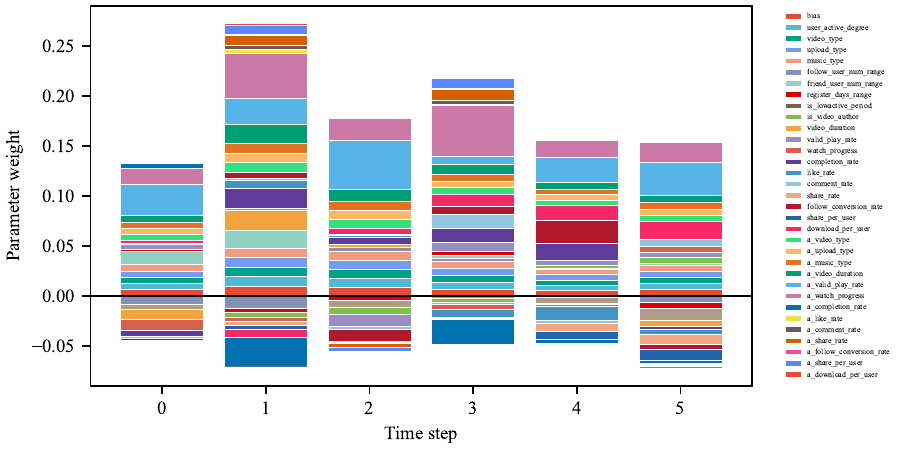}\vspace{-5pt}
		\end{minipage}
	}
	\subfigure[Cumulative reward of top-ranked feature vectors]
	{
		\begin{minipage}[b]{0.4\linewidth}
			\includegraphics[width=1\linewidth]{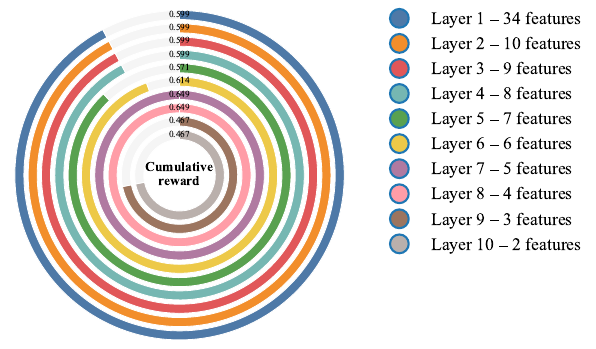}\vspace{-5pt}
		\end{minipage}
	}\hspace{-5mm}
	\caption{Interpretablity analysis on the Taobao dataset}\label{fig: interpretablity analysis}
\end{figure}

\textbf{Interpretability analysis and managerial implications.} We begin by visualizing the feature weights across different time steps. A total of 34 features are considered, encompassing both user- and video-related information. To assess feature importance, we compute the values of the learned linear coefficients at each time step and calculate each feature’s contribution proportion over the entire sequence. Fig.~\ref{fig: interpretablity analysis} (a) presents the top 10 feature classes ranked by their contribution proportion. The top three features, which are \texttt{user\_active\_degree} (11.42\%), \texttt{music\_type} (10.37\%), and \texttt{upload\_type} (9.32\%), have a significant influence on the learned policy. These observations highlight important factors influencing the policy’s decisions. The \texttt{user\_active\_degree} reflects the level of user engagement, which helps distinguish between passive and active users, enabling more personalized recommendations. The \texttt{music\_type} and \texttt{upload\_type} features indicate content genre and source, both related to user preferences and platform content diversification strategies. We next analyze the cumulative reward of the top-ranked features. As shown in Fig.~\ref{fig: interpretablity analysis} (b), the reward using all 34 features is 0.599. Surprisingly, the same reward is achieved when using only the top 4 features. Performance improves to 0.649 when using the top 5 or the top 4 features, then decreases when fewer than 4 are included or when more lower-ranked features are added. This counterintuitive result, which shows that using fewer features leads to better performance, suggests that a small subset of features captures the majority of decision-relevant information. 

\subsubsection{Case study on the Taobao dataset.}\label{experiment: taobao}
We further evaluate our method on the Ali\_Display\_Ad\_Click dataset\endnote{See \url{https://tianchi.aliyun.com/dataset/56\#1}.}, a large-scale real-world dataset released by Alibaba for display advertising on the Taobao platform. This dataset comprises user interaction logs collected from online ad impressions and feedback, and is designed for click-through rate (CTR) prediction. It includes detailed user responses along with comprehensive features extracted from both user profiles and ad metadata, making it a representative benchmark for evaluating batch RL in online advertising scenarios. A complete description of the experimental setup is provided in Appendix~\ref{appendix: taobao}.

\textbf{Performance comparison.} The experimental results shown in Fig.~\ref{fig: real performance} (b) indicate that LGD achieves strong performance, achieving accuracy comparable to that of KRR and DNN, while outperforming LASSO and LS. This demonstrates the effectiveness of our adaptive regularization method in capturing user-ad interaction patterns. Although KRR and DNN achieve slightly higher accuracy, their long training times (see Table \ref{tab:train_time} in Appendix \ref{appendix: time}) significantly limit their practicality in large-scale applications. In contrast, LGD offers a more efficient solution with comparable performance, highlighting the benefits of combining linear methods with spectral based estimation. These findings emphasize the practical value of our method in balancing accuracy and efficiency.

\begin{figure}[htbp]\vspace{-5mm}
	\centering
	\subfigure[Visualization of feature weights across different time steps]
	{
		\begin{minipage}[b]{0.52\linewidth}
			\includegraphics[width=1\linewidth]{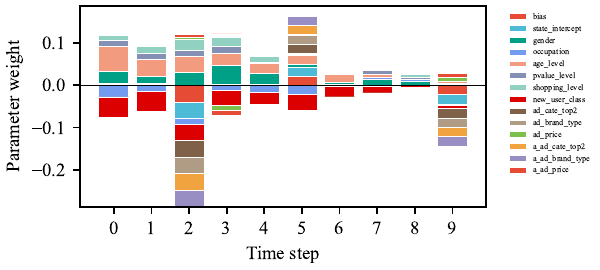}\vspace{-5pt}
		\end{minipage}
	}
	\subfigure[Cumulative reward of top-ranked feature vectors]
	{
		\begin{minipage}[b]{0.38\linewidth}
			\includegraphics[width=1\linewidth]{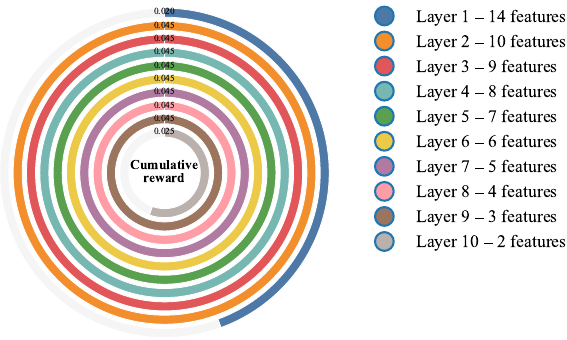}\vspace{-5pt}
		\end{minipage}
	}\hspace{-5mm}
	\caption{Interpretablity analysis on the Ali\_Display\_Ad\_Click dataset}\label{fig: interpretablity analysis taobao}
\end{figure}

\textbf{Interpretability analysis and managerial implications. } The analysis of feature weights in Fig.~\ref{fig: interpretablity analysis taobao} (a) reveals that user demographic and behavioral attributes such as city tier (\texttt{new\_user\_class}), age group (\texttt{age\_level}), and gender are among the most influential features, which underscores the importance of user profiling in personalized advertising. Moreover, action-related features such as product category and brand also contribute significantly, which indicates that effective ad targeting requires joint modeling of both user states and advertisement attributes. We next analyze the cumulative rewards associated with the top-ranked features. As shown in Fig. \ref{fig: interpretablity analysis taobao} (b), using all 14 features yields a cumulative reward of 0.02. When only the top 3 features are used, the reward increases to 0.045, whereas using only the top 2 features results in a reward of 0.025. This result suggests that retaining only the most informative features can lead to better performance.

From a managerial perspective, this has two important implications:

\begin{itemize}
    \item \textbf{Model Simplification and Deployment Efficiency:} In large-scale systems such as Kuaishou and Taobao platforms, the principle that “less is more” often proves effective, as using fewer but more informative features can yield comparable or even improved performance. Reducing the number of input features without compromising model accuracy can significantly lower computational costs and simplify the overall model structure. This reduction not only accelerates model training and decision making, but also enhances the system’s ability to serve a large number of users in real time. Furthermore, simpler models are typically easier to maintain, update, and deploy, which becomes particularly important in dynamic environments that require frequent iteration and timely adaptation. Therefore, selecting a compact and informative subset of features can lead to more efficient system deployment while maintaining reliable performance.
    \item \textbf{Feature Selection and Content Optimization:} Using the Kuaishou video recommendation scenario as an illustrative example, highlighting the most influential features, which include \texttt{user\_active\_degree}, \texttt{music\_type}, \texttt{upload\_type}, and \texttt{a\_valid\_play\_rate}, can guide content strategy. By focusing on these key features, the platform can more effectively monitor user preferences and enhance the overall user experience. For instance, encouraging greater diversity in upload types or adapting recommendations based on user activity levels can improve user satisfaction and promote long-term engagement. Such targeted efforts ensure that resources are directed toward the most influential factors, thereby supporting the platform’s sustainable growth.
\end{itemize}

In summary, our interpretability analysis not only explains the learned policy’s decision structure but also yields actionable guidance for system design and practical deployment of RL models.

\section{Conclusion and future work}\label{sec:conclusion}

This work proposes an adaptive spectral based linear RL framework for sequential decision making, aiming to prioritize interpretability while achieving competitive performance. Specifically, we develop a spectral based linear RL method that enhances numerical stability while mitigating the saturation phenomenon. Building on this framework, we propose an adaptive approach to select regularization parameters, guided by the bias–variance trade-off. Based on the relationship between batch Q-learning and multi-stage regression, we develop a novel error decomposition that incorporates a multi-stage error concept. This decomposition further supports the theoretical analysis, yielding near-optimal error bounds for parameter estimation and generalization. Experimental results on both simulated and real-world datasets from Kuaishou and Taobao indicate that our method outperforms existing baselines in decision quality. Interpretability analyses further show that the learned policies are transparent and trustworthy in practice. 

Future work will extend this framework to distributed settings, enabling scalable learning across decentralized data sources and constrained computational environments.

\begingroup \parindent 0pt \parskip 0.0ex \def\enotesize{\normalsize} \theendnotes \endgroup

%
%
%

\ACKNOWLEDGMENT{}



\bibliographystyle{informs2014} 
\bibliography{sample} 


\newpage
\begin{APPENDICES}
\section{Additional experimental details}

\subsection{Introduction to the performance comparison setting in synthetic simulations}\label{appendix: simulation}

This section details the procedure for generating trajectory data used in the performance comparison in Section~\ref{experiment: simulation}. We construct a synthetic environment that simulates a video recommendation scenario. The environment includes a pool of 10 users, where each user is represented by a fixed feature vector of dimension \(d_2 = 20\), sampled from a standard Gaussian distribution. The action space consists of 30 candidate videos, with each action associated with two types of feature vectors: an action feature vector of dimension \(d_3 = 24\), and a video feature vector of dimension \(d_1 = 28\), which determines the content of the video shown and forms part of the next state. We generate a total of 1000 trajectories, each with a time horizon of \(T = 20\). The dataset is split into training and test sets, each containing 50\% of the trajectories.

At the beginning of each trajectory, a video is randomly selected from the video pool to initialize the state. At each time step \(t\), the environment state \(s_t\) is constructed by concatenating the feature vector of the current user with that of the currently displayed video, resulting in a state vector of dimension \(d_1 + d_2\). Given the state \(s_t\), the agent selects an action according to the policy. The input vector \(x_t\) is then formed by appending the feature vector of the selected action to \(s_t\), yielding a combined input of dimension \(d = d_1 + d_2 + d_3 = 72\), which is subsequently normalized to unit norm. The observed $y_t$ is generated according to a linear model with time-varying parameters:
\[
y_t = r_t + \langle x_t, \theta^*_{t+1} \rangle + \varepsilon_t,
\]
where \(r_t\) is sampled from the uniform distribution \(\mathcal{U}(-0.5, 0.5)\), and \(\varepsilon_t \sim \mathcal{N}(0, \sigma^2)\) represents Gaussian noise with standard deviation \(\sigma = 0.5\). The parameter vector \(\theta^*_{t+1}\) is generated by concatenating two sub-vectors: the first half is drawn from \(\mathcal{N}(1, 0.2^2)\), and the second half from \(\mathcal{N}(-1, 0.2^2)\). The resulting vector is normalized to have unit \(\ell_2\)-norm to ensure consistency across time steps. After the reward is observed, the environment transitions to the next state by updating the video feature component of the state to match the video feature associated with the selected action. This process is repeated until the trajectory reaches its final time step.

\subsection{Introduction to the interpretability analysis setting in synthetic simulations}\label{appendix: simulation 2}

This section details the procedure for generating trajectory data used in the interpretability analysis in Section~\ref{experiment: simulation}. The generation of the context feature vector $x_t$ and reward $y_t$ follows the same approach as described in Appendix~\ref{appendix: simulation}. The main differences are that the dimensions of the video content, user profile, and action features are each set to 5, and the true feature weight parameter is identical across all features and time steps, reflecting a static yet unknown user preference. We generate a total of 1000 trajectories, each with a time horizon of $T = 6$.

\subsection{Introduction to the Kuaishou case study}\label{appendix:kuaishou}

This section discusses the application context and experimental configuration of the Kuaishou dataset in Section~\ref{experiment: kuaishou}. We conduct our experiments on the video browsing scenario (tab = 1) from the KuaiRand-1K dataset, which corresponds to the most typical recommendation setting on the Kuaishou platform, where the user interface is organized as a single column and videos are played in full-screen mode automatically. An illustration of this scenario is provided in Fig.~\ref{fig: video browsing scenario}. This scenario provides rich user feedback signals and a clearly defined sequential structure, making it particularly suitable for evaluating the interpretability of linear RL algorithms. Based on interaction timestamps, we segment each user's daily viewing history into trajectories, each consisting of $T=6$ consecutively watched videos, where each video corresponds to one time step in the sequence. We then identify the top 200 most popular videos over a four-week period (each with at least 30 user interactions) to define the action space and extract 619 training trajectories whose actions fall entirely within this set. Detailed descriptions of the state, action, and reward are provided below.
\begin{figure}[htp]
    \centering
    \includegraphics[width=0.6\linewidth]{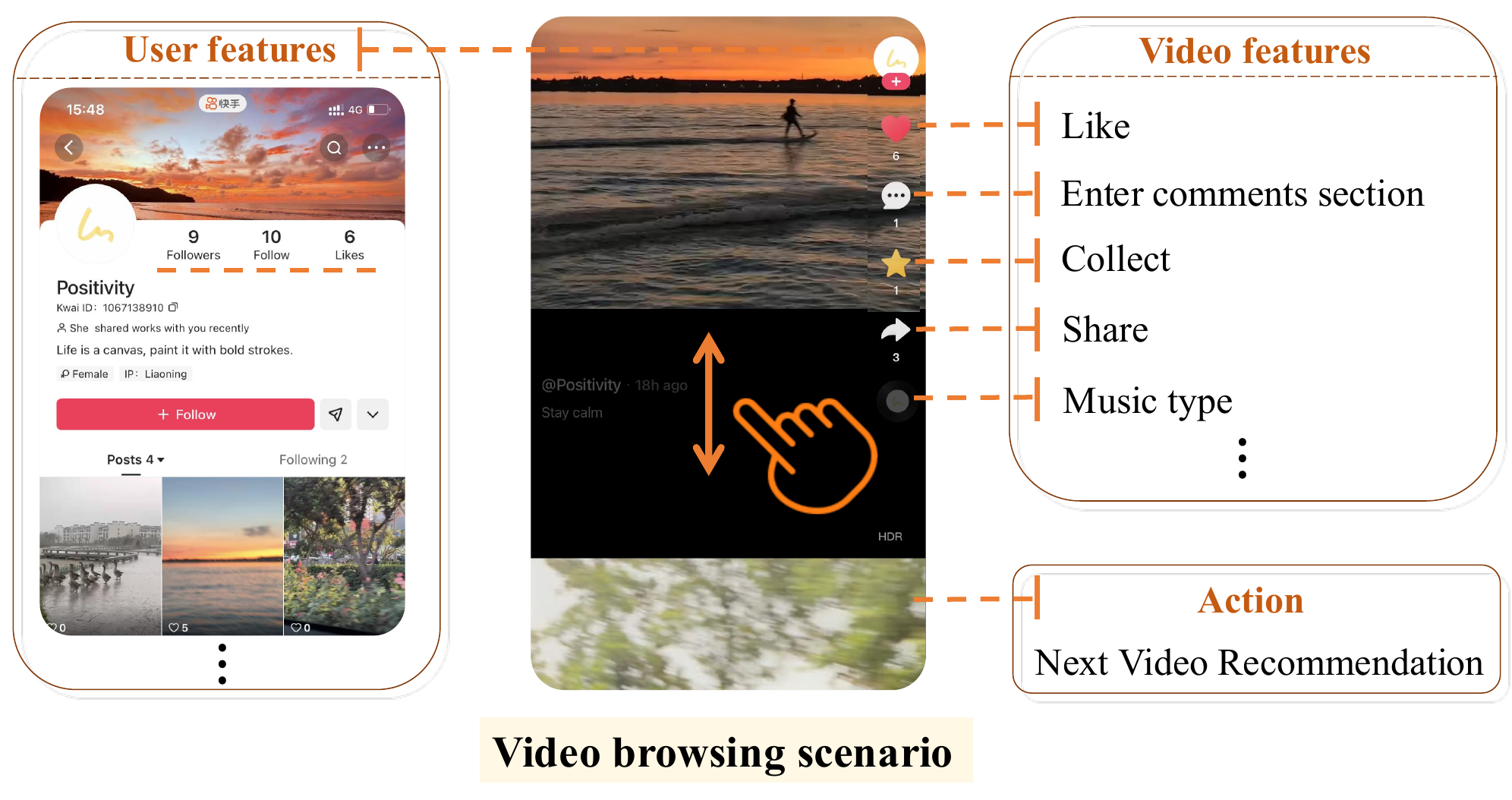}
    \caption{Illustration of the video browsing scenario}
    \label{fig: video browsing scenario}
\end{figure}

\begin{table}[htp]
\centering
\caption{Summary of state and action features on the KuaiRand-1K dataset}
\footnotesize
\begin{tabular}{lll}
\toprule
Feature Type & Feature Name & Description \\
\midrule
\multirow{7}{*}{State features\_User} 
  & user\_active\_degree & User activity level \\
  & is\_lowactive\_period & Low activity indicator (0/1) \\
  & is\_live\_streamer & Whether user is a live streamer (0/1) \\
  & is\_video\_author & Whether user is a video author (0/1) \\
  & follow\_user\_num\_range & Levels of followed users count ($0, 1, \ldots,7$) \\
  & friend\_user\_num\_range & Levels of friends count($0, 1, \ldots,6$)\\
  & register\_days\_range & Levels of registration day ($0, 1, \ldots,6$) \\
\midrule
\multirow{13}{*}{\makecell{State features\_Video\\(Action features)}} 
& (a)\_video\_type & Whether the video is an advertisement \\
& (a)\_upload\_type & Video type: three major categories, “other”  \\
  & (a)\_video\_duration & Video duration \\
  & (a)\_watch\_progress & Watch progress ratio (0--1) \\
  & (a)\_completion\_rate & Completion ratio (0--1) \\
  & (a)\_like\_rate & Like ratio (0--1) \\
  & (a)\_comment\_rate & Comment ratio (0--1) \\
  & (a)\_share\_rate & Share ratio (0--1) \\
  & (a)\_follow\_conversion\_rate & Follow conversion rate (0--1) \\
  & (a)\_share\_per\_user & Average number of shares per user \\
   & (a)\_download\_per\_user & Average number of downloads per user \\
  & (a)\_music\_type & Music type: two major categories, “other” \\
  & (a)\_valid\_play\_rate & Valid playback rate (0-Low, 1-Medium, 2-High) \\
\bottomrule
\end{tabular}
\label{tab:features}
\end{table}

The input feature vector at time $t$ is defined as $x_t = (s_t; a_t)$, where the state $s_t$ is formed by concatenating the feature vector of the currently viewed video with the user's profile features at time $t$, and the action $a_t$ corresponds to the feature vector of the video recommended at the next time step. The user profile captures behavioral characteristics, while video features include content descriptors and engagement metrics. Categorical variables are one-hot encoded, and numerical features are normalized to ensure consistency across dimensions. Table~\ref{tab:features} summarizes the complete list of features used to construct $x_t$. The transition from state $s_t$ to $s_{t+1}$ is determined by the action $a_t$, which specifies the recommended video category and thus influences the user feedback that leads to the next state. The reward $r_t$ is computed as a weighted sum of user feedback signals, where positive signals such as \textit{is\_click}, \textit{long\_view}, \textit{is\_like}, \textit{is\_comment}, \textit{is\_follow}, and \textit{is\_forward} each contribute one point, and the negative signal \textit{is\_hate} subtracts one point. This results in a reward value within the range from $-1$ to 6, representing overall user satisfaction.

\subsection{Introduction to the Taobao case study} \label{appendix: taobao}
This section provides additional details on the experimental design of the Taobao case study (Section~\ref{experiment: taobao}), with the ad recommendation scenario illustrated in Fig.~\ref{fig: ad recommendation scenario}. The dataset comprises over 26 million samples, with associated user demographic and behavioral features (e.g., age, gender, occupation), as well as ad-level features such as category, brand, and price. Each impression is associated with a scalar reward: $+1$ for a click and $-1$ for a non-click, followed by min-max normalization to the $[0,1]$ range. To simulate sequential decision making in an ad recommendation setting, each user’s behavior history is segmented into trajectories of length $T = 10$. The action space is defined by selecting the top 200 most popular ads in the dataset. Each time step $t$ in the sequence corresponds to an ad exposure event, with the input feature vector defined as $x_t = (s_t; a_t)$, where $s_t$ encodes the user profile and the most recent ad viewed, and $a_t$ represents the features of the ad to be recommended next. Table~\ref{tab:features taobao} summarizes the features of $x_t$.

\begin{figure}[htp]
    \centering
    \includegraphics[width=0.6\linewidth]{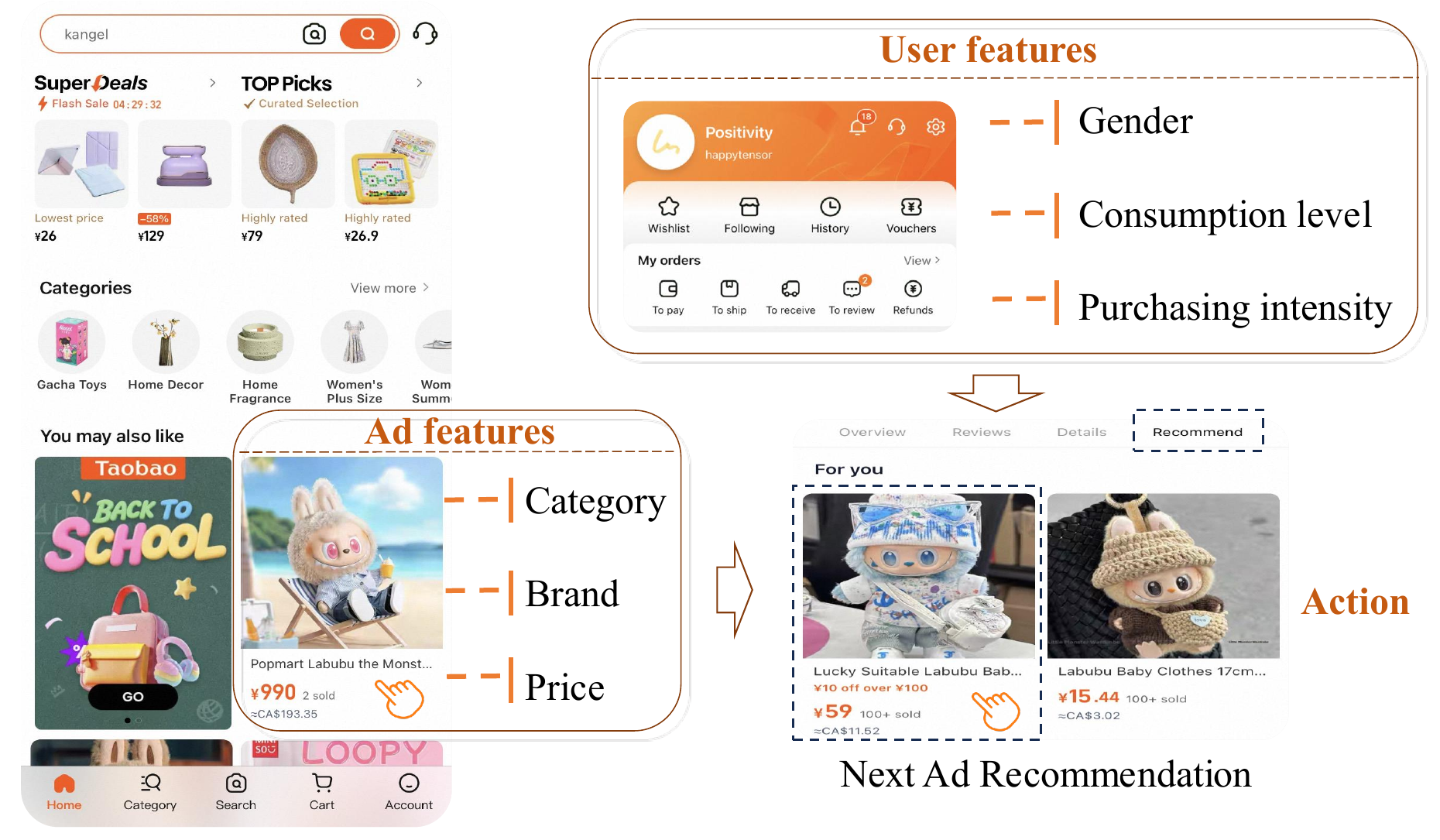}
    \caption{Illustration of the ad recommendation scenario}
    \label{fig: ad recommendation scenario}
\end{figure}

\begin{table}[htp]
\centering
\caption{Summary of state and action features on the Ali\_Display\_Ad\_Click dataset}
\footnotesize
\begin{tabular}{lll}
\toprule
Feature Type & Feature Name & Description \\
\midrule
\multirow{6}{*}{State features\_User} 
  & final\_gender\_code & Gender (1-Male, 2-Female) \\
  & occupation & University student status (1-Yes, 0-No) \\
  & age\_level & Age group \\
  & pvalue\_level & Consumption level (1-Low, 2-Medium, 3-High) \\
  & shopping\_level & User engagement level (1-Light user, 2-Moderate user, 3-Heavy user) \\
  & new\_user\_class\_level & City tier\\
\midrule
\multirow{3}{*}{\makecell{State features\_Ad\\(Action features)}} 
& (a)\_cate\_id & Product category ID \\
& (a)\_brand & Product brand  \\
  & (a)\_price & Product price \\
\bottomrule
\end{tabular}
\label{tab:features taobao}
\end{table}

\subsection{Training time} \label{appendix: time}
The average training times for all methods on synthetic and real-world datasets are summarized in Table~\ref{tab:train_time}. This detailed comparison highlights the computational efficiency of each method and provides insight into their practical applicability across different experimental settings.
\begin{table}[htbp]
\centering
\footnotesize
\caption{Average training time on simulation and real data}
\label{tab:train_time}
\begin{tabular}{cccccccc}
\toprule
Dataset & DNN & KRR & LS & LASSO & LRR & LGD & LCO \\
\midrule
Synthetic data & 1011.73 & 739.20 & 2.94 & 4.17 & 3.30 & 11.55 & 11.67 \\
KuaiRand-1K       & 207.92  & 162.43 & 6.64 & 5.59 & 5.36 & 9.73  & 5.84  \\
Ali\_Display\_Ad\_Click &  635.71 & 716.75 & 5.71 & 6.61 &5.83  & 10.7  &  6.06 \\
\bottomrule
\end{tabular}
\end{table}
\section{Theoretical challenges}\label{appendix: theoretical_challenges}

This section describes the theoretical challenges distinguishing linear RL from linear regression. As Fig.~\ref{fig:ql} illustrates, performing error analysis in linear RL is significantly more challenging. The key aspects are summarized as follows.

\begin{figure}[htp]
    \centering
    \includegraphics[width=\linewidth]{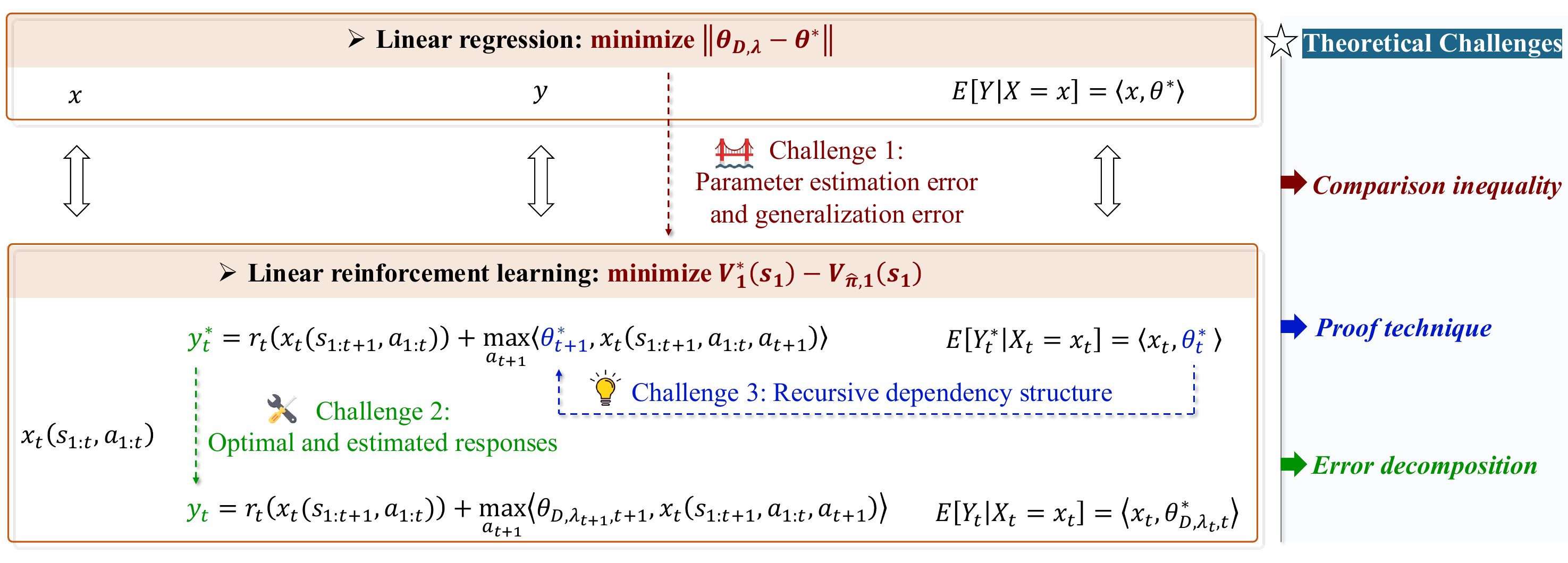}
    \caption{Theoretical challenges of linear reinforcement learning}
    \label{fig:ql}
\end{figure}

\textbf{Challenge 1: Bridging parameter estimation and generalization errors.} While linear regression aims to minimize parameter estimation error, linear RL instead targets the generalization error $V_1^*\left(s_1\right)-V_{\hat{\pi}, 1}\left(s_1\right)$. A central theoretical challenge lies in establishing a comparison inequality that connects the generalization error with the parameter estimation error, thereby enabling the derivation of generalization error bounds from the estimation error bounds.

\textbf{Challenge 2: Handling optimal and estimated responses.} Unlike linear regression, where the parameter $\theta_t^*$ directly characterizes the relationship between $x_t$ and the observed response $y_t$, the linear RL framework is more intricate. Here, the response $y_t^*$ depends on the optimal but inaccessible parameter $\theta_{t+1}^*$. In practice, this is replaced by the estimated response $y_t$ and the parameter $\theta_{D,\lambda_t,t}$ inferred from data. This substitution from the optimal response $y_t^*$ to its estimated counterpart $y_t$ complicates the analysis, particularly in decomposing parameter estimation errors.

\textbf{Challenge 3: Analyzing recursive dependency structures.} Linear RL differs fundamentally from linear regression because of the recursive dependency in its response. Specifically, the response $y_t^*$ defined in (\ref{ytstar}) depends on the next-step parameter $\theta_{t+1}^*$, whereas the empirical response $y_{i,t}$ in (\ref{prop:y bound}) is determined by the next-step estimate $\theta_{D,\lambda_{t+1},t+1}$. This recursive structure requires analytical techniques like recursive arguments to derive theoretical guarantees.

\section{Proofs}
This section presents the proofs of the generalization error bounds established in the paper. We begin with the adaptive spectral based linear regression method and then proceed to the adaptive spectral based linear reinforcement learning method by leveraging the connection between regression and reinforcement learning. We further derive a tighter generalization error bound under a margin-type condition. 

The proof sketch of Theorem \ref{thm:main} is illustrated in Fig. \ref{fig:proof sketch}. Specifically, for the adaptive spectral based linear regression, the triangle inequality allows the parameter estimation error to be decomposed into bias and variance, which are analyzed separately to obtain the overall estimation error. Since linear RL can be reformulated as a multi-stage linear regression problem, we establish a new error decomposition that includes bias, variance, and an additional multi-stage error term. While the analyses of bias and variance follow the regression setting, the analysis of the multi-stage error requires more technical arguments. Building on these results, we derive an iterative relationship among the parameter estimation errors and, through a recursive method, obtain the overall estimation error. Finally, by applying comparison inequality, we establish the generalization error bound, and under a margin-type condition, we further obtain a tighter comparison inequality together with the corresponding refined generalization error bound.

\begin{figure}[htp]
    \centering
    \includegraphics[width=\linewidth]{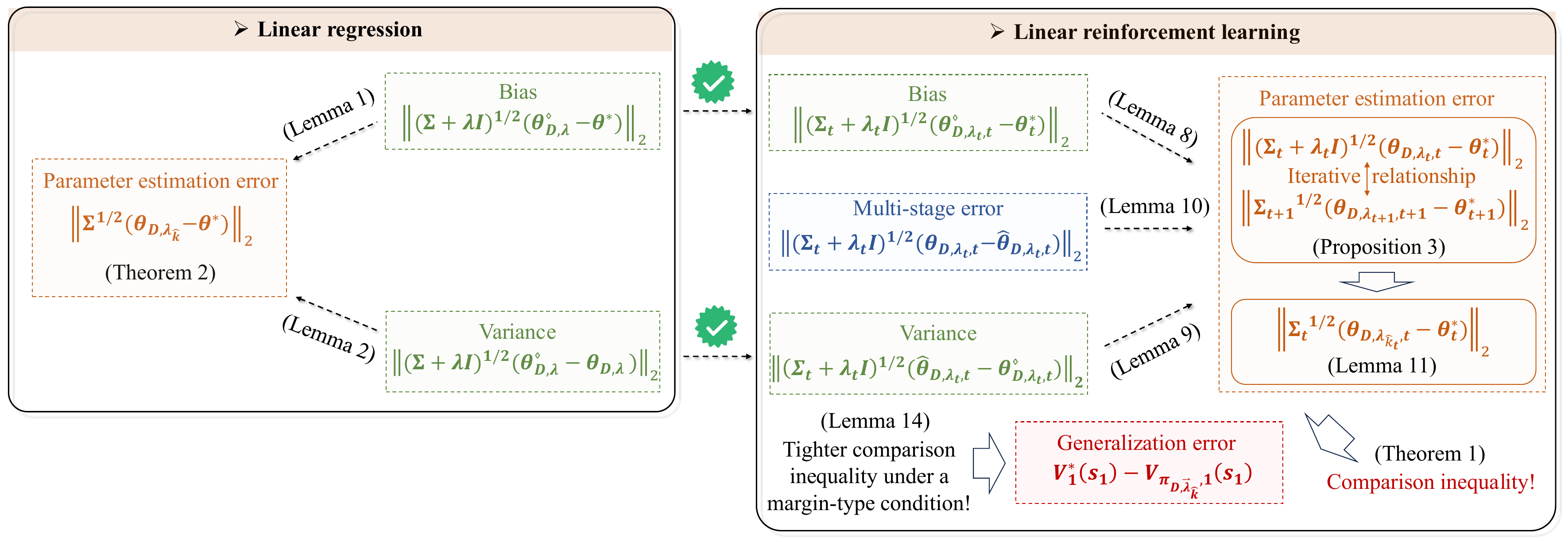}
    \caption{Proof sketch of Theorem \ref{thm:main}}
    \label{fig:proof sketch}
\end{figure}

\subsection{Proofs for adaptive spectral based linear regression method}

\subsubsection{Model formulation}
Let $ X \in \mathbb{R}^d $ and $ Y \in \mathbb{R} $ denote the random input feature vector and response variable, respectively. Given a dataset $ D = \{(x_i, y_i)\}_{i=1}^{|D|} $ containing $|D|$ identically distributed realizations sampled from the joint distribution $ P(X,Y) $, we consider the linear regression model with the following conditional expectation:
$$
    E[Y \mid X = x] = x^\top \theta^*,
$$
where $\theta^* \in \mathbb{R}^d$ represents the true parameter vector to be estimated. The covariance matrix and the empirical covariance matrix are defined as $
\Sigma = E\left[XX^{\top}\right]$ and $\widehat{\Sigma}_D = \widehat{E}_D\left[XX^{\top}\right] = \frac{1}{|D|} \sum_{i=1}^{|D|} x_i x_i^{\top}$, respectively. More generally, for any measurable function $f$, we define the empirical expectation operator:
$\widehat{E}_D[f] := \frac{1}{|D|} \sum_{i=1}^{|D|} f(x_i, y_i)$. 

We first outline the key assumptions required for the subsequent analysis.
\begin{assumption}\label{assump:mixingregression}
    The sequence $\left\{x_{i},y_{i}\right\}_{i=1}^{|D|}$ exhibits geometrically $\tau$-mixing with mixing coefficients $\tau_j$.
\end{assumption}
\begin{assumption}\label{assump:boundedregression}
There exists $C_x,M \ge 0$ such that $\|x\|_2\le C_x$ and $|y| \leq M$.
\end{assumption}

\begin{assumption}\label{assump:targetregression}
 For some $ r,C>0$, there holds $
\left\|\Sigma^{-r}\theta^*\right\|_2\le C.
$
\end{assumption}

\begin{assumption}\label{assump:effective dimensionregression}
There exists $s \in[0,1]$ such that the effective dimension $\mathcal{N}(\lambda)$ satisfies
$$
\mathcal{N}(\lambda)=\operatorname{Tr}\left(\Sigma\left(\Sigma+\lambda I\right)^{-1}\right) \leq C_0 \lambda^{-s},
$$
where $C_0 \geq 1$ is a constant independent of $\lambda$.
\end{assumption}

We adopt spectral based linear methods, where the estimator is defined as follows:
$$
 \theta_{D,\lambda}:= g_{\lambda}\left(\widehat{\Sigma}_D\right) \widehat{E}_D[XY].
$$

For the regularization parameter $\lambda$, denote $
K_{D,q}:=\log_q\left(\frac{C_{sa}}{q_0\sqrt{|D|_\gamma}}\right)
$ with $C_{sa}:=\frac{21C_x(1+2C_x)(\sqrt{C_0}+1)}{\tilde{c}}\log\frac{2}{\delta} (0<\delta<1)$, we choose $\lambda_{k}=q_0 q^{k}$ ($q_0>0$, $0<q<1$) with $k=K_{D,q}, \ldots, 1$, define $\hat{k}$ to be the first $k$ satisfying
\begin{equation}
\begin{aligned}
\left\|\left(\widehat{\Sigma}_{D}+\lambda_{k+1} I\right)^{1 / 2}\left({\theta}_{D, \lambda_{k+1}}-{\theta}_{D, \lambda_{k}}\right)\right\|_2
\ge168b\sqrt{\frac{1-\tilde{c}}{1-2 \tilde{c}}}\sqrt{\frac{1}{1-2 \tilde{c}}}M(1+C_x) \mathcal{W}_{D, \lambda_{k+1}}\log^2 \frac{2}{\delta},\label{adasregression}
\end{aligned}
\end{equation}
where $\mathcal{W}_{D, \lambda_{k+1}}$ is defined below in (\ref{notation:Wregression}). If there is no $k$ satisfying the above inequality, define $\hat{k}=K_{D,q}$. Therefore, for $k \geq \hat{k}$, there holds
$$
\left\|\left(\widehat{\Sigma}_{D}+\lambda_{k+1} I\right)^{1 / 2}\left({\theta}_{D, \lambda_{k+1}}-{\theta}_{D, \lambda_{k}}\right)\right\|_2
<168b\sqrt{\frac{1-\tilde{c}}{1-2 \tilde{c}}}\sqrt{\frac{1}{1-2 \tilde{c}}}M(1+C_x) \mathcal{W}_{D, \lambda_{k+1}}\log^2 \frac{2}{\delta} .
$$

\subsubsection{Key lemmas} 
To analyze error decomposition in parameter estimation, we introduce an auxiliary term defined as follows:
$$
\theta_{D, \lambda}^{\diamond}:=g_{\lambda}\left(\widehat{\Sigma}_{D}\right) \widehat{E}_D[XE[Y\mid X]].
$$
We now present the decomposition of the parameter estimation error. Specifically, by applying the triangle inequality, the error can be upper bounded as follows:
\begin{equation}
\begin{aligned}
 \left\|\left(\Sigma+\lambda I\right)^{1 / 2}\left(\theta_{D, \lambda}-\theta^*\right)\right\|_{2}  
\leq\left\|\left(\Sigma+\lambda I\right)^{1 / 2}\left(\theta_{D, \lambda}^{\diamond}-\theta^*\right)\right\|_{2}+ \left\|\left(\Sigma+\lambda I\right)^{1 / 2}\left(\theta_{D, \lambda}^{\diamond}-{\theta}_{D, \lambda}\right)\right\|_{2},\label{error decompositionregression}
\end{aligned}
\end{equation}
where the two terms on the right-hand side of (\ref{error decompositionregression}) correspond to the bias and variance, respectively. Next, we present several lemmas to support the subsequent analysis. Before this, we first introduce some notations.
\begin{equation}
\begin{aligned}
    & \mathcal{A}_{D,\lambda}=\left\|\left(\Sigma+\lambda I\right)^{1 / 2}g_{\lambda}\left(\widehat{\Sigma}_{D}\right)\left(\Sigma+\lambda I\right)^{1 / 2}\right\|,\label{notation:Aregression}
\end{aligned}
\end{equation}    
\begin{equation}
\begin{aligned}
&\mathcal{U}_{D,\lambda}=\left\|\left(\Sigma+\lambda I\right)^{-1 / 2}\left(\Sigma-\widehat{\Sigma}_{D}\right)\theta^*\right\|_{2} ,\label{notation:Uregression}
\end{aligned}
\end{equation}    
\begin{equation}
\begin{aligned}   &\mathcal{P}_{D,\lambda}=\left\|\left(\Sigma+\lambda I\right)^{-1/2}\left(\widehat{E}_D[XY]-\widehat{E}_D[X E[Y \mid X]]\right)\right\|_{2} ,\label{notation:Pregression}
\end{aligned}
\end{equation}    
\begin{equation}
\begin{aligned}
&\mathcal{W}_{D,\lambda}=\left(\frac{\left(1+4\left(\frac{13C_x }{\sqrt{\lambda \ell_3}}+\frac{21C_x^2 }{\lambda \ell_3}
\right)\right)\overline{\sqrt{\mathcal{N}_{\text{empirical}}(\lambda)}}}{\sqrt{|D|_\gamma}}+\frac{1}{|D|_\gamma \sqrt{\lambda}} \right),\label{notation:Wregression}
\end{aligned}
\end{equation}
where $\overline{\sqrt{\mathcal{N}_{\text{empirical}}(\lambda)}}:=\max\{\sqrt{\mathcal{N}_{\text{empirical}}(\lambda)} ,1\}$, $\ell_3=\frac{|D|b_0}{2\left(\max\left\{1 , \log \left(b_0c_0 |D|\frac{2\sqrt{d}}{C_x}\right)\right\}\right)^{1 / \gamma_0}}$, $|D|_\gamma:=\frac{|D|b_0}{2\left(\max\{1 , \log \left(c_1^* |D|\right)\}\right)^{1 / \gamma_0}}$ in which $c_1^*:=c_0 b_0\max\{\frac{\sqrt{2}\max\{M+2C_x\|\theta^*\|_2,C_x\}}{2C_xM},\frac{1}{C_x}\}. $
\begin{lemma}[Corollary 3.7 in \cite{blanchard2019concentration}]\label{37regression}
For a centered Hilbert-valued $\tau$-mixing sample $\left\{x_i\right\}_{i=1}^{|D|}$, assume there exist positive real constants $c, \sigma^2$ so that for all $i \in \mathbb{N}$:
$$
\begin{aligned}
\left\|x_i\right\| & \leq c, \quad \mathbb{P} \text {-almost surely; } \\
E\left[\left\|x_i\right\|^2\right] & \leq \sigma^2 .
\end{aligned}
$$ 
Then for any $0 \leq \delta \leq 1/2$, with probability at least $1-\delta$ it holds:
$$
\left\|\frac{1}{|D|} \sum_{i=1}^{|D|} x_i\right\| \leq \left(\frac{13 \sigma}{\sqrt{\ell^{\star}}}+\frac{21 c}{\ell^{\star}}\right)\log \frac{2}{\delta},
$$
where $\ell^{\star}:=\max \left\{1 \leq \ell \leq |D| \text{ s.t. }\tau_{\left\lfloor\frac{|D|}{\ell}\right\rfloor} \leq \max\{\frac{c}{\ell} , \frac{\sigma}{\sqrt{\ell}}\}\right\} \cup\{1\}$.
\end{lemma}

\begin{lemma}\label{empirical effective dimension}
With probability at least $1-\delta$, the relationship between the effective dimension and the empirical effective dimension is established as follows.
$$
\max \left(\frac{\overline{\sqrt{\mathcal{N}(\lambda)}}}{\overline{\sqrt{\mathcal{N}_{\text{empirical}}(\lambda)}}}, \frac{\overline{\sqrt{\mathcal{N}_{\text{empirical}}(\lambda)}}}{\overline{\sqrt{\mathcal{N}(\lambda)}}}\right)
\leq 1+4\left(\frac{13C_x }{\sqrt{\lambda \ell_3}}+\frac{21C_x^2 }{\lambda \ell_3}
    \right)\log \frac{2}{\delta}.\notag
$$
\end{lemma}

\proof{Proof.} 
Note that
\begin{equation*}
\begin{aligned}
&\mathcal{N}(\lambda_t)-\mathcal{N}_{\text{empirical}}(\lambda)
=\operatorname{Tr}(\Sigma(\Sigma+\lambda I)^{-1})-\operatorname{Tr}(\widehat{\Sigma}_D(\widehat{\Sigma}_D+\lambda I)^{-1})
=\operatorname{Tr}((\Sigma+\lambda I)^{-1}\Sigma -\left(\widehat{\Sigma}_D+\lambda I\right)^{-1}\widehat{\Sigma}_D),\notag
\end{aligned}
\end{equation*}
and because
$$
(\Sigma+\lambda I)^{-1}\Sigma -\left(\widehat{\Sigma}_D+\lambda I\right)^{-1}\widehat{\Sigma}_D =(\Sigma+\lambda I)^{-1}\left(\Sigma-\widehat{\Sigma}_D\right)+(\Sigma+\lambda I)^{-1}\left(\widehat{\Sigma}_D-\Sigma\right)\left(\widehat{\Sigma}_D+\lambda I\right)^{-1} \widehat{\Sigma}_D,
$$
we can obtain that
$$
\left|\mathcal{N}(\lambda)-\mathcal{N}_{\text{empirical}}(\lambda)\right|\leq \left|\operatorname{Tr}\left((\Sigma+\lambda I)^{-1}\left(\Sigma-\widehat{\Sigma}_D\right)\right)\right|+\left|\operatorname{Tr}\left((\Sigma+\lambda I)^{-1}\left(\widehat{\Sigma}_D-\Sigma\right)\left(\widehat{\Sigma}_D+\lambda I\right)^{-1} \widehat{\Sigma}_D\right)\right|.
$$
We next bound $\left|\operatorname{Tr}\left((\Sigma+\lambda I)^{-1}\left(\Sigma-\widehat{\Sigma}_D\right)\right)\right|$ and $\left|\operatorname{Tr}\left((\Sigma+\lambda I)^{-1}\left(\widehat{\Sigma}_D-\Sigma\right)\left(\widehat{\Sigma}_D+\lambda I\right)^{-1} \widehat{\Sigma}_D\right)\right|$, respectively. Firstly, we define $$\xi_{\operatorname{Tr}}(X)=\operatorname{Tr}((\Sigma+\lambda I)^{-1}XX^{\top}).$$ 
Note that
\begin{equation*}
\begin{aligned}
    &\left|\xi_{\operatorname{Tr}}(x_1)-\xi_{\operatorname{Tr}}(x_2)\right|\notag\\
    =&\left|\operatorname{Tr}((\Sigma+\lambda I)^{-1}(x_1x_1^{\top}-x_2x_2^\top))\right|
    \le\lambda^{-1}\left|\operatorname{Tr}(x_1x_1^{\top}-x_2x_2^\top)\right|\notag\\
    \le&\lambda^{-1}\sqrt{d}\sqrt{\|x_1(x_1^\top-x_2^\top)\|_F^2+\|(x_1-x_2)x_2^\top\|_F^2+2\langle x_1(x_1^\top-x_2^\top),(x_1-x_2)x_2^\top\rangle}\notag\\
    \le&2\lambda^{-1}\sqrt{d}C_x\|x_1-x_2\|_2. \notag
\end{aligned}
\end{equation*}
That is to say, the function $\xi_{\operatorname{Tr}}(X)=\operatorname{Tr}((\Sigma+\lambda I)^{-1}XX^{\top})$ is Lipschitz with constant $2\lambda^{-1}\sqrt{d}C_x$, from which we deduce that $(\xi_{\operatorname{Tr}}(x_i))_{i \geq 1}$ is $\tau$ mixing with rate $2\lambda^{-1}\sqrt{d}C_x\tau_j$. Then according to Lemma \ref{37regression} and the fact that 
$$\left|\xi_{\operatorname{Tr}}(X)\right|\le\left\|(\Sigma+\lambda I)^{-1}\right\|_2\operatorname{Tr}(XX^{\top})\le\frac{C_x^2}{\lambda},$$ 
$${E}(\xi_{\operatorname{Tr}}(X)^2)\le\frac{C_x^2}{\lambda}\operatorname{Tr}((\Sigma+\lambda I)^{-1}\Sigma)=\frac{C_x^2\mathcal{N}(\lambda)}{\lambda},$$ 
we can obtain that with probability at least $1-\delta$,
$$
    \left|\operatorname{Tr}\left((\Sigma+\lambda I)^{-1}\left(\Sigma-\widehat{\Sigma}_D\right)\right)\right|
    \le \left(\frac{13C_x \sqrt{\mathcal{N}(\lambda)}}{\sqrt{\lambda \ell_3}}+\frac{21C_x^2 }{\lambda \ell_3}\right)\log \frac{2}{\delta}.
$$
Secondly, note that
\begin{equation*}
\begin{aligned}
    &\left|\operatorname{Tr}\left((\Sigma+\lambda I)^{-1}\left(\widehat{\Sigma}_D-\Sigma\right)\left(\widehat{\Sigma}_D+\lambda I\right)^{-1} \widehat{\Sigma}_D\right)\right|\notag\\
    \le&\left\|(\Sigma+\lambda I)^{-1}\left(\widehat{\Sigma}_D-\Sigma\right)\right\|_F\left\|\left(\widehat{\Sigma}_D+\lambda I\right)^{-1} \widehat{\Sigma}_D\right\|_F\notag\\
    \le&\left|\operatorname{Tr}(\Sigma+\lambda I)^{-1}\left(\widehat{\Sigma}_D-\Sigma\right)\right|\sqrt{\operatorname{Tr}\left(\left(\widehat{\Sigma}_D+\lambda I\right)^{-1} \widehat{\Sigma}_D\right)\left\|\left(\widehat{\Sigma}_D+\lambda I\right)^{-1} \widehat{\Sigma}_D\right\|_2}\notag\\
    \le&\left|\operatorname{Tr}(\Sigma+\lambda I)^{-1}\left(\widehat{\Sigma}_D-\Sigma\right)\right|\sqrt{\operatorname{Tr}\left(\left(\widehat{\Sigma}_D+\lambda I\right)^{-1} \widehat{\Sigma}_D\right)\left\|\left(\widehat{\Sigma}_D+\lambda I\right)^{-1} \widehat{\Sigma}_D\right\|_2}\notag\\ \le&\left(\frac{13C_x \sqrt{\mathcal{N}(\lambda)}}{\sqrt{\lambda \ell_3}}+\frac{21C_x^2 }{\lambda \ell_3}\right)\log\frac{2}{\delta}  \sqrt{\mathcal{N}_{\text{empirical}}(\lambda)}.\notag
\end{aligned}
\end{equation*}
Then
\begin{equation*}
\begin{aligned}
    &\left|\mathcal{N}(\lambda)-\mathcal{N}_{\text{empirical}}(\lambda)\right|\notag\\
    \leq& \left(\frac{13C_x \sqrt{\mathcal{N}(\lambda)}}{\sqrt{\lambda \ell_3}}+\frac{21C_x^2 }{\lambda \ell_3}\right) \log \frac{2}{\delta}
    +\left(\frac{13C_x \sqrt{\mathcal{N}(\lambda)}}{\sqrt{\lambda \ell_3}}+\frac{21C_x^2 }{\lambda \ell_3}\right)\log\frac{2}{\delta}  \sqrt{\mathcal{N}_{\text{empirical}}(\lambda)}\notag\\
    \leq&\left(\frac{13C_x }{\sqrt{\lambda \ell_3}}+\frac{21C_x^2 }{\lambda \ell_3}
    \right)\left(\sqrt{\mathcal{N}(\lambda)}+1\right)\left(\sqrt{\mathcal{N}_{\text{empirical}}(\lambda)}+1\right)\log \frac{2}{\delta}\notag\\
    \le&4\left(\frac{13C_x }{\sqrt{\lambda \ell_3}}+\frac{21C_x^2 }{\lambda \ell_3}
    \right)\overline{\sqrt{\mathcal{N}(\lambda)}}\overline{\sqrt{\mathcal{N}_{\text{empirical}}(\lambda)}}\log \frac{2}{\delta}.\notag
\end{aligned}
\end{equation*}
Then 
\begin{equation*}
\begin{aligned}
&\left|\left(\overline{\sqrt{\mathcal{N}(\lambda)}}\right)^2-\left(\overline{\sqrt{\mathcal{N}_{\text{empirical}}(\lambda)}}\right)^2\right|
\le \left|\mathcal{N}(\lambda)-\mathcal{N}_{\text{empirical}}(\lambda)\right|
\le 4\left(\frac{13C_x }{\sqrt{\lambda \ell_3}}+\frac{21C_x^2 }{\lambda \ell_3}
    \right)\overline{\sqrt{\mathcal{N}(\lambda)}}\overline{\sqrt{\mathcal{N}_{\text{empirical}}(\lambda)}}\log \frac{2}{\delta},\notag
\end{aligned}
\end{equation*}
which yields that
\begin{equation*}
\begin{aligned}
&\max \left(\frac{\overline{\sqrt{\mathcal{N}(\lambda)}}}{\overline{\sqrt{\mathcal{N}_{\text{empirical}}(\lambda)}}}, \frac{\overline{\sqrt{\mathcal{N}_{\text{empirical}}(\lambda)}}}{\overline{\sqrt{\mathcal{N}(\lambda)}}}\right) \notag\\
\leq& 1+\left|\frac{\overline{\sqrt{\mathcal{N}(\lambda)}}}{\overline{\sqrt{\mathcal{N}_{\text{empirical}}(\lambda)}}}-\frac{\overline{\sqrt{\mathcal{N}_{\text{empirical}}(\lambda)}}}{\overline{\sqrt{\mathcal{N}(\lambda)}}}\right|\notag\\
\leq& 1+4\left(\frac{13C_x }{\sqrt{\lambda \ell_3}}+\frac{21C_x^2 }{\lambda \ell_3}
    \right)\log \frac{2}{\delta}.\notag
\end{aligned}
\end{equation*}
This completes the proof.
\endproof

\begin{lemma}\label{product of gregression}
    For $0< u \leqslant \nu_g$, we have
$$
\left\|\left(g_{\lambda}\left(\widehat{\Sigma}_{D}\right) \widehat{\Sigma}_{D}-I\right)\left(\lambda I+\widehat{\Sigma}_{D}\right)^u\right\| \leq 2^u\left(b+1+\gamma_u\right) \lambda^u .
$$
\end{lemma}

\proof{Proof.}
Note that the spectral norm of a matrix $A$ is defined as $\|A\|=\max \limits_{m \neq 0} \frac{\|Am\|_2}{\|m\|_2}$. Define $\widehat{\Sigma}_D=\sum_j\beta_j^xv_j^xv_j^{x\top}$, and then for any vector $m$, 
\begin{equation*}
\begin{aligned}
&\left\|\left(g_{\lambda}\left(\widehat{\Sigma}_D\right) \widehat{\Sigma}_D-I\right)\left(\lambda I+\widehat{\Sigma}_D\right)^um\right\|_2\notag\\
=&\left\|\sum_j\left(g_{\lambda}\left(\beta_j^x\right)\beta_j^x-1\right)\left(\lambda+\beta_j^x\right)^uv_j^xv_j^{x\top}m\right\|_2\notag\\
=&\left\{\sum_j\left[\left(g_{\lambda}\left(\beta_j^x\right)\beta_j^x-1\right)\left(\lambda+\beta_j^x\right)^uv_j^{x\top}m\right]^2\right\}^{\frac{1}{2}}\notag\\
\le&\left\{\sum_j\left[\left(g_{\lambda}\left(\beta_j^x\right)\beta_j^x-1\right)2^u\left(\lambda^u+(\beta_j^x)^u\right)v_j^{x\top}m\right]^2\right\}^{\frac{1}{2}}\notag\\
\le&2^u\left(b+1+\gamma_u\right) \lambda^u \left(\sum_j\left(v_j^{x\top}m\right)^2\right)^{\frac{1}{2}}\notag\\
\le&2^u\left(b+1+\gamma_u\right) \lambda^u \|m\|_2.\notag
\end{aligned}
\end{equation*}
This concludes that $\left\|\left(g_{\lambda}\left(\widehat{\Sigma}_D\right) \widehat{\Sigma}_D-I\right)\left(\lambda I+\widehat{\Sigma}_D\right)^u\right\| \leq 2^u\left(b+1+\gamma_u\right) \lambda^u $.
\endproof

\begin{lemma}\label{lemma512regression}
Under Assumptions \ref{assump:mixingregression}-\ref{assump:effective dimensionregression}, if $\left\|\left(\Sigma+\lambda I\right)^{-1/2}\left(\Sigma-\widehat{\Sigma}_{D}\right)\left(\Sigma+\lambda I\right)^{-1/2}\right\|\leq\tilde{c}< 1/2$, then with probability at least $1-\delta$, where $0<\delta \leq 1 / 2$, there simultaneously holds
\begin{equation}
\begin{aligned}
& \left\|\widehat{\Sigma}_{D}-\Sigma \right\|_F \le 84C_x^2 \frac{1}{\sqrt{|D|_\gamma}}\log \frac{2}{\delta},\label{differenceregression}
\end{aligned}
\end{equation}    
\begin{equation}
\begin{aligned}
&\left\|\left(\Sigma+\lambda I\right)^{1 / 2}\left(\widehat{\Sigma}_{D}+\lambda I\right)^{-1 / 2}\right\|  \leq  \sqrt{\frac{1}{1-\tilde{c}}}, \label{5.61regression}
\end{aligned}
\end{equation}    
\begin{equation}
\begin{aligned}
&\left\|\left(\widehat{\Sigma}_{D}+\lambda I\right)^{1 / 2}\left(\Sigma+\lambda I\right)^{-1 / 2}\right\|  \leq  \sqrt{\frac{1-\tilde{c}}{1-2 \tilde{c}}}, \label{5.62regression}
\end{aligned}
\end{equation}    
\begin{equation}
\begin{aligned}
&\mathcal{P}_{D,\lambda}
\leq  21M(1+C_x) \mathcal{W}_{D, \lambda}\log^2 \frac{2}{\delta}. \label{5.72regression}
\end{aligned}
\end{equation}
\end{lemma}

\proof{Proof.} We now establish the results one by one.
\begin{itemize}
    \item Bound (\ref{differenceregression}): We consider the random variable $\xi(X):=XX^\top-\Sigma.$ 
\end{itemize}
Note that
\begin{equation*}
\begin{aligned}
    &\|\xi(x_1)-\xi(x_2)\|_F 
    =\|x_1x_1^\top-x_2x_2^\top\|_F\notag\\
    \le & \sqrt{\|x_1(x_1^\top-x_2^\top)\|_F^2+\|(x_1-x_2)x_2^\top\|_F^2+2\langle x_1(x_1^\top-x_2^\top),(x_1-x_2)x_2^\top\rangle}
    \le  2C_x\|x_1-x_2\|_2,\notag
\end{aligned}
\end{equation*}
thus the function $\xi(X):=XX^\top-\Sigma$ is Lipschitz with constant $2C_x$, from which we deduce that $(\xi(x_i))_{i \geq 1}$ is $\tau$ mixing with rate $2C_x \tau_j$.
Then combined ${E}\left[\xi(x)\right]=0$, 
$$
\begin{aligned}
\left\|\xi(X)\right\|_2&\le\left\|\xi(X)\right\|_F  \leq 2C_x^2, \\
{E}\left[\left\|\xi(X)\right\|_2^2\right]&\le{E}\left[\left\|\xi(X)\right\|_F^2\right]  \leq 4C_x^4,
\end{aligned}
$$
with Lemma \ref{37regression} yields that with probability at least $1-\delta$:
$$
\left\|\widehat{\Sigma}_D-\Sigma\right\|_F \leq 21 \left(\frac{2C_x^2}{\sqrt{|D|_\gamma}}+\frac{2C_x^2}{|D|_\gamma}\right)\log \frac{2}{\delta} \leq \frac{84C_x^2 }{\sqrt{|D|_\gamma}}\log \frac{2} {\delta}.
$$
This completes the proof of (\ref{differenceregression}).
\begin{itemize}
    \item Bound (\ref{5.61regression}): Since the equation $A^{-1}-B^{-1}=B^{-1}(B-A) A^{-1}$ holds for positive matrices $A$ and $B$, we obtain
\end{itemize}
\begin{equation*}
\begin{aligned}
    &\left\|\left(\Sigma+\lambda I\right)^{1 / 2}\left(\widehat{\Sigma}_D+\lambda I\right)^{-1 / 2}\right\|^2\notag\\
   =& \left\|\left(\Sigma+\lambda I\right)^{1 / 2}\left(\widehat{\Sigma}_D+\lambda I\right)^{-1}\left(\Sigma+\lambda I\right)^{1 / 2}\right\| \notag\\
   =&\left\|\left(\Sigma+\lambda I\right)^{1 / 2}\left(\left(\widehat{\Sigma}_D+\lambda I\right)^{-1}-\left(\Sigma+\lambda I\right)^{-1}\right)\left(\Sigma+\lambda I\right)^{1 / 2}+I\right\|\notag\\
   =&\left\|\left(\Sigma+\lambda I\right)^{-1 / 2}\left(\Sigma-\widehat{\Sigma}_D\right)\left(\widehat{\Sigma}_D+\lambda I\right)^{-1}\left(\Sigma+\lambda I\right)^{1 / 2}+I \right\|\notag\\
   =&\left\|\left(\Sigma+\lambda I\right)^{-1 / 2}\left(\Sigma-\widehat{\Sigma}_D\right)\left(\Sigma+\lambda I\right)^{-1 / 2}\left(\Sigma+\lambda I\right)^{1 / 2}\left(\widehat{\Sigma}_D+\lambda I\right)^{-1}\left(\Sigma+\lambda I\right)^{1 / 2}+I \right\|\notag\\
   \le&1+\left\|\left(\Sigma+\lambda I\right)^{-1 / 2}\left(\Sigma-\widehat{\Sigma}_D\right)\left(\Sigma+\lambda I\right)^{-1 / 2}\right\|\left\|\left(\Sigma+\lambda I\right)^{1 / 2}\left(\widehat{\Sigma}_D+\lambda I\right)^{-1 / 2}\right\|^2,\notag
\end{aligned}
\end{equation*}
combined with the condition $\left\|\left(\Sigma+\lambda I\right)^{-\frac{1}{2}}\left(\Sigma-\widehat{\Sigma}_D\right)\left(\Sigma+\lambda I\right)^{-\frac{1}{2}}\right\|\leq\tilde{c}< 1 / 2$, we can further obtain that
$$
\left\|\left(\Sigma+\lambda I\right)^{1 / 2}\left(\widehat{\Sigma}_D+\lambda I\right)^{-1 / 2}\right\|^2\le 1+\tilde{c}\left\|\left(\Sigma+\lambda I\right)^{1 / 2}\left(\widehat{\Sigma}_D+\lambda I\right)^{-1 / 2}\right\|^2,
$$
that is to say,
$$
\left\|\left(\Sigma+\lambda I\right)^{1 / 2}\left(\widehat{\Sigma}_D+\lambda I\right)^{-1 / 2}\right\|\le\sqrt{\frac{1}{1-\tilde{c}}}.
$$
This completes the proof of (\ref{5.61regression}).
\begin{itemize}
    \item Bound (\ref{5.62regression}): Again by the equation $A^{-1}-B^{-1}=B^{-1}(B-A) A^{-1}$ holds for positive matrices $A$ and $B$, we can conclude that
\end{itemize}
\begin{equation}
\begin{aligned}
    &\left\|\left(\widehat{\Sigma}_D+\lambda I\right)^{1 / 2}\left(\Sigma+\lambda I\right)^{-1 / 2}\right\|^2\\
   =& \left\|\left(\widehat{\Sigma}_D+\lambda I\right)^{1 / 2}\left(\Sigma+\lambda I\right)^{-1}\left(\widehat{\Sigma}_D+\lambda I\right)^{1 / 2}\right\| \\
   =&\left\|\left(\widehat{\Sigma}_D+\lambda I\right)^{1 / 2}\left(\left(\Sigma+\lambda I\right)^{-1}-\left(\widehat{\Sigma}_D+\lambda I\right)^{-1}\right)\left(\widehat{\Sigma}_D+\lambda I\right)^{1 / 2}+I\right\|\\
   =&\left\|\left(\widehat{\Sigma}_D+\lambda I\right)^{-1 / 2}\left(\widehat{\Sigma}_D-\Sigma\right)\left(\Sigma+\lambda I\right)^{-1}\left(\widehat{\Sigma}_D+\lambda I\right)^{1 / 2}+I \right\|\\
   =&\left\|\left(\widehat{\Sigma}_D+\lambda I\right)^{-1 / 2}\left(\widehat{\Sigma}_D-\Sigma\right)\left(\widehat{\Sigma}_D+\lambda I\right)^{-1 / 2}\left(\widehat{\Sigma}_D+\lambda I\right)^{1 / 2}\left(\Sigma+\lambda I\right)^{-1}\left(\widehat{\Sigma}_D+\lambda I\right)^{1 / 2}+I \right\|\\
   \le&1+\left\|\left(\widehat{\Sigma}_D+\lambda I\right)^{-1 / 2}\left(\widehat{\Sigma}_D-\Sigma\right)\left(\widehat{\Sigma}_D+\lambda I\right)^{-1 / 2}\right\|\left\|\left(\widehat{\Sigma}_D+\lambda I\right)^{1 / 2}\left(\Sigma+\lambda I\right)^{-1 / 2}\right\|^2.\label{621}
\end{aligned}
\end{equation}
Based on the condition $\left\|\left(\Sigma+\lambda I\right)^{-\frac{1}{2}}\left(\Sigma-\widehat{\Sigma}_D\right)\left(\Sigma+\lambda I\right)^{-\frac{1}{2}}\right\|\leq\tilde{c}< 1 / 2$ and (\ref{5.61regression}),
\begin{equation}
\begin{aligned}
&\left\|\left(\widehat{\Sigma}_D+\lambda I\right)^{-1 / 2}\left(\widehat{\Sigma}_D-\Sigma\right)\left(\widehat{\Sigma}_D+\lambda I\right)^{-1 / 2}\right\| \\
\leq&\left\|\left(\widehat{\Sigma}_D+\lambda I\right)^{-1 / 2}\left(\Sigma+\lambda I\right)^{1 / 2}\right\|^2 \left\|\left(\Sigma+\lambda I\right)^{-\frac{1}{2}}\left(\Sigma-\widehat{\Sigma}_D\right)\left(\Sigma+\lambda I\right)^{-\frac{1}{2}}\right\|\leq \frac{\tilde{c}}{1-\tilde{c}} \label{622}.
\end{aligned}
\end{equation}
Combined (\ref{621}) with (\ref{622}) yields that
$$
\left\|\left(\widehat{\Sigma}_D+\lambda I\right)^{1 / 2}\left(\Sigma+\lambda I\right)^{-1 / 2}\right\|^2\le 1+\frac{\tilde{c}}{1-\tilde{c}} \left\|\left(\widehat{\Sigma}_D+\lambda I\right)^{1 / 2}\left(\Sigma+\lambda I\right)^{-1 / 2}\right\|^2,
$$
that is to say, 
$$
\left\|\left(\widehat{\Sigma}_D+\lambda I\right)^{1 / 2}\left(\Sigma+\lambda I\right)^{-1 / 2}\right\|\le\sqrt{\frac{1-\tilde{c}}{1-2 \tilde{c}}}.$$
This completes the proof of (\ref{5.62regression}).
\begin{itemize}
    \item Bound (\ref{5.72regression}): We consider the random variable:
$
\xi_{\mathcal{P}}(X, Y)=(\Sigma+\lambda I)^{-\frac{1}{2}}\left(XY-XX^\top\theta^*\right).
$
\end{itemize}
Note that
\begin{equation*}
\begin{aligned}
    &\|\xi_{\mathcal{P}}(x_1, y_1)-\xi_{\mathcal{P}}(x_2, y_2)\|_2 \notag\\
    =&\|(\Sigma+\lambda I)^{-\frac{1}{2}}\left(x_1 y_1-x_2y_2-(x_1x_1^\top\theta^*-x_2x_2^\top\theta^*)\right)\|_2 \notag\\
    \le& \lambda^{-\frac{1}{2}}\left(\|x_1 y_1-x_2y_2\|_2+\|x_1x_1^\top\theta^*-x_2x_2^\top\theta^*\|_2\right)  \notag\\
    =&\lambda^{-\frac{1}{2}}\left(\|x_1 y_1-x_2y_1+x_2y_1-x_2y_2\|_2+\|(x_1x_1^\top-x_1x_2^\top+x_1x_2^\top-x_2x_2^\top)\theta^*\|_2\right)  \notag\\
    \le&\lambda^{-\frac{1}{2}}\left(\|x_1 -x_2\|_2(3C_x\|\theta^*\|_2+M)+C_x|y_1-y_2|\right)  \notag\\
    \le&\sqrt{2}\lambda^{-\frac{1}{2}}\max\{3C_x\|\theta^*\|_2+M,C_x\}\sqrt{\|x_1 -x_2\|_2^2+(y_1-y_2)^2},\notag
\end{aligned}
\end{equation*}
thus the function $\xi_{\mathcal P}(X, Y)=(\Sigma+\lambda I)^{-\frac{1}{2}}\left(XY-XX^\top\theta^*\right)$ is Lipschitz with constant $\sqrt{2}\lambda^{-\frac{1}{2}}\max\{3C_x\|\theta^*\|_2+M,C_x\}$, from which we deduce that $\left(\xi_{\mathcal P}\left(x_i, y_i\right)\right)_{i \geq 1}$ is $\tau$ mixing with rate $\sqrt{2}\lambda^{-\frac{1}{2}}\max\{3C_x\|\theta^*\|_2+M,C_x\} \tau_j$.
Then combined ${E}\left[\xi_{\mathcal P}(X, Y)\right]=0$, 
$$
\left\|(\Sigma+\lambda I)^{-\frac{1}{2}}\left(X Y-XX^\top\theta^*\right)\right\|_2 \leq\left\|(\Sigma+\lambda I)^{-\frac{1}{2}}\right\|\left\|X Y-XX^\top\theta_*\right\|_2 \leq C_x M \lambda^{-\frac{1}{2}}  ,
$$
$$
\begin{aligned}
{E}\left[\left\|\xi_{\mathcal P}(X, Y)\right\|_2^2\right] & =E\left[\left( Y-X^\top\theta^*\right)^2X^\top(\Sigma+\lambda I)^{-1}X\right] \\
& \leq M^2 {E}\left[\operatorname{Tr}(X^\top(\Sigma+\lambda I)^{-1}X)\right]\\
& = M^2 {E}\left[XX^\top\operatorname{Tr}((\Sigma+\lambda I)^{-1})\right]\\
& = M^2 \operatorname{Tr}({E}\left[XX^\top\right](\Sigma+\lambda I)^{-1})\\
& =M^2\mathcal{N}(\lambda),
\end{aligned}
$$
with Lemmas \ref{37regression} and \ref{empirical effective dimension} yields that with probability at least $1-\delta$:
\begin{equation*}
\begin{aligned}
\mathcal{P}_{D,\lambda}=&\left\|\left(\Sigma+\lambda I\right)^{-1/2}\left(\widehat{E}_D[XY]-\widehat{E}_D[X E[Y \mid X]]\right)\right\|_{2}  \notag\\
\leq& 21M(1+C_x)  \left(\frac{\sqrt{\mathcal{N}(\lambda)}}{\sqrt{|D|_\gamma}}+\frac{1}{|D|_\gamma \sqrt{\lambda}} \right)\log \frac{2}{\delta}\notag\\
\leq& 21M(1+C_x)  \mathcal{W}_{D, \lambda}\log^2 \frac{2}{\delta}.\notag
\end{aligned}
\end{equation*}
This proves (\ref{5.72regression}) and finishes the proof of Lemma \ref{lemma512regression}.
\endproof

\begin{lemma}\label{lemma:Aregression}
    Under Assumptions \ref{assump:mixingregression}-\ref{assump:effective dimensionregression}, if $\left\|\left(\Sigma+\lambda I\right)^{-1/2}\left(\Sigma-\widehat{\Sigma}_{D}\right)\left(\Sigma+\lambda I\right)^{-1/2}\right\|\leq\tilde{c}< 1/2$, then with probability at least $1-\delta$, where $0<\delta \leq 1 / 2$, there holds$$\mathcal{A}_{D,\lambda}\le2b\sqrt{\frac{1}{1-\tilde{c}}}\sqrt{\frac{1-\tilde{c}}{1-2 \tilde{c}}}.$$
\end{lemma}
\proof{Proof.}
Due to Lemma \ref{lemma512regression} and Definition \ref{filter}, we have
$$
\begin{aligned}
    &\mathcal{A}_{D,\lambda}=\left\|\left(\Sigma+\lambda I\right)^{1 / 2}g_{\lambda}\left(\widehat{\Sigma}_{D}\right)\left(\Sigma+\lambda I\right)^{1 / 2}\right\|\\
    =&\left\|\left(\Sigma+\lambda I\right)^{1 / 2}\left(\widehat{\Sigma}_{D}+\lambda I\right)^{-1 / 2}g_{\lambda}\left(\widehat{\Sigma}_{D}\right)\left(\widehat{\Sigma}_{D}+\lambda I\right)\left(\widehat{\Sigma}_{D}+\lambda I\right)^{-1 / 2}\left(\Sigma+\lambda I\right)^{1 / 2}\right\|\\
    \le & 2b\sqrt{\frac{1}{1-\tilde{c}}}\sqrt{\frac{1-\tilde{c}}{1-2 \tilde{c}}}=2b\sqrt{\frac{1}{1-2 \tilde{c}}}.
\end{aligned}
$$
This completes the proof of Lemma \ref{lemma:Aregression}.
\endproof

\begin{lemma}\label{lemma:Uregression}
    Under Assumptions \ref{assump:mixingregression}-\ref{assump:effective dimensionregression}, with probability at least $1-\delta$, where $0<\delta \leq 1 / 2$, there holds$$\mathcal{U}_{D,\lambda}\le21(1+2C_x)M\mathcal{W}_{D,\lambda}\log \frac{2}{\delta}.$$
\end{lemma}
\proof{Proof.}
     We consider the random vector
$$
\xi_{\mathcal{U}}(X):=(\Sigma+\lambda I)^{-1/2}\left(XX^\top-\Sigma\right)\theta^*.
$$
Hence,
$$
\begin{aligned}
    &\|\xi_{\mathcal{U}}(x_{1})-\xi_{\mathcal{U}}(x_{2})\|_2 
    =\|(\Sigma_t+\lambda I)^{-1/2}\left(x_{1}x_{1}^\top-x_{2}x_{2}^\top\right)\theta^*\|_2\\
    \le & \lambda ^{-1/2}\left\| x_{1}(x_{1}^\top-x_{2}^\top)\theta^*+(x_{1}-x_{2})x_{2}^\top\theta^* \right\|_2
    \le \left(M+C_x\|\theta^*\|_2\right)\lambda ^{-1/2}\|x_{1}-x_{2}\|_{2},
\end{aligned}
$$
thus $\xi_{\mathcal{U}}(X):=(\Sigma+\lambda I)^{-1/2}\left(XX^\top-\Sigma\right)\theta^*$ is Lipschitz with constant $\left(M+C_x\|\theta^*\|_2\right)\lambda ^{-1/2}$, from which $(\xi_{\mathcal{U}}(x_{i}))_{i \geq 1}$ is $\tau$ mixing with rate $\left(M+C_x\|\theta^*\|_2\right)\lambda ^{-1/2} \tau_j$.
Then combined $E\left[\xi_{\mathcal{U}}(X)\right]=0$, 
$$
\begin{aligned}
&\left\|\xi_{\mathcal{U}}(X)\right\|_2 =\left\|(\Sigma+\lambda I)^{-1/2}\left(XX^\top-\Sigma\right)\theta^*\right\|_2\\
\le &\lambda ^{-1/2}\left\|XX^\top\theta^*\right\|_2+\lambda ^{-1/2}\left\|E[XX^\top\theta^*]\right\|_2
\le2C_xM\lambda ^{-1/2},
\end{aligned}
$$
and
$$
\begin{aligned}
&E\left[\left\|\xi_{\mathcal{U}}(X)\right\|_2^2\right]\\ 
=&E \left[\operatorname{Tr}\left(\left(\theta^*\right)^\top\left(XX^\top-\Sigma\right)(\Sigma+\lambda I)^{-1}\left(XX^\top-\Sigma\right)\theta^*\right) \right] \\
=&E \left[\operatorname{Tr}\left(\left(\theta^*\right)^\top XX^\top(\Sigma+\lambda I)^{-1}XX^\top\theta^*\right)\right]-\operatorname{Tr}(\left(\theta^*\right)^\top\Sigma(\Sigma+\lambda I)^{-1}\Sigma\theta^*)  \\
\leq& E \left[\operatorname{Tr}\left(\left(\theta^*\right)^\top XX^\top\theta^*\right)\operatorname{Tr}\left((\Sigma+\lambda I)^{-1}XX^\top\right)\right]
 \leq M^2 \mathcal{N}(\lambda) .
\end{aligned}
$$
with Lemma \ref{37regression} yields that with probability at least $1-\delta$:
$$
\left\|(\Sigma+\lambda I)^{-1/2}\left(\widehat{\Sigma}_{D}-\Sigma\right)\theta^*\right\|_2 
\leq  21 \left(\frac{M \sqrt{\mathcal{N}(\lambda)}}{\sqrt{|D|_\gamma}}+\frac{2 C_xM}{\sqrt{\lambda}|D|_\gamma}\right)\log \frac{2}{\delta}
\le21(1+2C_x)M\mathcal{W}_{D,\lambda}\log \frac{2}{\delta}.
$$ 
This completes the proof of Lemma \ref{lemma:Uregression}.
\endproof

\subsubsection{Proof of parameter estimation error}

\begin{lemma}\label{lemma:approximation errorregression}
Under Assumptions \ref{assump:mixingregression}-\ref{assump:effective dimensionregression}, with probability at least $1-\delta$:
$$
\left\|\left(\Sigma+\lambda I\right)^{1 / 2}\left(\theta_{D, \lambda}^{\diamond}-\theta^*\right)\right\|_2 
\leq C_{sa1}^{\prime}\left(\lambda^{\min\{1/2+r,\nu_g\}}+\lambda^{\min\{1/2,\nu_g\}}\left( \frac{1}{\sqrt{|D|_\gamma}}\log \frac{2}{\delta}\right)^{\min\{1,r\}}\mathbb{I}_{r>1/2}\right),
$$
where $C_{sa1}^{\prime}=\left(\frac{1}{1-\tilde{c}}\right)^{r+1/2}C\left(\gamma_{1/2+r} +b+1\right)\max\{1,r C_x^{2(r-1)}\} \left(84C_x^2\right)^{\min\{1,r\}}$.
\end{lemma}

\proof{Proof.}
Because 
$$
\begin{aligned}
    &\|\left(\Sigma+\lambda I\right)^{1 / 2}\left(\theta_{D, \lambda}^{\diamond}-\theta^*\right)\|_2\notag\\
    =&\left\|\left(\Sigma+\lambda I\right)^{1 / 2}\left(g_{\lambda}(\widehat{\Sigma}_{D})\widehat{E}_D[X E[Y \mid X]]-\theta^*\right)\right\|_2\notag\\
    =&\left\|\left(\Sigma+\lambda I\right)^{1 / 2}\left(g_{\lambda}(\widehat{\Sigma}_{D})\widehat{\Sigma}_{D}\theta^*-\theta^*\right)\right\|_2\notag\\
    =&\left\|\left(\Sigma+\lambda I\right)^{1/2}\left(\widehat{\Sigma}_{D}+\lambda_t I\right)^{-1/2}\left(\widehat{\Sigma}_{D}+\lambda I\right)^{1/2}\left(g_{\lambda}(\widehat{\Sigma}_{D})\widehat{\Sigma}_{D}-I\right)\theta^*\right\|_2\notag\\
    \le&\left\|\left(\Sigma+\lambda I\right)^{1/2}\left(\widehat{\Sigma}_{D}+\lambda I\right)^{-1/2}\right\|\left\|\left(\widehat{\Sigma}_{D}+\lambda I\right)^{1/2}\left(g_{\lambda}(\widehat{\Sigma}_{D})\widehat{\Sigma}_{D}-I\right)\Sigma^r\Sigma^{-r}\theta^*\right\|_2.\notag
\end{aligned}
$$
If $0 \le r \le \frac{1}{2}$, then Assumption~\ref{assump:targetregression}, together with Lemmas~\ref{product of gregression} and~\ref{lemma512regression}, implies that
$$
\begin{aligned}
    &\|\left(\Sigma+\lambda I\right)^{1 / 2}\left(\theta_{D, \lambda}^{\diamond}-\theta^*\right)\|_2\notag\\
    \le&\left\|\left(\Sigma+\lambda I\right)^{1/2}\left(\widehat{\Sigma}_{D}+\lambda I\right)^{-1/2}\right\|\left\|\left(\widehat{\Sigma}_{D}+\lambda I\right)^{1/2}\left(g_{\lambda}(\widehat{\Sigma}_{D})\widehat{\Sigma}_{D}-I\right)\left(\widehat{\Sigma}_{D}+\lambda I\right)^{r}\left(\widehat{\Sigma}_{D}+\lambda I\right)^{-r}\left(\Sigma+\lambda I\right)^{r}\Sigma^{-r}\theta^*\right\|_2\notag\\
    \le&C\left\|\left(\Sigma+\lambda I\right)^{1/2}\left(\widehat{\Sigma}_{D}+\lambda I\right)^{-1/2}\right\|^{2r+1}\left\|\left(\widehat{\Sigma}_{D}+\lambda I\right)^{1/2+r}\left(g_{\lambda}(\widehat{\Sigma}_{D})\widehat{\Sigma}_{D}-I\right)\right\|\notag\\
    \le&C\left(\frac{1}{1-\tilde{c}}\right)^{r+1/2}\left(\gamma_{1/2+r} +b+1\right)\lambda^{\min\{1/2+r,\nu_g\}}.\notag
\end{aligned}
$$
If $r>1/2$, then Assumption~\ref{assump:targetregression}, together with Lemmas~\ref{product of gregression} and~\ref{lemma512regression}, implies that
$$
\begin{aligned}
    &\|\left(\Sigma+\lambda I\right)^{1 / 2}\left(\theta_{D, \lambda}^{\diamond}-\theta^*\right)\|_2\notag\\
    \le&C\sqrt{\frac{1}{1-\tilde{c}}}\left\|\left(\widehat{\Sigma}_{D}+\lambda I\right)^{1/2}\left(g_{\lambda}(\widehat{\Sigma}_{D})\widehat{\Sigma}_{D}-I\right)\left(\Sigma^r-\widehat{\Sigma}_{D}^r+\widehat{\Sigma}_{D}^r\right)\right\|\notag\\
    \le&C\sqrt{\frac{1}{1-\tilde{c}}}\left\|\left(\widehat{\Sigma}_{D}+\lambda I\right)^{1/2+r}\left(g_{\lambda}(\widehat{\Sigma}_{D})\widehat{\Sigma}_{D}-I\right)\right\|+C\sqrt{\frac{1}{1-\tilde{c}}}\left\|\left(\widehat{\Sigma}_{D}+\lambda I\right)^{1/2}\left(g_{\lambda}(\widehat{\Sigma}_{D})\widehat{\Sigma}_{D}-I\right)\right\|\left\|\Sigma^r-\widehat{\Sigma}_{D}^r\right\|\notag\\
    \le&\left(\lambda^{\min\{1/2+r,\nu_g\}}+\lambda^{\min\{1/2,\nu_g\}}\max\{1,r C_x^{2(r-1)}\} \left(84C_x^2 \frac{1}{\sqrt{|D|_\gamma}}\log \frac{2}{\delta}\right)^{\min\{1,r\}}\right) C\sqrt{\frac{1}{1-\tilde{c}}}\left(\gamma_{1/2+r} +b+1\right).\notag
\end{aligned}
$$
The two cases can be further integrated by
\begin{equation}
\begin{aligned}
&\|\left(\Sigma+\lambda I\right)^{1 / 2}\left(\theta_{D, \lambda}^{\diamond}-\theta^*\right)\|_2 
\le C_{sa1}^{\prime}\left(\lambda^{\min\{1/2+r,\nu_g\}}+\lambda^{\min\{1/2,\nu_g\}}\left( \frac{1}{\sqrt{|D|_\gamma}}\log \frac{2}{\delta}\right)^{\min\{1,r\}}\mathbb{I}_{r>1/2}\right),\label{vr2}
\end{aligned}
\end{equation}
where $C_{sa1}^{\prime}=\left(\frac{1}{1-\tilde{c}}\right)^{r+1/2}C\left(\gamma_{1/2+r} +b+1\right)\max\{1,r C_x^{2(r-1)}\} \left(84C_x^2\right)^{\min\{1,r\}}$.
This completes the proof of Lemma \ref{lemma:approximation errorregression}.
\endproof

\begin{lemma}\label{lemma:sample errorregression}
There holds that
$$
\left\|\left(\Sigma+\lambda I\right)^{1 / 2}\left(\theta_{D, \lambda}^{\diamond}-{\theta}_{D, \lambda}\right)\right\|_{2} \leq \mathcal{A}_{D,\lambda}\mathcal{P}_{D,\lambda}   .
$$
\end{lemma}

\proof{Proof.}
By (\ref{notation:Aregression}) and (\ref{notation:Pregression})  we have
$$
\begin{aligned}
	&\left\|\left(\Sigma+\lambda I\right)^{1 / 2}\left(\theta_{D, \lambda}^{\diamond}-{\theta}_{D, \lambda}\right)\right\|_{2} \notag\\
	=&\left\|\left(\Sigma+\lambda I\right)^{1 / 2}g_{\lambda}\left(\widehat{\Sigma}_{D}\right)\left(\Sigma+\lambda I\right)^{1 / 2}\left(\Sigma+\lambda I\right)^{-1 / 2}\left(\widehat{E}_D[XY]-\widehat{E}_D[X E[Y \mid X]]\right)\right\|_{2}
	\le\mathcal{A}_{D,\lambda}\mathcal{P}_{D,\lambda} .\notag
\end{aligned}
$$
This completes the proof of Lemma \ref{lemma:sample errorregression}.
\endproof

\begin{proposition}\label{prop:error decompositionregression}
Under Assumptions \ref{assump:mixingregression}-\ref{assump:effective dimensionregression}, with probability at least $1-\delta$, we have
$$
\begin{aligned}
& \left\|\left(\Sigma+\lambda I\right)^{1 / 2}\left(\theta_{D, \lambda}-\theta^*\right)\right\|_{2}\\
\leq & C_{sa1}^{\prime}\left(\lambda^{\min\{1/2+r,\nu_g\}}+\lambda^{\min\{1/2,\nu_g\}}\left( \frac{1}{\sqrt{|D|_\gamma}}\log \frac{2}{\delta}\right)^{\min\{1,r\}}\mathbb{I}_{r>1/2}\right)
+42b\sqrt{\frac{1}{1-2 \tilde{c}}}M(1+C_x) \mathcal{W}_{D, \lambda}\log^2 \frac{2}{\delta}.\notag
\end{aligned}
$$
\end{proposition}

\proof{Proof.}
Inserting Lemmas \ref{lemma:approximation errorregression} and \ref{lemma:sample errorregression} into (\ref{error decompositionregression}), we obtain
\begin{equation}
\begin{aligned}
& \left\|\left(\Sigma+\lambda I\right)^{1 / 2}\left(\theta_{D, \lambda}-\theta^*\right)\right\|_{2} \\
\leq  &C_{sa1}^{\prime}\left(\lambda^{\min\{1/2+r,\nu_g\}}+\lambda^{\min\{1/2,\nu_g\}}\left( \frac{1}{\sqrt{|D|_\gamma}}\log \frac{2}{\delta}\right)^{\min\{1,r\}}\mathbb{I}_{r>1/2}\right)+\mathcal{A}_{D, \lambda}\mathcal{P}_{D, \lambda}.\label{mediumregression}
\end{aligned}
\end{equation}
Substituting (\ref{5.72regression}), Lemmas \ref{lemma:Aregression} and \ref{lemma:Uregression} into (\ref{mediumregression}) yields the following:
    $$
\begin{aligned}
& \left\|\left(\Sigma+\lambda I\right)^{1 / 2}\left(\theta_{D, \lambda}-\theta^*\right)\right\|_{2}\\
\leq & C_{sa1}^{\prime}\left(\lambda^{\min\{1/2+r,\nu_g\}}+\lambda^{\min\{1/2,\nu_g\}}\left( \frac{1}{\sqrt{|D|_\gamma}}\log \frac{2}{\delta}\right)^{\min\{1,r\}}\mathbb{I}_{r>1/2}\right)
+42b\sqrt{\frac{1}{1-2 \tilde{c}}}M(1+C_x) \mathcal{W}_{D, \lambda}\log^2 \frac{2}{\delta}.\notag
\end{aligned}
$$
This finishes the proof of Proposition \ref{prop:error decompositionregression}.
\endproof

\begin{proposition}\label{prop: adasgregression}
If Assumptions \ref{assump:mixingregression}-\ref{assump:effective dimensionregression} hold, then for any  $\lambda<\lambda^{\prime}$ satisfying 
$$
\begin{aligned}&\left\|\left(\Sigma+\lambda I\right)^{-1/2}\left(\Sigma-\widehat{\Sigma}_{D}\right)\left(\Sigma+\lambda I\right)^{-1/2}\right\|\leq\tilde{c}< 1 / 2,\notag\\
&\left\|\left(\Sigma+\lambda^\prime I\right)^{-1/2}\left(\Sigma-\widehat{\Sigma}_{D}\right)\left(\Sigma+\lambda^\prime I\right)^{-1/2}\right\|\leq\tilde{c}< 1 / 2,\notag
\end{aligned}
$$
with probability at least $1-\delta$, we have
$$
\begin{aligned}
& \left\|\left(\widehat{\Sigma}_{D}+\lambda I\right)^{1 / 2}\left({\theta}_{D,\lambda}-{\theta}_{D,\lambda^{\prime}}\right)\right\|_2 \notag\\
\le&\sqrt{\frac{1-\tilde{c}}{1-2 \tilde{c}}}\left[2C_{sa1}^{\prime}\left((\lambda_t^\prime)^{\min\{1/2+r,\nu_g\}}+(\lambda_t^\prime)^{\min\{1/2,\nu_g\}}\left( \frac{1}{\sqrt{|D|_\gamma}}\log \frac{2}{\delta}\right)^{\min\{1,r\}}\mathbb{I}_{r>1/2}\right)\right.\notag\\
&\left.+84b\sqrt{\frac{1}{1-2 \tilde{c}}}M(1+C_x) \mathcal{W}_{D, \lambda}\log^2 \frac{2}{\delta}\right].\notag
\end{aligned}
$$
\end{proposition}
\proof{Proof.}
By triangle inequality and (\ref{5.62regression}), we can obtain that
\begin{equation}
\begin{aligned}
& \left\|\left(\widehat{\Sigma}_{D}+\lambda I\right)^{1 / 2}\left({\theta}_{D,\lambda}-{\theta}_{D,\lambda^{\prime}}\right)\right\|_2 
\le  \sqrt{\frac{1-\tilde{c}}{1-2 \tilde{c}}}\left(\left\|\left(\Sigma+\lambda I\right)^{1 / 2}\left({\theta}_{D,\lambda}-\theta^*\right)\right\|_2+\left\|\left(\Sigma+\lambda I\right)^{1 / 2}\left({\theta}_{D,\lambda^{\prime}}-\theta^*\right)\right\|_2\right).\label{tingjiregression}
\end{aligned}
\end{equation}
It remains to bound $\left\|\left(\Sigma+\lambda I\right)^{1 / 2}\left({\theta}_{D,\lambda}-\theta^*\right)\right\|_2$ and $\left\|\left(\Sigma+\lambda I\right)^{1 / 2}\left({\theta}_{D,\lambda^{\prime}}-\theta^*\right)\right\|_2$. By Proposition \ref{prop:error decomposition}, there holds
\begin{equation}
\begin{aligned}
    &\left\|\left(\Sigma+\lambda I\right)^{1 / 2}\left({\theta}_{D,\lambda}-\theta^*\right)\right\|_2\\
    \le&C_{sa1}^{\prime}\left(\lambda^{\min\{1/2+r,\nu_g\}}+\lambda^{\min\{1/2,\nu_g\}}\left( \frac{1}{\sqrt{|D|_\gamma}}\log \frac{2}{\delta}\right)^{\min\{1,r\}}\mathbb{I}_{r>1/2}\right)
+42b\sqrt{\frac{1}{1-2 \tilde{c}}}M(1+C_x) \mathcal{W}_{D, \lambda}\log^2 \frac{2}{\delta}.\label{tingji1regression}
\end{aligned}
\end{equation}
Similarly, if $\lambda<\lambda^{\prime}$, we can obtain that
\begin{equation}
\begin{aligned}
&\left\|\left(\Sigma+\lambda I\right)^{1 / 2}\left({\theta}_{D,\lambda^{\prime}}-\theta^*\right)\right\|_2\\
\le&C_{sa1}^{\prime}\left((\lambda^\prime)^{\min\{1/2+r,\nu_g\}}+(\lambda^\prime)^{\min\{1/2,\nu_g\}}\left( \frac{1}{\sqrt{|D|_\gamma}}\log \frac{2}{\delta}\right)^{\min\{1,r\}}\mathbb{I}_{r>1/2}\right)\\
&+42b\sqrt{\frac{1}{1-2 \tilde{c}}}M(1+C_x) \mathcal{W}_{D, \lambda}\log^2 \frac{2}{\delta}.\label{tingji2regression}
\end{aligned}
\end{equation}
Substituting (\ref{tingji1regression}) and (\ref{tingji2regression}) into (\ref{tingjiregression}) yields that
$$
\begin{aligned}
& \left\|\left(\widehat{\Sigma}_{D}+\lambda I\right)^{1 / 2}\left({\theta}_{D,\lambda}-{\theta}_{D,\lambda^{\prime}}\right)\right\|_2 \notag\\
\le & \sqrt{\frac{1-\tilde{c}}{1-2 \tilde{c}}}\left(\left\|\left(\Sigma+\lambda I\right)^{1 / 2}\left({\theta}_{D,\lambda}-\theta^*\right)\right\|_2+\left\|\left(\Sigma+\lambda I\right)^{1 / 2}\left({\theta}_{D,\lambda^{\prime}}-\theta^*\right)\right\|_2\right)\notag\\
\le&\sqrt{\frac{1-\tilde{c}}{1-2 \tilde{c}}}\left[2C_{sa1}^{\prime}\left((\lambda^\prime)^{\min\{1/2+r,\nu_g\}}+(\lambda^\prime)^{\min\{1/2,\nu_g\}}\left( \frac{1}{\sqrt{|D|_\gamma}}\log \frac{2}{\delta}\right)^{\min\{1,r\}}\mathbb{I}_{r>1/2}\right)\right.\notag\\
&\left.+84b\sqrt{\frac{1}{1-2 \tilde{c}}}M(1+C_x) \mathcal{W}_{D, \lambda}\log^2 \frac{2}{\delta}\right].\notag
\end{aligned}
$$
This completes the proof of Proposition \ref{prop: adasgregression}.
\endproof
\begin{theorem}\label{thm:asregression}
Under Assumptions \ref{assump:mixingregression}-\ref{assump:effective dimensionregression}, and $\lambda_{\hat{k}}$ obtained by (\ref{adasregression}), then with probability at least $1-\delta\ (\delta \in(0,1/2))$, there holds
\begin{equation}
\begin{aligned}
&\left\|\Sigma^{1/2}\left({\theta}_{D, \lambda_{\hat{k}}}-\theta^*\right)\right\|_2
\le C_2|D|_\gamma^{-\frac{r+1/2}{2r+s+1}} (\log d)^{\frac{2}{\gamma_0}}\log^2 \frac{2}{\delta}\log_q\left(|D|_\gamma^{-1/2}\right) \left(1+\left(\log \frac{2}{\delta}\right)^{\min\{1,r\}}\mathbb{I}_{r>1/2}\right) ,\end{aligned}
\end{equation}
where $C_2$ is the constant independent of $|D|$ and $\delta$.
\end{theorem}

\proof{Proof of Theorem \ref{thm:asregression}.}
There exists a $k_{0} \in\left[1, K_{D,q}\right]$ such that $\lambda_{k_{0}}=q_{0} q^{k_{0}} \sim |D|_\gamma^{-\frac{1}{2 r+s+1}}$. If $k_{0} \leq \hat{k}$, i.e., $\lambda_{k_{0}} \geq \lambda_{\hat{k}}$, we obtain from the definition of $\hat{k}$ that
\begin{equation}\begin{aligned}
&168b\sqrt{\frac{1-\tilde{c}}{1-2 \tilde{c}}}\sqrt{\frac{1}{1-2 \tilde{c}}}M(1+C_x) \mathcal{W}_{D, \lambda_{\hat{k}+1}}\log^2 \frac{2}{\delta} 
<\left\|\left(\widehat{\Sigma}_{D}+\lambda_{\hat{k}+1} I\right)^{1 / 2}\left({\theta}_{D, \lambda_{\hat{k}+1}}-{\theta}_{D, \lambda_{\hat{k}}}\right)\right\|_2.\label{constant11regression}
\end{aligned}
\end{equation}
Due to Proposition \ref{prop: adasgregression}, it follows that
\begin{equation}
\begin{aligned}
& \left\|\left(\widehat{\Sigma}_{D}+\lambda_{\hat{k}+1} I\right)^{1 / 2}\left({\theta}_{D, \lambda_{\hat{k}+1}}-{\theta}_{D, \lambda_{\hat{k}}}\right)\right\|_2\\
\le&\sqrt{\frac{1-\tilde{c}}{1-2 \tilde{c}}}\left[2C_{sa1}^{\prime}\left((\lambda_{\hat{k}})^{\min\{1/2+r,\nu_g\}}+(\lambda_{\hat{k}})^{\min\{1/2,\nu_g\}}\left( \frac{1}{\sqrt{|D|_\gamma}}\log \frac{2}{\delta}\right)^{\min\{1,r\}}\mathbb{I}_{r>1/2}\right)\right.\\
&+\left.84b\sqrt{\frac{1}{1-2 \tilde{c}}}M(1+C_x) \mathcal{W}_{D, \lambda_{\hat{k}+1}}\log^2 \frac{2}{\delta}\right].\label{constant21regression}
\end{aligned}
\end{equation}
Combining (\ref{constant11regression}) and (\ref{constant21regression}) leads to
\begin{equation}\begin{aligned}
&42b\sqrt{\frac{1}{1-2 \tilde{c}}}M(1+C_x) \mathcal{W}_{D, \lambda_{\hat{k}+1}}\log^2 \frac{2}{\delta}\\
\leq &C_{sa1}^{\prime}\left((\lambda_{\hat{k}})^{\min\{1/2+r,\nu_g\}}+(\lambda_{\hat{k}})^{\min\{1/2,\nu_g\}}\left( \frac{1}{\sqrt{|D|_\gamma}}\log \frac{2}{\delta}\right)^{\min\{1,r\}}\mathbb{I}_{r>1/2}\right).\label{constant3regression}
\end{aligned}
\end{equation}
Therefore, we can further obtain that
$$
\begin{aligned}
& \left\|\left(\widehat{\Sigma}_{D}+\lambda_{\hat{k}} I\right)^{1 / 2}\left({\theta}_{D, \lambda_{\hat{k}}}-\theta^*\right)\right\|_2 \\
=&\left\|\left(\widehat{\Sigma}_{D}+\lambda_{\hat{k}} I\right)^{1 / 2}\left(\Sigma+\lambda_{\hat{k}} I\right)^{-1 / 2}\left(\Sigma+\lambda_{\hat{k}} I\right)^{1 / 2}\left({\theta}_{D, \lambda_{\hat{k}}}-\theta^*\right)\right\|_2\\
\le&\sqrt{\frac{1-\tilde{c}}{1-2 \tilde{c}}}\left\|\left(\Sigma+\lambda_{\hat{k}} I\right)^{1 / 2}\left({\theta}_{D, \lambda_{\hat{k}}}-\theta^*\right)\right\|_2\\
\leq & \sqrt{\frac{1-\tilde{c}}{1-2 \tilde{c}}} C_{sa1}^{\prime}\left(\lambda_{\hat{k}}^{\min\{1/2+r,\nu_g\}}+\lambda_{\hat{k}}^{\min\{1/2,\nu_g\}}\left( \frac{1}{\sqrt{|D|_\gamma}}\log \frac{2}{\delta}\right)^{\min\{1,r\}}\mathbb{I}_{r>1/2}\right)\notag\\
&+\sqrt{\frac{1-\tilde{c}}{1-2 \tilde{c}}} 42b\sqrt{\frac{1}{1-2 \tilde{c}}}M(1+C_x)\mathcal{W}_{D,\lambda_{\hat{k}},t}\log^2 \frac{2}{\delta}\notag\\
\leq & \sqrt{\frac{1-\tilde{c}}{1-2 \tilde{c}}} C_{sa1}^{\prime}\left(\lambda_{\hat{k}}^{\min\{1/2+r,\nu_g\}}+\lambda_{\hat{k}}^{\min\{1/2,\nu_g\}}\left( \frac{1}{\sqrt{|D|_\gamma}}\log \frac{2}{\delta}\right)^{\min\{1,r\}}\mathbb{I}_{r>1/2}\right)\notag\\
&+\sqrt{\frac{1-\tilde{c}}{1-2 \tilde{c}}} 42b\sqrt{\frac{1}{1-2 \tilde{c}}}M(1+C_x)\mathcal{W}_{D,\lambda_{\hat{k}+1},t}\log^2 \frac{2}{\delta}\notag\\
\leq & 2\sqrt{\frac{1-\tilde{c}}{1-2 \tilde{c}}} C_{sa1}^{\prime}\left(\lambda_{\hat{k}}^{\min\{1/2+r,\nu_g\}}+\lambda_{\hat{k}}^{\min\{1/2,\nu_g\}}\left( \frac{1}{\sqrt{|D|_\gamma}}\log \frac{2}{\delta}\right)^{\min\{1,r\}}\mathbb{I}_{r>1/2}\right)\notag\\
\leq & 2\sqrt{\frac{1-\tilde{c}}{1-2 \tilde{c}}} C_{sa1}^{\prime}\lambda_{\hat{k}}^{1/2+r}\left(1+\left(\log \frac{2}{\delta}\right)^{\min\{1,r\}}\mathbb{I}_{r>1/2}\right).
\end{aligned}
$$
Then we have
\begin{equation}
\begin{aligned}
&\left\|\Sigma^{1/2}\left({\theta}_{D, \lambda_{\hat{k}}}-\theta^*\right)\right\|_2\le\left\|\left(\Sigma+\lambda_{\hat{k}}I\right)^{1/2}\left({\theta}_{D, \lambda_{\hat{k}}}-\theta^*\right)\right\|_2\\
=&\left\|\left(\Sigma+\lambda_{\hat{k}}I\right)^{1/2}\left(\widehat{\Sigma}_{D}+\lambda_{\hat{k}} I\right)^{-1 / 2}\left(\widehat{\Sigma}_{D}+\lambda_{\hat{k}} I\right)^{1 / 2}\left({\theta}_{D, \lambda_{\hat{k}}}-\theta^*\right)\right\|_2\\
\le&\sqrt{\frac{1}{1-\tilde{c}}}\left\|\left(\widehat{\Sigma}_{D}+\lambda_{\hat{k}} I\right)^{1 / 2}\left({\theta}_{D, \lambda_{\hat{k}}}-\theta^*\right)\right\|_2\\
\le&2\sqrt{\frac{1}{1-2 \tilde{c}}} C_{sa1}^{\prime}\lambda_{\hat{k}}^{r+1/2}\left(1+\left(\log \frac{2}{\delta}\right)^{\min\{1,r\}}\mathbb{I}_{r>1/2}\right)\\
\le&2\sqrt{\frac{1}{1-2 \tilde{c}}} C_{sa1}^{\prime}\lambda_{k_{0}}^{r+1/2}\left(1+\left(\log \frac{2}{\delta}\right)^{\min\{1,r\}}\mathbb{I}_{r>1/2}\right)\\
=&2\sqrt{\frac{1}{1-2 \tilde{c}}} C_{sa1}^{\prime}|D|_\gamma^{-\frac{r+1/2}{2 r+s+1}}\left(1+\left(\log \frac{2}{\delta}\right)^{\min\{1,r\}}\mathbb{I}_{r>1/2}\right)\\
\le &C_{1}|D|_\gamma^{-\frac{r+1/2}{2 r+s+1}}\left(1+\left(\log \frac{2}{\delta}\right)^{\min\{1,r\}}\mathbb{I}_{r>1/2}\right),\label{2var:bias1regression}
\end{aligned}
\end{equation}
where $C_{1}=2\sqrt{\frac{1}{1-2 \tilde{c}}} C_{sa1}^{\prime}$ is a constant independent of $\delta$. If $k_{0}>\hat{k}$, i.e., $\lambda_{k_{0}}<\lambda_{\hat{k}}$. Note that
$$
\begin{aligned}
    &\left\|\Sigma^{1/2}\left({\theta}_{D, \lambda_{\hat{k}}}-\theta^*\right)\right\|_2 
\leq  \left\|\Sigma^{1/2}\left({\theta}_{D, \lambda_{\hat{k}}}-{\theta}_{D, \lambda_{k_{0}}}\right)\right\|_2+\left\|\Sigma^{1/2}\left({\theta}_{D, \lambda_{k_{0}}}-\theta^*\right)\right\|_2\notag\\
\leq & \left\|(\Sigma+\lambda_{\hat{k}}I)^{1/2}\left({\theta}_{D, \lambda_{\hat{k}}}-{\theta}_{D, \lambda_{k_{0}}}\right)\right\|_2+\left\|(\Sigma+\lambda_{k_{0}}I)^{1/2}\left({\theta}_{D, \lambda_{k_{0}}}-\theta^*\right)\right\|_2\notag\\
\le &\sqrt{\frac{1}{1-\tilde{c}}}\left\|\left(\widehat{\Sigma}_{D}+\lambda_{\hat{k}} I\right)^{1 / 2}\left({\theta}_{D, \lambda_{\hat{k}}}-{\theta}_{D, \lambda_{k_{0}}}\right)\right\|_2+\left\|(\Sigma+\lambda_{k_{0}}I)^{1/2}\left({\theta}_{D, \lambda_{k_{0}}}-\theta^*\right)\right\|_2.\notag
\end{aligned}
$$
Based on (\ref{2var:bias1regression}), we obtain
$$\left\|(\Sigma+\lambda_{k_{0}}I)^{1/2}\left({\theta}_{D, \lambda_{k_{0}}}-\theta^*\right)\right\|_2
\le C_{1}|D|_\gamma^{-\frac{r+1/2}{2 r+s+1}}\left(1+\left(\log \frac{2}{\delta}\right)^{\min\{1,r\}}\mathbb{I}_{r>1/2}\right),$$
it remains to bound $\sqrt{\frac{1}{1-\tilde{c}}}\left\|\left(\widehat{\Sigma}_{D}+\lambda_{\hat{k}} I\right)^{1 / 2}\left({\theta}_{D, \lambda_{\hat{k}}}-{\theta}_{D, \lambda_{k_{0}}}\right)\right\|_2$. Due to the definition of $\hat{k}$ yields that
\begin{equation}
\begin{aligned}
& \sqrt{\frac{1}{1-\tilde{c}}}\left\|\left(\widehat{\Sigma}_{D}+\lambda_{\hat{k}} I\right)^{1 / 2}\left({\theta}_{D, \lambda_{\hat{k}}}-{\theta}_{D, \lambda_{k_{0}}}\right)\right\|_2\\
\leq &\sqrt{\frac{1}{1-\tilde{c}}} \sum_{k=\hat{k}-1}^{k_{0}}\left\|\left(\widehat{\Sigma}_{D}+\lambda_{k+1} I\right)^{1 / 2}\left({\theta}_{D, \lambda_{k+1}}-{\theta}_{D, \lambda_{k}}\right)\right\|_2 \\
\leq &\sqrt{\frac{1}{1-\tilde{c}}}\sum_{k=\hat{k}-1}^{k_{0}} 168b\sqrt{\frac{1-\tilde{c}}{1-2 \tilde{c}}}\sqrt{\frac{1}{1-2 \tilde{c}}}M(1+C_x) \mathcal{W}_{D,\lambda_{k+1}}\log^2 \frac{2}{\delta}.\label{WWW1regression}
\end{aligned}
\end{equation}
And because $\lambda_{k+1}>\lambda_{k_{0}}=|D|_\gamma^{-\frac{1}{2 r+s+1}}, \lambda_{k+1} |D|_\gamma>1$, then
$$
\begin{aligned}
    &\mathcal{W}_{D, \lambda_{k+1}} \notag\\
    =&\frac{\left(1+4\left(\frac{13C_x }{\sqrt{\lambda_{k+1} \ell_3}}+\frac{21C_x^2 }{\lambda_{k+1} \ell_3}\right)\right)\overline{\sqrt{\mathcal{N}_{\text{empirical}}(\lambda_{k+1})}}}{\sqrt{|D|_\gamma}}+\frac{1}{|D|_\gamma \sqrt{\lambda_{k+1}}} \notag\\
    \le &\left(\frac{\left(1+4\left(\frac{13C_x }{\sqrt{\lambda_{k+1}\bar{C}_3|D|_\gamma(\log d)^{-\frac{1}{\gamma_0}}}}+\frac{21C_x^2 }{\lambda_{k+1}  \bar{C}_3|D|_\gamma(\log d)^{-\frac{1}{\gamma_0}}}
    \right)\right)^2\overline{\sqrt{\mathcal{N}(\lambda_{k+1})}}}{\sqrt{|D|_\gamma}}+\frac{1}{|D|_\gamma \sqrt{\lambda_{k+1}}} \right)\log\frac{2}{\delta}\notag\\
    \le &\left(\frac{\left(1+84\frac{C_x^2}{\bar{C}_3}(\log d)^{\frac{1}{\gamma_0}}\left(\frac{1}{\sqrt{\lambda_{k+1} |D|_\gamma}}+\frac{1 }{\lambda_{k+1}  |D|_\gamma}
    \right)\right)^2\sqrt{C_0}\lambda_{k_{0}}^{-s/2}}{\sqrt{|D|_\gamma}}+\frac{1}{|D|_\gamma \sqrt{\lambda_{k+1}}} \right)\log\frac{2}{\delta}.\notag
\end{aligned}
$$
Therefore, (\ref{WWW1regression}) can be further bounded by
\begin{equation}
\begin{aligned}
& \sqrt{\frac{1}{1-\tilde{c}}}\left\|\left(\widehat{\Sigma}_{D}+\lambda_{\hat{k}} I\right)^{1 / 2}\left({\theta}_{D, \lambda_{\hat{k}}}-{\theta}_{D, \lambda_{k_{0}}}\right)\right\|_2\\
\le &168b\frac{1}{1-2 \tilde{c}}M(1+C_x) \left(\lambda_{k_{0}}^{-s/2}|D|_\gamma^{-1/2}+\lambda_{k_{0}}^{-1/2}|D|_\gamma^{-1}\right) (\log d)^{\frac{2}{\gamma_0}}\log^2 \frac{2}{\delta}\log_q\left(\frac{C_{sa}}{q_0\sqrt{|D|_\gamma}}\right) \\
\le & 336b\frac{1}{1-2 \tilde{c}}M(1+C_x)|D|_\gamma^{-\frac{r+1/2}{2r+s+1}} (\log d)^{\frac{2}{\gamma_0}}\log^2 \frac{2}{\delta}\log_q\left(|D|_\gamma^{-1/2}\right).\label{computaleW2regression}
\end{aligned}
\end{equation}
Together all the above results, we have
\begin{equation}\begin{aligned}
&\left\|\Sigma^{1/2}\left({\theta}_{D, \lambda_{\hat{k}}}-\theta^*\right)\right\|_2
\le C_2|D|_\gamma^{-\frac{r+1/2}{2r+s+1}} (\log d)^{\frac{2}{\gamma_0}}\log^2 \frac{2}{\delta}\log_q\left(|D|_\gamma^{-1/2}\right) \left(1+\left(\log \frac{2}{\delta}\right)^{\min\{1,r\}}\mathbb{I}_{r>1/2}\right),\label{proofTregression}\end{aligned}\end{equation}
where $C_2$ is independent of $|D|$ and $\delta$. This completes the proof.
\endproof

\subsection{Proofs for adaptive spectral based linear RL method}
\subsubsection{Key lemmas} 
Following the theoretical analysis in linear regression, we first define some notations. Since linear RL can be viewed as a $T$-stage linear regression, these notations are indexed by $t$. Furthermore, due to Challenge 2, where $\theta_t^*-\theta_{D, \lambda_t, t}^{*}$ appears, the form of $\mathcal{U}_{D,\lambda_t,t}$ changes from $\left\|\left(\Sigma_t+\lambda_t I\right)^{-1 / 2}\left(\Sigma_t-\widehat{\Sigma}_{D, t}\right)\theta_t^*\right\|_{2}$ to $\left\|\left(\Sigma_t+\lambda_t I\right)^{-1 / 2}\left(\Sigma_t-\widehat{\Sigma}_{D, t}\right)\left(\theta_t^*-\theta_{D, \lambda_t, t}^{*}\right)\right\|_{2}$, and $\mathcal{S}_{D,\lambda_t,t}$ also needs to be introduced. The specific definitions are provided below.
\begin{equation}
\begin{aligned}
    & \mathcal{A}_{D,\lambda_t,t}=\left\|\left(\Sigma_{t}+\lambda_t I\right)^{1 / 2}g_{\lambda_t}\left(\widehat{\Sigma}_{D, t}\right)\left(\Sigma_{t}+\lambda_t I\right)^{1 / 2}\right\|,\label{notation:A}
 \end{aligned}
\end{equation}    
\begin{equation}
\begin{aligned}   &\mathcal{U}_{D,\lambda_t,t}=\left\|\left(\Sigma_t+\lambda_t I\right)^{-1 / 2}\left(\Sigma_t-\widehat{\Sigma}_{D, t}\right)\left(\theta_t^*-\theta_{D, \lambda_t, t}^{*}\right)\right\|_{2} ,\label{notation:U}
\end{aligned}
\end{equation}    
\begin{equation}
\begin{aligned}
    &\mathcal{P}_{D,\lambda_t,t}=\left\|\left(\Sigma_t+\lambda_t I\right)^{-1/2}\left(\widehat{E}_D[X_t Y_t^*]-\widehat{E}_D[X_t E[Y_t^* \mid X_t]]\right)\right\|_{2} ,\label{notation:P}
    \end{aligned}
\end{equation}    
\begin{equation}
\begin{aligned}
    &\mathcal{S}_{D,\lambda_t,t}=\left\|\left(\Sigma_{t}+\lambda_t I\right)^{-1 / 2}\left(\widehat{E}_{D}[X_t(Y_t^*-Y_t)]-\widehat{\Sigma}_{D, t}\left(\theta_t^*-\theta_{D, \lambda_t, t}^{*}\right) \right)\right\|_{2}, \label{notation:S}
    \end{aligned}
\end{equation}    
\begin{equation}
\begin{aligned}
    &\mathcal{W}_{D,\lambda_t,t}=\left(\frac{\left(1+4\left(\frac{13C_x }{\sqrt{\lambda_t \ell_3}}+\frac{21C_x^2 }{\lambda_t \ell_3}
\right)\right)\overline{\sqrt{\mathcal{N}_{\text{empirical}}(\lambda_t)}}}{\sqrt{|D|_\gamma}}+\frac{1}{|D|_\gamma \sqrt{\lambda_t}} \right),\label{notation:W}
\end{aligned}
\end{equation}
where $\overline{\sqrt{\mathcal{N}_{\text{empirical}}(\lambda_t)}}:=\max\{\sqrt{\mathcal{N}_{\text{empirical}}(\lambda_t)} ,1\}$, $\ell_3=\frac{|D|b_0}{2\left(\max\left\{1 , \log \left(b_0c_0 |D|\frac{2\sqrt{d}}{C_x}\right)\right\}\right)^{1 / \gamma_0}}$, $|D|_\gamma:=\frac{|D|b_0}{2\left(\max\{1 , \log \left(c_1^* |D|\right)\}\right)^{1 / \gamma_0}}$ in which $c_1^*:=c_0 b_0\max\{\frac{\sqrt{2}\max\{(T-t+2)M+\Phi_{t+1}+2C_x(\|\theta_t^*\|_2+\|\theta_{D,\lambda_t,t}^*\|_2),C_x\}}{2C_x\left((T-t+2)M+\Phi_{t+1}\right)},\frac{1}{C_x}\}. $

Equipped with the notations outlined above, we present the following key lemmas. Beyond the specific proof techniques, the main difference between the analyses of RL and regression lies in the index $t$ and the change from the regression condition $|y|<M$ to the RL condition $|y_t^*|<(T-t+2)M$.

\begin{lemma}\label{product of g}
    For $0< u \leqslant \nu_g$, we have
$$
\left\|\left(g_{\lambda_t}\left(\widehat{\Sigma}_{D,t}\right) \widehat{\Sigma}_{D,t}-I\right)\left(\lambda_t I+\widehat{\Sigma}_{D,t}\right)^u\right\| \leq 2^u\left(b+1+\gamma_u\right) \lambda_t^u .
$$
\end{lemma}

\begin{lemma}\label{lemma512}
Under Assumptions \ref{assump:mixing}-\ref{assump:effective dimension}, if $\left\|\left(\Sigma_t+\lambda_t I\right)^{-1/2}\left(\Sigma_t-\widehat{\Sigma}_{D,t}\right)\left(\Sigma_t+\lambda_t I\right)^{-1/2}\right\|\leq\tilde{c}< 1/2$, then with probability at least $1-\delta$, where $0<\delta \leq 1 / 2$, there simultaneously holds
\begin{equation}
    \begin{aligned}
& \left\|\widehat{\Sigma}_{D,t}-\Sigma_t \right\|_F \le 84C_x^2 \frac{1}{\sqrt{|D|_\gamma}}\log \frac{2}{\delta},\label{difference}
\end{aligned}
\end{equation}    
\begin{equation}
\begin{aligned}
&\left\|\left(\Sigma_t+\lambda_t I\right)^{1 / 2}\left(\widehat{\Sigma}_{D,t}+\lambda_t I\right)^{-1 / 2}\right\|  \leq  \sqrt{\frac{1}{1-\tilde{c}}}, \label{5.61}
\end{aligned}
\end{equation}    
\begin{equation}
\begin{aligned}
&\left\|\left(\widehat{\Sigma}_{D,t}+\lambda_t I\right)^{1 / 2}\left(\Sigma_t+\lambda_t I\right)^{-1 / 2}\right\|  \leq  \sqrt{\frac{1-\tilde{c}}{1-2 \tilde{c}}}, \label{5.62}
\end{aligned}
\end{equation}    
\begin{equation}
\begin{aligned}
&\mathcal{P}_{D,\lambda_t,t}
\leq  21(T-t+2)M(1+C_x) \mathcal{W}_{D, \lambda_t,t}\log^2 \frac{2}{\delta}. \label{5.72}
\end{aligned}
\end{equation}
\end{lemma}

\begin{lemma}\label{lemma:A}
    Under Assumptions \ref{assump:mixing}-\ref{assump:effective dimension}, if $\left\|\left(\Sigma_t+\lambda_t I\right)^{-1/2}\left(\Sigma_t-\widehat{\Sigma}_{D,t}\right)\left(\Sigma_t+\lambda_t I\right)^{-1/2}\right\|\leq\tilde{c}< 1/2$, then with probability at least $1-\delta$, where $0<\delta \leq 1 / 2$, there holds$$\mathcal{A}_{D,\lambda_t,t}\le2b\sqrt{\frac{1}{1-\tilde{c}}}\sqrt{\frac{1-\tilde{c}}{1-2 \tilde{c}}}.$$
\end{lemma}

\begin{lemma}\label{lemma:U}
    Under Assumptions \ref{assump:mixing}-\ref{assump:effective dimension}, with probability at least $1-\delta$, where $0<\delta \leq 1 / 2$, there holds$$\mathcal{U}_{D,\lambda_t,t}\le21(1+2C_x)\left((T-t+2)M+\Phi_{t+1}\right)\mathcal{W}_{D,\lambda_t,t}\log \frac{2}{\delta},$$
    where $\Phi_{t+1}$ is the upper bound of $| \langle\theta_{D, \lambda_{t+1}, t+1},x_t\left( s_{1:t+1},a_{1:t}, a_{t+1}\right)\rangle |$.
\end{lemma}
\proof{Proof.}
Unlike Lemma \ref{lemma:Uregression}, the random vector we construct here is given by
$$
\xi_{\mathcal{U}}(X_t):=(\Sigma_t+\lambda_t I)^{-1/2}\left(X_tX_t^\top-\Sigma_t\right)\left(\theta_t^*-\theta_{D, \lambda_t, t}^{*}\right).
$$ 
The remaining proof follows the same structure as that of Lemma \ref{lemma:Uregression}. To maintain consistency and ensure the completeness of the proof, we provide the full proof here as well. Let $\Phi_{t+1}$ denote the upper bound of $| \langle \theta_{D, \lambda_{t+1}, t+1}, x_t(s_{1:t+1}, a_{1:t}, a_{t+1}) \rangle|$. Then, from (\ref{qstard}), it follows that $|x_t^\top \theta_{D, \lambda_t, t}^{*}| \le M + \Phi_{t+1}$. Hence,
$$
\begin{aligned}
    &\|\xi_{\mathcal{U}}(x_{1,t})-\xi_{\mathcal{U}}(x_{2,t})\|_2 \notag\\
    =&\|(\Sigma_t+\lambda_t I)^{-1/2}\left(x_{1,t}x_{1,t}^\top-x_{2,t}x_{2,t}^\top\right)\left(\theta_t^*-\theta_{D, \lambda_t, t}^{*}\right)\|_2\notag\\
    \le & \lambda_t ^{-1/2}\left\| x_{1,t}(x_{1,t}^\top-x_{2,t}^\top)\left(\theta_t^*-\theta_{D, \lambda_t, t}^{*}\right)+(x_{1,t}-x_{2,t})x_{2,t}^\top\left(\theta_t^*-\theta_{D, \lambda_t, t}^{*}\right) \right\|_2\notag\\
    \le & \left((T-t+2)M+\Phi_{t+1}+C_x(\|\theta_t^*\|_2+\|\theta_{D, \lambda_t, t}^{*}\|_2)\right)\lambda_t ^{-1/2}\|x_{1,t}-x_{2,t}\|_{2},\notag
\end{aligned}
$$
thus $\xi_{\mathcal{U}}(X):=(\Sigma_t+\lambda_t I)^{-1/2}\left(X_tX_t^\top-\Sigma_t\right)\left(\theta_t^*-\theta_{D, \lambda_t, t}^{*}\right)$ is Lipschitz with constant $\left((T-t+2)M+\Phi_{t+1}+C_x(\|\theta_t^*\|_2+\|\theta_{D, \lambda_t, t}^{*}\|_2)\right)\lambda_t ^{-1/2}$, from which $(\xi_{\mathcal{U}}(x_{i,t}))_{i \geq 1}$ is $\tau$ mixing with rate $\left((T-t+2)M+\Phi_{t+1}+C_x(\|\theta_t^*\|_2+\|\theta_{D, \lambda_t, t}^{*}\|_2)\right)\lambda_t ^{-1/2} \tau_j$.
Then combined $E\left[\xi_{\mathcal{U}}(X_t)\right]=0$, 
$$
\begin{aligned}
&\left\|\xi_{\mathcal{U}}(X_t)\right\|_2 =\left\|(\Sigma_t+\lambda_t I)^{-1/2}\left(X_tX_t^\top-\Sigma_t\right)\left(\theta_t^*-\theta_{D, \lambda_t, t}^{*}\right)\right\|_2\notag\\
\le &\lambda_t ^{-1/2}\left\|X_t\left(X_t^\top\theta_t^*-X_t^\top\theta_{D, \lambda_t, t}^{*}\right)\right\|_2+\lambda_t ^{-1/2}\left\|E[X_tX_t^\top\theta_t^*]-E[X_tX_t^\top\theta_{D, \lambda_t, t}^{*}]\right\|_2\notag\\
\le&2C_x\left((T-t+2)M+\Phi_{t+1}\right)\lambda_t ^{-1/2}\notag,
\end{aligned}
$$
and
$$
\begin{aligned}
&E\left[\left\|\xi_{\mathcal{U}}(X_t)\right\|_2^2\right]\\ 
=&E \left[\operatorname{Tr}\left(\left(\theta_t^*-\theta_{D, \lambda_t, t}^{*}\right)^\top\left(X_tX_t^\top-\Sigma_t\right)(\Sigma_t+\lambda_t I)^{-1}\left(X_tX_t^\top-\Sigma_t\right)\left(\theta_t^*-\theta_{D, \lambda_t, t}^{*}\right)\right) \right] \\
=&E \left[\operatorname{Tr}\left(\left(\theta_t^*-\theta_{D, \lambda_t, t}^{*}\right)^\top X_tX_t^\top(\Sigma_t+\lambda_t I)^{-1}X_tX_t^\top\left(\theta_t^*-\theta_{D, \lambda_t, t}^{*}\right)\right)\right]\notag\\
&-\operatorname{Tr}(\left(\theta_t^*-\theta_{D, \lambda_t, t}^{*}\right)^\top\Sigma_t(\Sigma_t+\lambda_t I)^{-1}\Sigma_t\left(\theta_t^*-\theta_{D, \lambda_t, t}^{*}\right))  \\
\leq& E \left[\operatorname{Tr}\left(\left(\theta_t^*-\theta_{D, \lambda_t, t}^{*}\right)^\top X_tX_t^\top\left(\theta_t^*-\theta_{D, \lambda_t, t}^{*}\right)\right)\operatorname{Tr}\left((\Sigma_t+\lambda_t I)^{-1}X_tX_t^\top\right)\right]\\
 \leq& \left((T-t+2)M+\Phi_{t+1}\right)^2 \mathcal{N}(\lambda_t) ,
\end{aligned}
$$
with Lemma \ref{37regression} yields that with probability at least $1-\delta$:
$$
\begin{aligned}
&\left\|(\Sigma_t+\lambda_t I)^{-1/2}\left(\widehat{\Sigma}_{D,t}-\Sigma_t\right)\left(\theta_t^*-\theta_{D, \lambda_t, t}^{*}\right)\right\|_2 \notag\\
\leq & 21 \left(\frac{\left((T-t+2)M+\Phi_{t+1}\right) \sqrt{\mathcal{N}(\lambda_t)}}{\sqrt{|D|_\gamma}}+\frac{2 C_x\left((T-t+2)M+\Phi_{t+1}\right)}{\sqrt{\lambda_t}|D|_\gamma}\right)\log \frac{2}{\delta}\notag\\
\le&21(1+2C_x)\left((T-t+2)M+\Phi_{t+1}\right)\mathcal{W}_{D,\lambda_t,t}\log \frac{2}{\delta}.\notag
\end{aligned} 
$$
This completes the proof of Lemma \ref{lemma:U}.
\endproof

The following lemma is unique to linear RL and does not appear in the proof process of linear regression.

\begin{lemma}\label{lemma:S}
    Under Assumptions \ref{assump:mixing}-\ref{assump:effective dimension}, with probability at least $1-\delta$, where $0<\delta \leq 1 / 2$, there holds$$\mathcal{S}_{D,\lambda_t,t}\le42((T-t+2)M+\Phi_{t+1})(1+C_x)\mathcal{W}_{D,\lambda_t,t}\log \frac{2}{\delta}.$$
\end{lemma}

\proof{Proof.}
    We consider the random vector:
$$
\xi_{\mathcal{S}}(X_t, Y_t,Y_t^*)=(\Sigma_t+\lambda_t I)^{-1/2}\left(X_t (Y_t^*-Y_t)-X_tX_t^\top(\theta_t^*-\theta_{D,\lambda_t,t}^*)\right).
$$
Note that
$$
\begin{aligned}
    &\|\xi_{\mathcal{S}}(x_{1,t}, y_{1,t},y_{1,t}^*)-\xi_{\mathcal{S}}(x_{2,t}, y_{2,t},y_{2,t}^*)\|_2 \notag\\
    \le& \lambda_t^{-1/2}\left(\|x_{1,t} (y_{1,t}^*-y_{1,t})-x_{2,t}(y_{2,t}^*-y_{2,t})\|_2+\|x_{1,t}x_{1,t}^\top(\theta_t^*-\theta_{D,\lambda_t,t}^*)-x_{2,t}x_{2,t}^\top(\theta_t^*-\theta_{D,\lambda_t,t}^*)\|_2\right)  \notag\\
    =&\lambda_t^{-1/2}\left(\|x_{1,t} (y_{1,t}^*-y_{1,t})-x_{2,t}(y_{1,t}^*-y_{1,t})+x_{2,t}(y_{1,t}^*-y_{1,t})-x_{2,t}(y_{2,t}^*-y_{2,t})\|_2\right.\notag\\
    &\left.+\|(x_{1,t}x_{1,t}^\top-x_{1,t}x_{2,t}^\top+x_{1,t}x_{2,t}^\top-x_{2,t}x_{2,t}^\top)(\theta_t^*-\theta_{D,\lambda_t,t}^*)\|_2\right)  \notag\\
    \le&\lambda_t^{-1/2}\left(\|x_{1,t} -x_{2,t}\|_2|y_{1,t}^*-y_{1,t}|+\|x_{2,t}\|_2|y_{1,t}^*-y_{1,t}-(y_{2,t}^*-y_{2,t})|\right.\notag\\
    &\left.+\left\| x_{1,t}(x_{1,t}^\top-x_{2,t}^\top)+(x_{1,t}-x_{2,t})x_{2,t}^\top\right\|_F\|\theta_t^*-\theta_{D,\lambda_t,t}^*\|_2\right)  \notag\\
    \le&\lambda_t^{-1/2}\left(\|x_{1,t} -x_{2,t}\|_2((T-t+2)M+\Phi_{t+1}+2C_x(\|\theta_t^*\|_2+\|\theta_{D,\lambda_t,t}^*\|_2))+C_x|(y_{1,t}^*-y_{1,t})-(y_{2,t}^*-y_{2,t})|\right)  \notag\\
    \le&\sqrt{2}\lambda_t^{-1/2}\max\{(T-t+2)M+\Phi_{t+1}+2C_x(\|\theta_t^*\|_2+\|\theta_{D,\lambda_t,t}^*\|_2),C_x\}\sqrt{\|x_{1,t} -x_{2,t}\|_2^2+((y_{1,t}^*-y_{1,t})-(y_{2,t}^*-y_{2,t}))^2},\notag
\end{aligned}
$$
thus the function $\xi_{\mathcal{S}}(X_t, Y_t,Y_t^*)=(\Sigma_t+\lambda_t I)^{-1/2}\left(X_t (Y_t^*-Y_t)-X_tX_t^\top(\theta_t^*-\theta_{D,\lambda_t,t}^*)\right)$ is Lipschitz with constant $\sqrt{2}\lambda_t^{-1/2}\max\{(T-t+2)M+\Phi_{t+1}+2C_x(\|\theta_t^*\|_2+\|\theta_{D,\lambda_t,t}^*\|_2),C_x\}$, from which we deduce that $\left(\xi_{\mathcal{S}}\left(x_{i,t}, y_{i,t},y_{i,t}^*\right)\right)_{i \geq 1}$ is $\tau$ mixing with rate $\sqrt{2}\lambda_t^{-1/2}\max\{(T-t+2)M+\Phi_{t+1}+2C_x(\|\theta_t^*\|_2+\|\theta_{D,\lambda_t,t}^*\|_2),C_x\} \tau_j$.
Then combined $E\left[\xi_{\mathcal{S}}(X_t, Y_t,Y_t^*)\right]=0$, 
$$
\|\xi_{\mathcal{S}}(X_t, Y_t,Y_t^*)\|_2
=\left\|(\Sigma_t+\lambda_t I)^{-1/2}\left(X_t (Y_t^*-Y_t)-X_tX_t^\top(\theta_t^*-\theta_{D,\lambda_t,t}^*)\right)\right\|_2 
\leq 2C_x ((T-t+2)M+\Phi_{t+1}) \lambda_t^{-1/2}  ,
$$
and
$$
\begin{aligned}
E\left[\left\|\xi_{\mathcal{S}}(X_t, Y_t,Y_{t}^*)\right\|_2^2\right] & =E\left[\left( (Y_t^*-Y_t)-X_t^\top(\theta_t^*-\theta_{D,\lambda_t,t}^*)\right)^2X_t^\top(\Sigma_t+\lambda_t I)^{-1}X_t\right] \\
& \leq 4((T-t+2)M+\Phi_{t+1})^2 E\left[\operatorname{Tr}(X_t^\top(\Sigma_t+\lambda_t I)^{-1}X_t)\right]\\
& = 4((T-t+2)M+\Phi_{t+1})^2  E\left[\operatorname{Tr}(X_tX_t^\top(\Sigma_t+\lambda_t I)^{-1})\right]\\
& = 4((T-t+2)M+\Phi_{t+1})^2  \operatorname{Tr}(E\left[X_tX_t^\top\right](\Sigma_t+\lambda_t I)^{-1})\\
& =4((T-t+2)M+\Phi_{t+1})^2 \mathcal{N}(\lambda_t),
\end{aligned}
$$
with Lemma \ref{37regression} yields that with probability at least $1-\delta$:
$$
\begin{aligned}
&\left\|\left(\Sigma_{t}+\lambda_t I\right)^{-1 / 2}\left(\widehat{E}_{D}[X_t(Y_t^*-Y_t)]-\widehat{\Sigma}_{D, t}\left(\theta_t^*-\theta_{D, \lambda_t, t}^{*}\right) \right)\right\|_{2} \notag\\
\leq& 21  \left(\frac{2((T-t+2)M+\Phi_{t+1}) \sqrt{\mathcal{N}(\lambda_t)}}{\sqrt{|D|_\gamma}}+\frac{2C_x((T-t+2)M+\Phi_{t+1})}{\sqrt{\lambda_t }|D|_\gamma}\right)\log \frac{2}{\delta}\notag\\
\le& 42((T-t+2)M+\Phi_{t+1})(1+C_x)\mathcal{W}_{D,\lambda_t,t}\log \frac{2}{\delta}.\notag
\end{aligned}
$$
This completes the proof of Lemma \ref{lemma:S}.
\endproof

\subsubsection{Proof of parameter estimation error}

As shown in (\ref{error decomposition}), the error consists of three components: bias, variance, and multi-stage error. Since the first two components are analyzed in the same manner as in regression, we directly present their results (Lemmas \ref{lemma:approximation error} and \ref{lemma:sample error}) and concentrate on the multi-stage error (Lemma \ref{lemma:multistage error}).

\begin{lemma}\label{lemma:approximation error}
Under Assumptions \ref{assump:mixing}-\ref{assump:effective dimension}, with probability at least $1-\delta$, we have
$$
\left\|\left(\Sigma_{t}+\lambda_t I\right)^{1 / 2}\left(\theta_{D, \lambda_t, t}^{\diamond}-\theta_t^*\right)\right\|_2 
\leq C_{sa1}^{\prime}\left(\lambda_t^{\min\{1/2+r,\nu_g\}}+\lambda_t^{\min\{1/2,\nu_g\}}\left( \frac{1}{\sqrt{|D|_\gamma}}\log \frac{2}{\delta}\right)^{\min\{1,r\}}\mathbb{I}_{r>1/2}\right),$$
where $C_{sa1}^{\prime}=\left(\frac{1}{1-\tilde{c}}\right)^{r+1/2}C\left(\gamma_{1/2+r} +b+1\right)\max\{1,r C_x^{2(r-1)}\} \left(84C_x^2\right)^{\min\{1,r\}}$.
\end{lemma}

\begin{lemma}\label{lemma:sample error}
For any $t=1, \ldots, T$, it holds that
$$
\left\|\left(\Sigma_{t}+\lambda_t I\right)^{1 / 2}\left(\theta_{D, \lambda_t, t}^{\diamond}-\hat{\theta}_{D, \lambda_t, t}\right)\right\|_{2} \leq \mathcal{A}_{D,\lambda_t,t}\mathcal{P}_{D,\lambda_t,t}   .
$$
\end{lemma}

\begin{lemma}\label{lemma:multistage error}
Under Assumption \ref{assump:conditional probability}, for any $t=1, \ldots, T$, it holds that
$$
 \left\|\left(\Sigma_{t}+\lambda_t I\right)^{1 / 2}\left(\theta_{D, \lambda_t, t}-\hat{\theta}_{D, \lambda_t, t}\right)\right\|_{2} 
\leq  \mathcal{A}_{D, \lambda_t, t} \mathcal{S}_{D, \lambda, t}+\mathcal{A}_{D, \lambda_t, t} \mathcal{U}_{D, \lambda_t, t}+\mu^{1 / 2} \mathcal{A}_{D, \lambda_t, t}\left\|\theta_{D, \lambda_{t+1}, t+1}-\theta_{t+1}^*\right\|_{\Sigma_{t+1}}.
$$
\end{lemma}

\proof{Proof.}
For any $t=1, \ldots, T$, we have
\begin{equation}
\begin{aligned}
& \left\|\left(\Sigma_{t}+\lambda_t I\right)^{1 / 2}\left(\theta_{D, \lambda_t, t}-\hat{\theta}_{D, \lambda_t, t}\right)\right\|_{2}\\
=&\left\|\left(\Sigma_{t}+\lambda_t I\right)^{1 / 2}g_{\lambda_t}\left(\widehat{\Sigma}_{D, t}\right) \widehat{E}_{D}\left[X_t (Y_t^*-Y_t)\right]\right\|_{2} \\
\leq & \left\|\left(\Sigma_{t}+\lambda_t I\right)^{1 / 2}g_{\lambda_t}\left(\widehat{\Sigma}_{D, t}\right)\left(\widehat{E}_{D}[X_t(Y_t^*-Y_t)]-\widehat{\Sigma}_{D, t}\left(\theta_t^*-\theta_{D, \lambda_t, t}^{*}\right) \right)\right\|_{2} \\
&+  \left\|\left(\Sigma_t+\lambda_t I\right)^{1 / 2}g_{\lambda_t}\left(\widehat{\Sigma}_{D, t}\right)\left(\widehat{\Sigma}_{D, t}-\Sigma_t\right)\left(\theta_t^*-\theta_{D, \lambda_t, t}^{*}\right)\right\|_{2}\\
&+  \left\|\left(\Sigma_t+\lambda_t I\right)^{1 / 2}g_{\lambda_t}\left(\widehat{\Sigma}_{D, t}\right)\Sigma_t\left(\theta_t^*-\theta_{D, \lambda_t, t}^{*}\right)\right\|_{2}.\label{trouble}
\end{aligned}
\end{equation}
First, we analyze the first term of (\ref{trouble}), by (\ref{notation:A}) and (\ref{notation:S}), we have
\begin{equation}\begin{aligned}
&\left\|\left(\Sigma_{t}+\lambda_t I\right)^{1 / 2}g_{\lambda_t}\left(\widehat{\Sigma}_{D, t}\right)\left(\widehat{E}_{D}[X_t(Y_t^*-Y_t)]-\widehat{\Sigma}_{D, t}\left(\theta_t^*-\theta_{D, \lambda_t, t}^{*}\right) \right)\right\|_{2} 
\le \mathcal{A}_{D,\lambda_t,t}\mathcal{S}_{D,\lambda_t,t}.\label{firstterm}
\end{aligned}
\end{equation}
Then, we analyze the second term of (\ref{trouble}), by (\ref{notation:A}) and (\ref{notation:U}), we have
\begin{equation}\begin{aligned}
&\left\|\left(\Sigma_t+\lambda_t I\right)^{1 / 2}g_{\lambda_t}\left(\widehat{\Sigma}_{D, t}\right)\left(\widehat{\Sigma}_{D, t}-\Sigma_t\right)\left(\theta_t^*-\theta_{D, \lambda_t, t}^{*}\right)\right\|_{2} 
\leq  \mathcal{A}_{D, \lambda_t, t} \mathcal{U}_{D, \lambda_t, t}.\label{secondterm}
\end{aligned}
\end{equation}
To bound the last term of (\ref{trouble}), by (\ref{notation:A}) and the property $\|z\|_A^2=z^\top Az=z^\top A^{1/2}A^{1/2}z=\|A^{1/2}z\|_2^2$, there holds
\begin{equation}\begin{aligned}
&\left\|\left(\Sigma_t+\lambda_t I\right)^{1 / 2}g_{\lambda_t}\left(\widehat{\Sigma}_{D, t}\right)\Sigma_t\left(\theta_t^*-\theta_{D, \lambda_t, t}^{*}\right)\right\|_{2} \\
\le&\left\|\left(\Sigma_t+\lambda_t I\right)^{1 / 2}g_{\lambda_t}\left(\widehat{\Sigma}_{D, t}\right)\left(\Sigma_t+\lambda_t I\right)^{1 / 2}\Sigma_t^{1/2}\left(\theta_t^*-\theta_{D, \lambda_t, t}^{*}\right)\right\|_{2}\\
\leq&  \mathcal{A}_{D, \lambda_t, t}\left\|\Sigma_t^{1/2}\left(\theta_t^*-\theta_{D, \lambda_t, t}^{*}\right)\right\|_{2} \\
=&\mathcal{A}_{D, \lambda_t, t}\left\|\theta_t^*-\theta_{D, \lambda_t, t}^{*}\right\|_{\Sigma_t} .\label{thirdterm1}
\end{aligned}
\end{equation}
Due to Assumption \ref{assump:conditional probability}, (\ref{property:q}) and (\ref{qstard}), there holds
\begin{equation}\begin{aligned}
&\left\|\theta_t^*-\theta_{D, \lambda_t, t}^{*}\right\|_{\Sigma_t}^2\\
=&\left(\theta_t^*-\theta_{D, \lambda_t, t}^{*}\right)^\top E[X_tX_t^\top]\left(\theta_t^*-\theta_{D, \lambda_t, t}^{*}\right)
=E[\left(\theta_t^*-\theta_{D, \lambda_t, t}^{*}\right)^\top X_tX_t^\top\left(\theta_t^*-\theta_{D, \lambda_t, t}^{*}\right)\mid D]\\
=&E\left[\langle\theta_{D, \lambda_t, t}^{*}-\theta_t^*,X_t\rangle^2 \mid D\right] 
=E\left[\left(\langle\theta_{D, \lambda_t, t}^{*},X_t\rangle-\langle\theta_t^*,X_t\rangle\right)^2 \mid D\right] \\
=& E\left[\left(\max _{a_{t+1}} \langle\theta_{D, \lambda_{t+1}, t+1},X_t\left(A_{1:t},S_{1:t+1}, a_{t+1}\right)\rangle-\max _{a_{t+1}} \langle\theta_{t+1}^*,X_t\left(A_{1:t},S_{1: t+1}, a_{t+1}\right)\rangle\right)^2 \mid D\right] \\
 \leq& E\left[\max _{a_{t+1}}\left(\langle\theta_{D, \lambda_{t+1}, t+1},X_t\left(A_{1:t},S_{1: t+1}, a_{t+1}\right)\rangle-\langle\theta_{t+1}^*,X_t\left(A_{1:t},S_{1: t+1}, a_{t+1}\right)\rangle\right)^2 \mid D\right]\\
 \leq &E\left[\mu \sum_{a \in \mathcal{A}_{t+1}}\left(\langle\theta_{D, \lambda_{t+1}, t+1},X_t\left(A_{1:t},S_{1: t+1}, a\right)\rangle-\langle\theta_{t+1}^*,X_t\left(A_{1:t},S_{1: t+1}, a\right)\rangle\right)^2 p_t\left(a \mid A_{1:t},S_{1:t+1}\right) \mid D\right]\\
=&\mu E\left[\left(\langle\theta_{D, \lambda_{t+1}, t+1},X_{t+1}\rangle-\langle\theta_{t+1}^*,X_{t+1}\rangle\right)^2 \mid D\right]\\
=&\mu\left\|\theta_{D, \lambda_{t+1}, t+1}-\theta_{t+1}^*\right\|_{\Sigma_{t+1}}^2 .\label{mutheta}
\end{aligned}
\end{equation}
Then, substituting (\ref{mutheta}) into (\ref{thirdterm1}) yields
\begin{equation}\begin{aligned}
\left\|\left(\Sigma_t+\lambda_t I\right)^{1 / 2}g_{\lambda_t}\left(\widehat{\Sigma}_{D, t}\right)\Sigma_t\left(\theta_t^*-\theta_{D, \lambda_t, t}^{*}\right)\right\|_{2}  \leq \mu^{1 / 2} \mathcal{A}_{D, \lambda_t, t}\left\|\theta_{D, \lambda_{t+1}, t+1}-\theta_{t+1}^*\right\|_{\Sigma_{t+1}}.\label{thirdterm}
\end{aligned}
\end{equation}
Therefore, by substituting (\ref{firstterm}), (\ref{secondterm}), and (\ref{thirdterm}) into (\ref{trouble}), we obtain
$$
\begin{aligned}
& \left\|\left(\Sigma_{t}+\lambda_t I\right)^{1 / 2}\left(\theta_{D, \lambda_t, t}-\hat{\theta}_{D, \lambda_t, t}\right)\right\|_{2} 
\leq  \mathcal{A}_{D, \lambda_t, t} \mathcal{S}_{D, \lambda, t}+\mathcal{A}_{D, \lambda_t, t} \mathcal{U}_{D, \lambda_t, t}+\mu^{1 / 2} \mathcal{A}_{D, \lambda_t, t}\left\|\theta_{D, \lambda_{t+1}, t+1}-\theta_{t+1}^*\right\|_{\Sigma_{t+1}}.
\end{aligned}
$$
This finishes the proof of Lemma \ref{lemma:multistage error}.
\endproof

\begin{proposition}\label{prop:error decomposition}
Under Assumptions \ref{assump:mixing}-\ref{assump:conditional probability}, with probability at least $1-\delta$, we have
$$
\begin{aligned}
& \left\|\left(\Sigma_{ t}+\lambda_t I\right)^{1 / 2}\left(\theta_{D, \lambda_t, t}-\theta_t^*\right)\right\|_{2}\\
\leq & C_{sa1}^{\prime}\left(\lambda_t^{\min\{1/2+r,\nu_g\}}+\lambda_t^{\min\{1/2,\nu_g\}}\left( \frac{1}{\sqrt{|D|_\gamma}}\log \frac{2}{\delta}\right)^{\min\{1,r\}}\mathbb{I}_{r>1/2}\right)\notag\\
&+2b\sqrt{\frac{1}{1-2 \tilde{c}}}\left(84((T-t+2)M+\Phi_{t+1})(1+C_x)\mathcal{W}_{D,\lambda_t,t}\log^2 \frac{2}{\delta}\right)\notag\\
&+\mu^{1/2} \mathcal{A}_{D, \lambda_t, t}\left\|\Sigma_{t+1}^{1/2}\left(\theta_{D, \lambda_{t+1}, t+1}-\theta_{t+1}^*\right)\right\|_{2}.\notag
\end{aligned}
$$
\end{proposition}

\proof{Proof.}
Inserting Lemmas \ref{lemma:approximation error}, \ref{lemma:sample error} and \ref{lemma:multistage error} into (\ref{error decomposition}), we obtain for any $t=1,2, \ldots, T$,
\begin{equation}
\begin{aligned}
& \left\|\left(\Sigma_{ t}+\lambda_t I\right)^{1 / 2}\left(\theta_{D, \lambda_t, t}-\theta_t^*\right)\right\|_{2} \\
\leq & C_{sa1}^{\prime}\left(\lambda_t^{\min\{1/2+r,\nu_g\}}+\lambda_t^{\min\{1/2,\nu_g\}}\left( \frac{1}{\sqrt{|D|_\gamma}}\log \frac{2}{\delta}\right)^{\min\{1,r\}}\mathbb{I}_{r>1/2}\right)\\
&+\mathcal{A}_{D, \lambda_t, t}\left(\mathcal{P}_{D, \lambda_t, t}+\mathcal{S}_{D, \lambda_t, t}+\mathcal{U}_{D, \lambda_t, t}\right)+\mu^{1/2} \mathcal{A}_{D, \lambda_t, t}\left\|\theta_{D, \lambda_{t+1}, t+1}-\theta_{t+1}^*\right\|_{\Sigma_{t+1}} \\
\leq & C_{sa1}^{\prime}\left(\lambda_t^{\min\{1/2+r,\nu_g\}}+\lambda_t^{\min\{1/2,\nu_g\}}\left( \frac{1}{\sqrt{|D|_\gamma}}\log \frac{2}{\delta}\right)^{\min\{1,r\}}\mathbb{I}_{r>1/2}\right)+\mathcal{A}_{D, \lambda_t, t}\\
&\cdot\left(\mathcal{P}_{D, \lambda_t, t}+\mathcal{S}_{D, \lambda_t, t}+\mathcal{U}_{D, \lambda_t, t}\right)+\mu^{1/2} \mathcal{A}_{D, \lambda_t, t}\left\|\Sigma_{t+1}^{1/2}\left(\theta_{D, \lambda_{t+1}, t+1}-\theta_{t+1}^*\right)\right\|_{2} .\label{medium}
\end{aligned}
\end{equation}
Substituting (\ref{5.72}), Lemmas \ref{lemma:A}, \ref{lemma:U}, and \ref{lemma:S} into (\ref{medium}) yields the following:
    $$
\begin{aligned}
& \left\|\left(\Sigma_{ t}+\lambda_t I\right)^{1 / 2}\left(\theta_{D, \lambda_t, t}-\theta_t^*\right)\right\|_{2}\\
\leq & C_{sa1}^{\prime}\left(\lambda_t^{\min\{1/2+r,\nu_g\}}+\lambda_t^{\min\{1/2,\nu_g\}}\left( \frac{1}{\sqrt{|D|_\gamma}}\log \frac{2}{\delta}\right)^{\min\{1,r\}}\mathbb{I}_{r>1/2}\right)\notag\\
&+2b\sqrt{\frac{1}{1-2 \tilde{c}}}\left(84((T-t+2)M+\Phi_{t+1})(1+C_x)\mathcal{W}_{D,\lambda_t,t}\log^2 \frac{2}{\delta}\right)\notag\\
&+\mu^{1/2} \mathcal{A}_{D, \lambda_t, t}\left\|\Sigma_{t+1}^{1/2}\left(\theta_{D, \lambda_{t+1}, t+1}-\theta_{t+1}^*\right)\right\|_{2}.\notag
\end{aligned}
$$
This finishes the proof of Proposition \ref{prop:error decomposition}.
\endproof

Leveraging the iterative relationship among the parameter estimation errors outlined in Proposition \ref{prop:error decomposition}, and motivated by the proof sketch of parameter estimation error bound under adaptive parameter selection in linear regression, we proceed to derive the parameter estimation error for linear RL under adaptive parameter selection.
\begin{lemma}\label{thm:as}
Let $\delta \in(0,1/2)$. Under Assumptions \ref{assump:mixing}-\ref{assump:conditional probability}, and $\lambda_{\hat{k}_t}$ obtained by (\ref{adas}), then with probability at least $1-\delta$, there holds
\begin{equation}
\begin{aligned}
&\left\|\Sigma_t^{1/2}\left({\theta}_{D, \lambda_{\hat{k}_t},t}-\theta_t^*\right)\right\|_2\le C_{6}\sum_{\ell=t}^T\left((T-\ell+2)M+\Phi_{\ell+1}\right)\\
&\cdot|D|_\gamma^{-\frac{r+1/2}{2r+s+1}} (\log d)^{\frac{2}{\gamma_0}}\log^2 \frac{2}{\delta}\log_q\left(|D|_\gamma^{-1/2}\right) \left(1+\left(\log \frac{2}{\delta}\right)^{\min\{1,r\}}\mathbb{I}_{r>1/2}\right) ,\end{aligned}
\end{equation}
where $C_6$ is the constant independent of $|D|$ and $\delta$.
\end{lemma}


\proof{Proof of Lemma \ref{thm:as}.}
We employ a recursive method for the proof. First, we analyze the parameter estimation error at step $T$, then use Proposition \ref{prop:error decomposition} to compute the parameter estimation error at step $T-1$. Next, we substitute the parameter estimation error at step $T-1$ into Proposition \ref{prop:error decomposition} to compute the error at step $T-2$, and continue this process recursively, ultimately deriving the general parameter estimation error at step $t$.

\textbf{(a) For step $\boldsymbol{t=T}$:} There exists a $k_{T,0} \in\left[1, K_{D,q,T}\right]$ such that $\lambda_{k_{T,0}}=q_{T,0} q^{k_{T,0}} \sim |D|_\gamma^{-\frac{1}{2 r+s+1}}$. Following the proof idea of Theorem \ref{thm:asregression}, we also analyze by considering two separate cases.
If $k_{T,0} \leq \hat{k}_T$, i.e., $\lambda_{k_{T,0}} \geq \lambda_{\hat{k}_T}$, then by the definition of $\hat{k}_T$, Proposition \ref{prop:error decomposition}, and the fact that $\theta_{D, \lambda_{T+1}^\prime, T+1} =\theta_{D, \lambda_{T+1}, T+1} = \theta_{T+1}^* = 0$, we have
\begin{equation}
\begin{aligned}
&2b\sqrt{\frac{1}{1-2 \tilde{c}}}\left(84(2M+\Phi_{T+1})(1+C_x)\mathcal{W}_{D,\lambda_{\hat{k}_T+1},T}\log^2 \frac{2}{\delta}\right)\\
\leq &C_{sa1}^{\prime}\left((\lambda_{\hat{k}_T})^{\min\{1/2+r,\nu_g\}}+(\lambda_{\hat{k}_T})^{\min\{1/2,\nu_g\}}\left( \frac{1}{\sqrt{|D|_\gamma}}\log \frac{2}{\delta}\right)^{\min\{1,r\}}\mathbb{I}_{r>1/2}\right).\label{constant3}
\end{aligned}
\end{equation}
Therefore, we can further obtain that
$$
\begin{aligned}
& \left\|\left(\widehat{\Sigma}_{D,T}+\lambda_{\hat{k}_T} I\right)^{1 / 2}\left({\theta}_{D, \lambda_{\hat{k}_T},T}-\theta_T^*\right)\right\|_2 \\
\le&\sqrt{\frac{1-\tilde{c}}{1-2 \tilde{c}}}\left\|\left(\Sigma_T+\lambda_{\hat{k}_T} I\right)^{1 / 2}\left({\theta}_{D, \lambda_{\hat{k}_T},T}-\theta_T^*\right)\right\|_2\\
\leq & \sqrt{\frac{1-\tilde{c}}{1-2 \tilde{c}}} C_{sa1}^{\prime}\left(\lambda_{\hat{k}_T}^{\min\{1/2+r,\nu_g\}}+\lambda_{\hat{k}_T}^{\min\{1/2,\nu_g\}}\left( \frac{1}{\sqrt{|D|_\gamma}}\log \frac{2}{\delta}\right)^{\min\{1,r\}}\mathbb{I}_{r>1/2}\right)\notag\\
&+\sqrt{\frac{1-\tilde{c}}{1-2 \tilde{c}}} 2b\sqrt{\frac{1}{1-2 \tilde{c}}}\left(84(2M+\Phi_{T+1})(1+C_x)\mathcal{W}_{D,\lambda_{\hat{k}_T},t}\log^2 \frac{2}{\delta}\right)\notag\\
\leq & 2\sqrt{\frac{1-\tilde{c}}{1-2 \tilde{c}}} C_{sa1}^{\prime}\left(\lambda_{\hat{k}_T}^{\min\{1/2+r,\nu_g\}}+\lambda_{\hat{k}_T}^{\min\{1/2,\nu_g\}}\left( \frac{1}{\sqrt{|D|_\gamma}}\log \frac{2}{\delta}\right)^{\min\{1,r\}}\mathbb{I}_{r>1/2}\right)\notag\\
\leq & 2\sqrt{\frac{1-\tilde{c}}{1-2 \tilde{c}}} C_{sa1}^{\prime}\lambda_{\hat{k}_T}^{1/2+r}\left(1+\left(\log \frac{2}{\delta}\right)^{\min\{1,r\}}\mathbb{I}_{r>1/2}\right).
\end{aligned}
$$
Then we have
\begin{equation}
\begin{aligned}
&\left\|\Sigma_T^{1/2}\left({\theta}_{D, \lambda_{\hat{k}_T},T}-\theta_T^*\right)\right\|_2\le\left\|\left(\Sigma_T+\lambda_{\hat{k}_T}I\right)^{1/2}\left({\theta}_{D, \lambda_{\hat{k}_T},T}-\theta_T^*\right)\right\|_2\\
\le&\sqrt{\frac{1}{1-\tilde{c}}}\left\|\left(\widehat{\Sigma}_{D,T}+\lambda_{\hat{k}_T} I\right)^{1 / 2}\left({\theta}_{D, \lambda_{\hat{k}_T},T}-\theta_T^*\right)\right\|_2
\le2\sqrt{\frac{1}{1-2 \tilde{c}}} C_{sa1}^{\prime}\lambda_{\hat{k}_T}^{r+1/2}\left(1+\left(\log \frac{2}{\delta}\right)^{\min\{1,r\}}\mathbb{I}_{r>1/2}\right)\\
\le&2\sqrt{\frac{1}{1-2 \tilde{c}}} C_{sa1}^{\prime}\lambda_{k_{T,0}}^{r+1/2}\left(1+\left(\log \frac{2}{\delta}\right)^{\min\{1,r\}}\mathbb{I}_{r>1/2}\right)
\le C_{3}|D|_\gamma^{-\frac{r+1/2}{2 r+s+1}}\left(1+\left(\log \frac{2}{\delta}\right)^{\min\{1,r\}}\mathbb{I}_{r>1/2}\right),\label{2var:bias1}
\end{aligned}
\end{equation}
where $C_{3}=2\sqrt{\frac{1}{1-2 \tilde{c}}} C_{sa1}^{\prime}$ is a constant independent of $T$ and $\delta$. If $k_{T,0}>\hat{k}_T$, i.e., $\lambda_{k_{T,0}}<\lambda_{\hat{k}_T}$. Note that
$$
\begin{aligned}
    &\left\|\Sigma_T^{1/2}\left({\theta}_{D, \lambda_{\hat{k}_T},T}-\theta_T^*\right)\right\|_2 \\
\le &\sqrt{\frac{1}{1-\tilde{c}}}\left\|\left(\widehat{\Sigma}_{D,T}+\lambda_{\hat{k}_T} I\right)^{1 / 2}\left({\theta}_{D, \lambda_{\hat{k}_T},T}-{\theta}_{D, \lambda_{k_{T,0}},T}\right)\right\|_2
+\left\|(\Sigma_T+\lambda_{k_{T,0}}I)^{1/2}\left({\theta}_{D, \lambda_{k_{T,0}},T}-\theta_T^*\right)\right\|_2.\notag
\end{aligned}
$$
Based on (\ref{2var:bias1}), we obtain
$$\left\|(\Sigma_T+\lambda_{k_{T,0}}I)^{1/2}\left({\theta}_{D, \lambda_{k_{T,0}},T}-\theta_T^*\right)\right\|_2
\le C_{3}|D|_\gamma^{-\frac{r+1/2}{2 r+s+1}}\left(1+\left(\log \frac{2}{\delta}\right)^{\min\{1,r\}}\mathbb{I}_{r>1/2}\right),$$
it remains to bound $\sqrt{\frac{1}{1-\tilde{c}}}\left\|\left(\widehat{\Sigma}_{D,T}+\lambda_{\hat{k}_T} I\right)^{1 / 2}\left({\theta}_{D, \lambda_{\hat{k}_T},T}-{\theta}_{D, \lambda_{k_{T,0}},T}\right)\right\|_2$. Due to the definition of $\hat{k}_T$ yields that
$$
    \begin{aligned}
& \sqrt{\frac{1}{1-\tilde{c}}}\left\|\left(\widehat{\Sigma}_{D,T}+\lambda_{\hat{k}_T} I\right)^{1 / 2}\left({\theta}_{D, \lambda_{\hat{k}_T},T}-{\theta}_{D, \lambda_{k_{T,0}},T}\right)\right\|_2\\
\leq &\sqrt{\frac{1}{1-\tilde{c}}} \sum_{k_T=\hat{k}_T-1}^{k_{T,0}}\left\|\left(\widehat{\Sigma}_{D,T}+\lambda_{k_T+1} I\right)^{1 / 2}\left({\theta}_{D, \lambda_{k_T+1},T}-{\theta}_{D, \lambda_{k_T},T}\right)\right\|_2 \\
\leq &\sqrt{\frac{1}{1-\tilde{c}}}\sum_{k_T=\hat{k}_T-1}^{k_{T,0}} 8b\sqrt{\frac{1-\tilde{c}}{1-2 \tilde{c}}}\sqrt{\frac{1}{1-2 \tilde{c}}}\left(84(2M+\Phi_{T+1})(1+C_x)\mathcal{W}_{D,\lambda_{k_T+1},T}\log^2 \frac{2}{\delta}\right)\\
\le &\frac{672b}{1-2 \tilde{c}}(2M+\Phi_{T+1})(1+C_x)\left(\lambda_{k_{T,0}}^{-s/2}|D|_\gamma^{-1/2}+\lambda_{k_{T,0}}^{-1/2}|D|_\gamma^{-1}\right) (\log d)^{\frac{2}{\gamma_0}}\log^2 \frac{2}{\delta}\log_q\left(\frac{C_{sa}}{q_T\sqrt{|D|_\gamma}}\right) \\
\le & \frac{672b}{1-2 \tilde{c}}(2M+\Phi_{T+1})(1+C_x)|D|_\gamma^{-\frac{r+1/2}{2r+s+1}} (\log d)^{\frac{2}{\gamma_0}}\log^2 \frac{2}{\delta}\log_q\left(|D|_\gamma^{-1/2}\right).
\end{aligned}
$$
Together all the above results, we have
\begin{equation}\begin{aligned}
&\left\|\Sigma_T^{1/2}\left({\theta}_{D, \lambda_{\hat{k}_T},T}-\theta_T^*\right)\right\|_2\\
\le& C_4(2M+\Phi_{T+1})|D|_\gamma^{-\frac{r+1/2}{2r+s+1}} (\log d)^{\frac{2}{\gamma_0}}\log^2 \frac{2}{\delta}\log_q\left(|D|_\gamma^{-1/2}\right) \left(1+\left(\log \frac{2}{\delta}\right)^{\min\{1,r\}}\mathbb{I}_{r>1/2}\right),\label{proofTT}\end{aligned}
\end{equation}
where $C_4$ is the constant independent of $|D|$ and $\delta$.
\textbf{(b) For step $\boldsymbol{t=T-1}$:} Combined (\ref{proofTT}) and Proposition \ref{prop:error decomposition}, there holds
$$
\begin{aligned}
& \left\|\left(\Sigma_{T-1}+\lambda_{T-1} I\right)^{1 / 2}\left(\theta_{D, \lambda_{T-1}, T-1}-\theta_{T-1}^*\right)\right\|_{2}\\
\leq & C_{sa1}^{\prime}\left(\lambda_{T-1}^{\min\{1/2+r,\nu_g\}}+\lambda_{T-1}^{\min\{1/2,\nu_g\}}\left( \frac{1}{\sqrt{|D|_\gamma}}\log \frac{2}{\delta}\right)^{\min\{1,r\}}\mathbb{I}_{r>1/2}\right)\notag\\
&+2b\sqrt{\frac{1}{1-2 \tilde{c}}}\left(84(3M+\Phi_{T})(1+C_x)\mathcal{W}_{D,\lambda_{T-1},T-1}\log^2 \frac{2}{\delta}\right)\notag\\
&+\mu^{1/2} 2b\sqrt{\frac{1}{1-2 \tilde{c}}}C_4(2M+\Phi_{T+1})|D|_\gamma^{-\frac{r+1/2}{2r+s+1}} (\log d)^{\frac{2}{\gamma_0}}\log^2 \frac{2}{\delta}
\log_q\left(|D|_\gamma^{-1/2}\right) \left(1+\left(\log \frac{2}{\delta}\right)^{\min\{1,r\}}\mathbb{I}_{r>1/2}\right).\notag
\end{aligned}
$$
Similarly, there exists a $k_{T-1,0} \in\left[1, K_{D,q,T-1}\right]$ such that $\lambda_{k_{T-1,0}}=q_{T,0} q^{k_{T-1,0}} \sim |D|_\gamma^{-\frac{1}{2 r+s+1}}$. Similarly, if $k_{T-1,0} \leq \hat{k}_{T-1}$, i.e., $\lambda_{k_{T-1,0}} \geq \lambda_{\hat{k}_{T-1}}$, then by the definition of $\hat{k}_{T-1}$, we have
\begin{equation}\begin{aligned}
&2b\sqrt{\frac{1}{1-2 \tilde{c}}}\left(84(2M+\Phi_{T+1})(1+C_x)\mathcal{W}_{D,\lambda_{\hat{k}_T+1},T}\log^2 \frac{2}{\delta}\right)\\
\leq &C_{sa1}^{\prime}\left((\lambda_{\hat{k}_T})^{\min\{1/2+r,\nu_g\}}+(\lambda_{\hat{k}_T})^{\min\{1/2,\nu_g\}}\left( \frac{1}{\sqrt{|D|_\gamma}}\log \frac{2}{\delta}\right)^{\min\{1,r\}}\mathbb{I}_{r>1/2}\right)\\
&+\mu^{1/2} 2bC_4(2M+\Phi_{T+1})|D|_\gamma^{-\frac{r+1/2}{2r+s+1}} (\log d)^{\frac{2}{\gamma_0}}\log^2 \frac{2}{\delta}\log_q\left(|D|_\gamma^{-1/2}\right) \left(1+\left(\log \frac{2}{\delta}\right)^{\min\{1,r\}}\mathbb{I}_{r>1/2}\right).\label{constant32}
\end{aligned}
\end{equation}
Therefore, we can further obtain that
$$
\begin{aligned}
& \left\|\left(\widehat{\Sigma}_{D,T-1}+\lambda_{\hat{k}_{T-1}} I\right)^{1 / 2}\left({\theta}_{D, \lambda_{\hat{k}_{T-1}},T-1}-\theta_{T-1}^*\right)\right\|_2 
\le\sqrt{\frac{1-\tilde{c}}{1-2 \tilde{c}}}\left\|\left(\Sigma_{T-1}+\lambda_{\hat{k}_{T-1}} I\right)^{1 / 2}\left({\theta}_{D, \lambda_{\hat{k}_{T-1}},T-1}-\theta_{T-1}^*\right)\right\|_2\\
\leq & 2\sqrt{\frac{1-\tilde{c}}{1-2 \tilde{c}}} C_{sa1}^{\prime}\lambda_{\hat{k}_{T-1}}^{1/2+r}\left(1+\left(\log \frac{2}{\delta}\right)^{\min\{1,r\}}\mathbb{I}_{r>1/2}\right)+\frac{\sqrt{1-\tilde{c}}}{1-2 \tilde{c}}\mu^{1/2} 2bC_4(2M+\Phi_{T+1})\notag\\
&\cdot|D|_\gamma^{-\frac{r+1/2}{2r+s+1}} (\log d)^{\frac{2}{\gamma_0}}\log^2 \frac{2}{\delta}\log_q\left(|D|_\gamma^{-1/2}\right) \left(1+\left(\log \frac{2}{\delta}\right)^{\min\{1,r\}}\mathbb{I}_{r>1/2}\right)\notag.
\end{aligned}
$$
Then we have
\begin{equation}
\begin{aligned}
&\left\|\Sigma_{T-1}^{1/2}\left({\theta}_{D, \lambda_{\hat{k}_{T-1}},T-1}-\theta_{T-1}^*\right)\right\|_2\\
\le&\sqrt{\frac{1}{1-\tilde{c}}}\left\|\left(\widehat{\Sigma}_{D,T-1}+\lambda_{\hat{k}_{T-1}} I\right)^{1 / 2}\left({\theta}_{D, \lambda_{\hat{k}_{T-1}},T-1}-\theta_{T-1}^*\right)\right\|_2\\
\le&2\sqrt{\frac{1}{1-2 \tilde{c}}} C_{sa1}^{\prime}\lambda_{\hat{k}_{T-1}}^{r+1/2}\left(1+\left(\log \frac{2}{\delta}\right)^{\min\{1,r\}}\mathbb{I}_{r>1/2}\right)+\frac{1}{1-2 \tilde{c}}\mu^{1/2} 2bC_4(2M+\Phi_{T+1})\\
&\cdot|D|_\gamma^{-\frac{r+1/2}{2r+s+1}} (\log d)^{\frac{2}{\gamma_0}}\log^2 \frac{2}{\delta}\log_q\left(|D|_\gamma^{-1/2}\right) \left(1+\left(\log \frac{2}{\delta}\right)^{\min\{1,r\}}\mathbb{I}_{r>1/2}\right)\\
\le&2\sqrt{\frac{1}{1-2 \tilde{c}}} C_{sa1}^{\prime}\lambda_{k_{T-1,0}}^{r+1/2}\left(1+\left(\log \frac{2}{\delta}\right)^{\min\{1,r\}}\mathbb{I}_{r>1/2}\right)+\frac{1}{1-2 \tilde{c}}\mu^{1/2} 2bC_4(2M+\Phi_{T+1})\\
&\cdot|D|_\gamma^{-\frac{r+1/2}{2r+s+1}} (\log d)^{\frac{2}{\gamma_0}}\log^2 \frac{2}{\delta}\log_q\left(|D|_\gamma^{-1/2}\right) \left(1+\left(\log \frac{2}{\delta}\right)^{\min\{1,r\}}\mathbb{I}_{r>1/2}\right)\\
\le &C_{5}|D|_\gamma^{-\frac{r+1/2}{2 r+s+1}}\left(1+\left(\log \frac{2}{\delta}\right)^{\min\{1,r\}}\mathbb{I}_{r>1/2}\right)+\frac{1}{1-2 \tilde{c}}\mu^{1/2} 2bC_4(2M+\Phi_{T+1})\\
&\cdot|D|_\gamma^{-\frac{r+1/2}{2r+s+1}} (\log d)^{\frac{2}{\gamma_0}}\log^2 \frac{2}{\delta}\log_q\left(|D|_\gamma^{-1/2}\right) \left(1+\left(\log \frac{2}{\delta}\right)^{\min\{1,r\}}\mathbb{I}_{r>1/2}\right),\label{2var:bias2}
\end{aligned}
\end{equation}
where $C_{5}=2\sqrt{\frac{1}{1-2 \tilde{c}}} C_{sa1}^{\prime}$ is a constant independent of $T-1$ and $\delta$. If $k_{T-1,0}>\hat{k}_{T-1}$, i.e., $\lambda_{k_{T-1,0}}<\lambda_{\hat{k}_{T-1}}$. Note that
$$\begin{aligned}
    \left\|\Sigma_{T-1}^{1/2}\left({\theta}_{D, \lambda_{\hat{k}_{T-1}},T-1}-\theta_{T-1}^*\right)\right\|_2 
\le &\sqrt{\frac{1}{1-\tilde{c}}}\left\|\left(\widehat{\Sigma}_{D,T-1}+\lambda_{\hat{k}_{T-1}} I\right)^{1 / 2}\left({\theta}_{D, \lambda_{\hat{k}_{T-1}},T-1}-{\theta}_{D, \lambda_{k_{T-1,0}},T-1}\right)\right\|_2\notag\\
&+\left\|(\Sigma_{T-1}+\lambda_{k_{T-1,0}}I)^{1/2}\left({\theta}_{D, \lambda_{k_{T-1,0}},T-1}-\theta_{T-1}^*\right)\right\|_2.\notag
\end{aligned}$$
Based on (\ref{2var:bias2}), we obtain
$$\begin{aligned}&\left\|(\Sigma_{T-1}+\lambda_{k_{T-1,0}}I)^{1/2}\left({\theta}_{D, \lambda_{k_{T-1,0}},T-1}-\theta_{T-1}^*\right)\right\|_2\notag\\
\le& C_{3}|D|_\gamma^{-\frac{r+1/2}{2 r+s+1}}\left(1+\left(\log \frac{2}{\delta}\right)^{\min\{1,r\}}\mathbb{I}_{r>1/2}\right)+\frac{1}{1-2 \tilde{c}}\mu^{1/2} 2bC_4(2M+\Phi_{T+1})\notag\\
&\cdot|D|_\gamma^{-\frac{r+1/2}{2r+s+1}} (\log d)^{\frac{2}{\gamma_0}}\log^2 \frac{2}{\delta}\log_q\left(|D|_\gamma^{-1/2}\right) \left(1+\left(\log \frac{2}{\delta}\right)^{\min\{1,r\}}\mathbb{I}_{r>1/2}\right),\notag
\end{aligned}$$
it remains to bound $\sqrt{\frac{1}{1-\tilde{c}}}\left\|\left(\widehat{\Sigma}_{D,{T-1}}+\lambda_{\hat{k}_{T-1}} I\right)^{1 / 2}\left({\theta}_{D, \lambda_{\hat{k}_{T-1}},T-1}-{\theta}_{D, \lambda_{k_{T-1,0}},T-1}\right)\right\|_2$. Due to the definition of $\hat{k}_{T-1}$ yields that
$$
\begin{aligned}
& \sqrt{\frac{1}{1-\tilde{c}}}\left\|\left(\widehat{\Sigma}_{D,{T-1}}+\lambda_{\hat{k}_{T-1}} I\right)^{1 / 2}\left({\theta}_{D, \lambda_{\hat{k}_{T-1}},T-1}-{\theta}_{D, \lambda_{k_{T-1,0}},T-1}\right)\right\|_2\\
\leq &\sqrt{\frac{1}{1-\tilde{c}}} \sum_{k_{T-1}=\hat{k}_{T-1}-1}^{k_{T-1,0}}\left\|\left(\widehat{\Sigma}_{D,T-1}+\lambda_{k_{T-1}+1} I\right)^{1 / 2}\left({\theta}_{D, \lambda_{k_{T-1}+1},T}-{\theta}_{D, \lambda_{k_{T-1}},T-1}\right)\right\|_2 \\
\leq &\sqrt{\frac{1}{1-\tilde{c}}}\sum_{k_{T-1}=\hat{k}_{T-1}-1}^{k_{T-1,0}} 8b\sqrt{\frac{1-\tilde{c}}{1-2 \tilde{c}}}\sqrt{\frac{1}{1-2 \tilde{c}}}\left(84(3M+\Phi_{T})(1+C_x)\mathcal{W}_{D,\lambda_{k_{T-1}+1},T}\log^2 \frac{2}{\delta}\right)\\
\le & \frac{672b}{1-2 \tilde{c}}(3M+\Phi_{T})(1+C_x)|D|_\gamma^{-\frac{r+1/2}{2r+s+1}} (\log d)^{\frac{2}{\gamma_0}}\log^2 \frac{2}{\delta}\log_q\left(|D|_\gamma^{-1/2}\right).
\end{aligned}
$$
Together all the above results, we have
\begin{equation}\begin{aligned}
&\left\|\Sigma_{T-1}^{1/2}\left({\theta}_{D, \lambda_{\hat{k}_{T-1}},T-1}-\theta_{T-1}^*\right)\right\|_2\\
\le& C_6(2M+\Phi_{T+1})|D|_\gamma^{-\frac{r+1/2}{2r+s+1}} (\log d)^{\frac{2}{\gamma_0}}\log^2 \frac{2}{\delta}\log_q\left(|D|_\gamma^{-1/2}\right) \left(1+\left(\log \frac{2}{\delta}\right)^{\min\{1,r\}}\mathbb{I}_{r>1/2}\right)\\
&+C_6(3M+\Phi_{T})(1+C_x)|D|_\gamma^{-\frac{r+1/2}{2r+s+1}} (\log d)^{\frac{2}{\gamma_0}}\log^2 \frac{2}{\delta}\log_q\left(|D|_\gamma^{-1/2}\right).\label{proofT}\end{aligned}
\end{equation}
By repeating this process, we can inductively establish the result for step $t$, thereby completing the proof.
\endproof

\subsubsection{Proof of the generalization
error}

It remains to bound $\Phi_{t+1}$, for which the following lemma establishes a recursive relationship between $\Phi_{t}$ and $\Phi_{t+1}$.

\begin{lemma}\label{lemma:Phi recursive relationship}
Let $0 \leq \delta \leq 1/2$ satisfy
\begin{equation}
\begin{aligned}
\delta \geq 2 \exp \left\{-\frac{\sqrt{2 r+s}}{(\log d)^{\frac{1}{\gamma_0}}\sqrt{\log_q(|D|_\gamma^{-1/2})}}|D|_\gamma^{\frac{r}{4 r+2 s+1}} \right\} .\label{condition of delta}
\end{aligned}
\end{equation}
Under Assumptions \ref{assump:mixing}-\ref{assump:conditional probability} with $r \geq 0$ and $0\leq s \leq 1$, if $\lambda_{\hat{k}_t}$ is chosen by (\ref{adas}) for $t=1, \ldots, T$, then with probability at least $1-\delta$, it holds that
$$
\Phi_t+M \leq C_8 \sum_{\ell=t}^T(T-\ell+2)\left(\Phi_{\ell+1}+M\right), \quad t=1,2, \ldots, T.
$$
\end{lemma}

\proof{Proof.}
Since $\lambda_{\hat{k}_t}$ is determined by (\ref{adas}) for $t=1, \ldots, T$, and by Lemma \ref{thm:as}, with probability at least $1-\delta$
$$
\begin{aligned}
& \left\|\theta_{D, \lambda_{\hat{k}_t}, t}-\theta_t^*\right\|_{2}  \\
\leq & C_{6}\sum_{\ell=t}^T\left((T-\ell+2)M+\Phi_{\ell+1}\right)|D|_\gamma^{-\frac{r}{2r+s+1}} (\log d)^{\frac{2}{\gamma_0}}\log^2 \frac{2}{\delta}\log_q\left(|D|_\gamma^{-1/2}\right)\left(1+\left(\log \frac{2}{\delta}\right)^{\min\{1,r\}}\mathbb{I}_{r>1/2}\right)   \\
\leq &C_7 {(2 r+s) \sum_{\ell=t}^T(T-\ell+2)\left(\Phi_{\ell+1}+M\right)},
\end{aligned}
$$
where $C_7$ is the constant independent of $|D|$ and $\delta$.
Therefore, we have
$$
\begin{aligned}
& |x_t^\top\theta_{D, \lambda_t, t}|+M \leq C_x\left\|\theta_{D, \lambda_t, t}\right\|_{2}+M 
\leq C_x\left\|\theta_{D, \lambda_t, t}-\theta_t^*\right\|_{2}+C_x\left\|\Sigma_t^r\Sigma_t^{-r}\theta_t^*\right\|_{2}+M \\
\leq& C_x\left\|\theta_{D, \lambda_t, t}-\theta_t^*\right\|_{2}+CC_x^{2r+1}+M 
\leq  C_8 \sum_{\ell=t}^T(T-\ell+2)\left(\Phi_{\ell+1}+M\right),
\end{aligned}
$$
where $C_8$ is the constant independent of $|D|$ and $\delta$. This completes the proof of Lemma \ref{lemma:Phi recursive relationship}.
\endproof

Based on the above lemma, we can derive an upper bound of $\Phi_t$.
\begin{proposition}\label{bound of Phi}
Let $0 \leq \delta \leq 1/2$ with $\delta$ satisfying (\ref{condition of delta}). Under Assumptions \ref{assump:mixing}-\ref{assump:conditional probability} with $r \geq 0$ and $0\leq s \leq 1$, if $\lambda_{\hat{k}_t}$ is chosen by (\ref{adas}), then with probability at least $1-\delta$, it holds that
$$
\Phi_t \leq 2 C_8 M\prod_{\ell=t}^{T-1}\left(C_8(T-\ell+2)+1\right) -M.
$$
\end{proposition}
\proof{Proof.}
Since for any $\xi_t, \eta_t>0, \xi_t \leq \sum_{\ell=t}^T \eta_{\ell} \xi_{\ell+1}$ implies $\xi_t \leq \prod_{\ell=t}^{T-1}\left(\eta_{\ell}+1\right) \eta_T \xi_{T+1}$. Set $\xi_t=\Phi_t+M$ and $\eta_\ell=C_8(T-\ell+2)$. We have from $\theta_{D, \lambda_{T+1}, T+1}=0$ that
$$
\Phi_t+M \leq \prod_{\ell=t}^{T-1}\left(C_8(T-\ell+2)+1\right) 2 C_8 M.
$$
This completes the proof of Proposition \ref{bound of Phi}.
\endproof

\proof{Proof of Theorem \ref{thm:main}.}
Because
$$
\begin{aligned}
&E\left[V_1^*\left(S_1\right)-V_{\pi_{D, \vec{\lambda}_{\hat{k}}}, 1}\left(S_1\right)\right] \leq \sum_{t=1}^T 2 \mu^{t / 2} \left\|\theta_{D,\lambda_{\hat{k}_t},t}-\theta_t^*\right\|_{\Sigma_t},\notag\\
\le&\sum_{t=1}^T 2 \mu^{t / 2} C_{6}\sum_{\ell=t}^T\left((T-\ell+2)M+\Phi_{\ell+1}\right)\notag\\
&\cdot|D|_\gamma^{-\frac{r+1/2}{2r+s+1}} (\log d)^{\frac{2}{\gamma_0}}\log^2 \frac{2}{\delta}\log_q\left(|D|_\gamma^{-1/2}\right) \left(1+\left(\log \frac{2}{\delta}\right)^{\min\{1,r\}}\mathbb{I}_{r>1/2}\right)\notag\\
\le&\sum_{t=1}^T 2 \mu^{t / 2} C_{6}\sum_{\ell=t}^T\left((T-\ell+2)M+2 C_4 M\prod_{k=\ell+1}^{T-1}\left(C_4(T-k+2)+1\right) -M\right)\notag\\
&\cdot|D|_\gamma^{-\frac{r+1/2}{2r+s+1}} (\log d)^{\frac{2}{\gamma_0}}\log^2 \frac{2}{\delta}\log_q\left(|D|_\gamma^{-1/2}\right) \left(1+\left(\log \frac{2}{\delta}\right)^{\min\{1,r\}}\mathbb{I}_{r>1/2}\right).\notag
\end{aligned}
$$
This completes the proof.
\endproof
\subsection{Comparison inequality}
\begin{assumption}\label{assump:margin assumption}
    There exist some constants $C_m>0$ and $\alpha \geq 0$ such that
    \begin{footnotesize}
$$
P\left(\max _{a_t \in \mathcal{A}_t} \langle X_t\left(s_{1:t},a_{1:t-1}, a_t\right),\theta_t^*\rangle-\max _{a_t \in \mathcal{A}_t \backslash \arg \max _{a_t} \langle X_t\left(s_{1:t},a_{1:t-1}, a_t\right),\theta_t^*\rangle} \langle X_t\left(s_{1:t},a_{1:t-1}, a_t\right),\theta_t^*\rangle \leq \epsilon_t\right) \leq C \epsilon_t^\alpha\notag
$$
\end{footnotesize}
for all positive $\epsilon_t$ for $t=1, \ldots, T$.
\end{assumption}
\begin{lemma}[\cite{murphy2005generalization}]\label{lemma:equation}
Given policies $\tilde{\pi}$ and $\pi$,
$$
E[V_{\tilde{\pi},1}\left(S_1\right)-V_{\pi,1}\left(S_1\right)]=-E_\pi\left[\sum_{t=1}^T Q_{\tilde{\pi}, t}\left(S_{1:t}, A_{1:t}\right)-V_{\tilde{\pi}, t}\left(S_{1:t}, A_{1:t}\right)\right].
$$
Set $\tilde{\pi}=\pi^*$, we can further obtain that
$$
E[V_1^*\left(S_1\right)-V_{\pi,1}\left(S_1\right)]=-E_\pi\left[\sum_{t=1}^T Q^*_{ t}\left(S_{1:t}, A_{1:t}\right)-V^*_{ t}\left(S_{1:t}, A_{1:t}\right)\right].
$$
\end{lemma}
Based on Lemma \ref{lemma:equation} and equation (\ref{vq general}), we further derive that
\begin{equation}
\begin{aligned}
&E[V_1^*\left(S_1\right)-V_{\pi,1}\left(S_1\right)]\\
=&-E_\pi\left[\sum_{t=1}^T Q^*_{ t}\left(S_{1:t}, A_{1:t}\right)-V^*_{ t}\left(S_{1:t}, A_{1:t}\right)\right]\\
=&E_\pi\left[\sum_{t=1}^T \max _{a_t} Q_t^*\left(S_{1: t}, A_{1: t-1},a_t\right)-Q^*_{ t}\left(S_{1:t}, A_{1:t}\right)\right]\\
=&E_\pi\left[\sum_{t=1}^T \max _{a_t} \langle X_t\left(S_{1: t}, A_{1: t-1},a_t\right),\theta_t^*\rangle- \langle X_t\left(S_{1: t}, A_{1: t}\right),\theta_t^*\rangle\right].\label{murphyproperty}
\end{aligned}
\end{equation}
\begin{lemma}
Suppose Assumptions \ref{assump:conditional probability} and \ref{assump:margin assumption} hold. Then for any parameter vector $\theta_t$, and the policy $\pi=\left(\pi_1, \ldots, \pi_T\right)$ is defined by $\pi_t\left(s_{1: t}, a_{1: t-1}\right)=\arg \max _{a_t \in \mathcal{A}_t} \langle\theta_t,x_t\left(s_{1: t}, a_{1: t-1}, a_t\right)\rangle$, the following inequality holds.
$$
\begin{aligned}
&E\left[V_1^*\left(S_1\right)-V_{\pi, 1}\left(S_1\right)\right] \leq \sum_{t=1}^T C_{1,t} \left\{E\left[\langle \theta_t-\theta_t^*,X_t\rangle^2\right]\right\}^{(1+\alpha)/(2+\alpha)}\notag\\
=&\sum_{t=1}^T C_{1,t} \left\|\theta_t-\theta_t^*\right\|_{E[X_tX_t^\top]}^{(2+2\alpha)/(2+\alpha)}:=\sum_{t=1}^T 2 \mu^{t / 2} \left\|\theta_t-\theta_t^*\right\|_{\Sigma_t}^{(2+2\alpha)/(2+\alpha)},\notag
\end{aligned}
$$
where $\|z\|_A^2=z^\top A z$ denotes the weighted 2-norm of the vector $z\in\mathbb{R}^d$ with respect to a positive definite matrix $A\in\mathbb{R}^{d\times d}$.
\end{lemma}
\proof{Proof.}
For any policy $\pi=\left(\pi_1, \ldots, \pi_T\right)$, denote
$$
\Delta \left(\langle X_t\left(S_{1:t},A_{1:t-1}\right),\theta_t^*\rangle \right)=\max _{a_t} \langle X_t\left(S_{1: t}, A_{1: t-1},a_t\right),\theta_t^*\rangle- \langle X_t\left(S_{1: t}, A_{1: t}\right),\theta_t^*\rangle
$$
for $t=1, \ldots, T$. Following (\ref{murphyproperty}), we have
$$
\begin{aligned}
&E[V_1^*\left(S_1\right)-V_{\pi,1}(S_1)]\notag\\
=&E_\pi\left[\sum_{t=1}^T\left[\max _{a_t} \langle X_t\left(S_{1: t}, A_{1: t-1},a_t\right),\theta_t^*\rangle- \langle X_t\left(S_{1: t}, A_{1: t}\right),\theta_t^*\rangle\right]\right]
=\sum_{t=1}^T E_\pi\left[\Delta \left(\langle X_t\left(S_{1:t},A_{1:t-1}\right),\theta_t^*\rangle \right)\right] .\notag
\end{aligned}
$$
Define the event
$$
\Omega_{\epsilon_t, t}=\left\{\max _{a_t \in \mathcal{A}_t} \langle X_t\left(S_{1: t}, A_{1: t-1},a_t\right),\theta_t^*\rangle-\max _{a_t \in \mathcal{A}_t \backslash \arg \max _{a_t} \langle X_t\left(S_{1: t}, A_{1: t-1},a_t\right),\theta_t^*\rangle} \langle X_t\left(S_{1: t}, A_{1: t-1},a_t\right),\theta_t^*\rangle \leq \epsilon_t\right\} .
$$
Then on the event $\Omega_{\epsilon_t, t}^c$, we have $\Delta \left(\langle X_t\left(S_{1:t},A_{1:t-1}\right),\theta_t^*\rangle \right) \leq\left[\Delta \left(\langle X_t\left(S_{1:t},A_{1:t-1}\right),\theta_t^*\rangle \right)\right]^2 / \epsilon_t$. Thus, given that $2\sqrt{ab}\le a+b$ for $a,b\ge 0$, by setting $a=\frac{\Delta \left(\langle X_t\left(S_{1:t},A_{1:t-1}\right),\theta_t^*\rangle \right)}{\sqrt{\epsilon_t}}$ and $b=\frac{\sqrt{\epsilon}}{2}$, we have
\begin{equation}
\begin{aligned}
&E[V_1^*\left(S_1\right)-V_{\pi,1}(S_1)]\\
=&\sum_{t=1}^T E_\pi\left[1_{\Omega_{\epsilon_t, t}^C} \Delta \left(\langle X_t\left(S_{1:t},A_{1:t-1}\right),\theta_t^*\rangle \right)+1_{\Omega_{\epsilon_t, t}} \Delta \left(\langle X_t\left(S_{1:t},A_{1:t-1}\right),\theta_t^*\rangle \right)\right]\\
 \leq& \sum_{t=1}^T E_\pi\left[1_{\Omega_{\epsilon_t, t}^C} \frac{\left(\Delta \left(\langle X_t\left(S_{1:t},A_{1:t-1}\right),\theta_t^*\rangle \right)\right)^2}{\epsilon_t}+1_{\Omega_{\epsilon_t, t}}\left(\frac{\left(\Delta \left(\langle X_t\left(S_{1:t},A_{1:t-1}\right),\theta_t^*\rangle \right)\right)^2}{\epsilon_t}+\frac{\epsilon_t}{4}\right)\right] \\
 =&\sum_{t=1}^T\left[\frac{1}{\epsilon_t} E_\pi\left(\Delta \left(\langle X_t\left(S_{1:t},A_{1:t-1}\right),\theta_t^*\rangle \right)\right)^2+\frac{\epsilon_t}{4} E_\pi\left(1_{\Omega_{\epsilon_t, t}}\right)\right].\label{comparison total}
\end{aligned}
\end{equation}
By Assumptions \ref{assump:conditional probability} and \ref{assump:margin assumption}, there holds
\begin{equation}
\begin{aligned}
E_\pi (1_{\Omega_{\epsilon_t, t}})=E\left[\prod_{\ell=1}^{t-1} \frac{1_{A_\ell=\pi_\ell\left(S_{1:\ell},A_{1:\ell-1}\right)}}{p_\ell\left(A_\ell \mid S_{1:\ell},A_{1:\ell-1}\right)} 1_{\Omega_{\epsilon_t, t}}\right] \leq \mu^{t-1} C_m \epsilon_t^\alpha.\label{comparison1}
\end{aligned}
\end{equation}
In addition, note that
\begin{equation}
\begin{aligned}
&E_\pi\left[  \Delta \left(\langle X_t\left(S_{1:t},A_{1:t-1}\right),\theta_t^*\rangle \right)\right]^2 \\
= & E_\pi\left[\max _{a_t} \langle X_t\left(S_{1: t}, A_{1: t-1},a_t\right),\theta_t^*\rangle-\max _{a_t} \langle X_t\left(S_{1: t}, A_{1: t-1},a_t\right),\theta_t\rangle\right.\\
&\left.+\langle X_t\left(S_{1: t}, A_{1: t-1},\pi_t(S_{1:t},A_{1:t-1})\right),\theta_t\rangle-\langle X_t\left(S_{1: t}, A_{1: t}\right),\theta_t^*\rangle\right]^2 \\
\leq & 2 E_\pi\left[\max _{a_t} \langle X_t\left(S_{1: t}, A_{1: t-1},a_t\right),\theta_t^*\rangle-\max _{a_t} \langle X_t\left(S_{1: t}, A_{1: t-1},a_t\right),\theta_t\rangle\right]^2 \\
& +2 E_\pi\left[\langle X_t\left(S_{1: t}, A_{1: t-1},\pi_t(S_{1:t},A_{1:t-1})\right),\theta_t\rangle-\langle X_t\left(S_{1: t}, A_{1: t-1},\pi_t(S_{1:t},A_{1:t-1})\right),\theta_t^*\rangle\right]^2 \\
\leq & 4 E_\pi\left(\max _{a_t}\left[\langle X_t\left(S_{1: t}, A_{1: t-1},a_t\right),\theta_t^*\rangle-\langle X_t\left(S_{1: t}, A_{1: t-1},a_t\right),\theta_t\rangle\right]^2\right) \\
= & 4 E\left(\prod_{\ell=1}^{t-1} \frac{1_{A_\ell=\pi_\ell\left(S_{1:\ell},A_{1:\ell-1}\right)}}{p_\ell\left(A_\ell \mid S_{1:\ell},A_{1:\ell-1}\right)} \frac{1_{A_t \in \arg \max _{a_t}\left[\langle X_t\left(S_{1: t}, A_{1: t-1},a_t\right),\theta_t^*\rangle-\langle X_t\left(S_{1: t}, A_{1: t-1},a_t\right),\theta_t\rangle\right]^2}}{p_t\left(A_t \mid S_{1:t},A_{1:t-1}\right)}\right. \\
& \left.\times\left[\langle X_t\left(S_{1: t}, A_{1: t-1},A_t\right),\theta_t^*\rangle-\langle X_t\left(S_{1: t}, A_{1: t-1},A_t\right),\theta_t\rangle\right]^2\right) \\
\leq & 4 \mu^t E\left[\langle X_t\left(S_{1: t}, A_{1: t-1},A_t\right),\theta_t^*\rangle-\langle X_t\left(S_{1: t}, A_{1: t-1},A_t\right),\theta_t\rangle\right]^2.\label{comparison 2}
\end{aligned}
\end{equation}
Plugging (\ref{comparison1}) and (\ref{comparison 2}) into (\ref{comparison total}) yields
$$
\begin{aligned}
&E[V_1^*\left(S_1\right)-V_{\pi,1}(S_1)] \notag\\
\leq& \sum_{t=1}^T\left[\frac{1}{\epsilon_t} 4 \mu^t E\left[\langle X_t\left(S_{1: t}, A_{1: t-1},A_t\right),\theta_t^*\rangle-\langle X_t\left(S_{1: t}, A_{1: t-1},A_t\right),\theta_t\rangle\right]^2+\frac{1}{4} \mu^{t-1} C_m \epsilon_t^{\alpha+1}\right].\notag
\end{aligned}
$$
By choosing 
$$\epsilon_t=\left\{16 \mu E\left[\langle X_t\left(S_{1: t}, A_{1: t-1},A_t\right),\theta_t^*\rangle-\langle X_t\left(S_{1: t}, A_{1: t-1},A_t\right),\theta_t\rangle\right]^2 /[(1+\alpha) C_m]\right\}^{1 /(2+\alpha)}$$ to minimize the above upper bound, we have
$$
E[V_1^*\left(S_1\right)-V_{\pi,1}(S_1)] \leq \sum_{t=1}^T C_{1, t}\left\{E\left[\langle X_t,\theta_t^*\rangle-\langle X_t,\theta_t\rangle\right]^2\right\}^{(1+\alpha) /(2+\alpha)},
$$
where $C_{1, t}=(2+\alpha)\left[2^{2 \alpha}(1+\alpha)^{-(1+\alpha)} \mu^{(2+\alpha) t-1} C_m\right]^{1 /(2+\alpha)}$.
\endproof

\end{APPENDICES}

\end{document}